\documentclass[journal]{IEEEtran}
\usepackage[english]{babel}
\usepackage{cite}

\pdfoutput=1 

\ifCLASSINFOpdf
  \usepackage[pdftex]{graphicx}
\else
  \usepackage[dvips]{graphicx}
  \graphicspath{{./Figures/eps/}}
\fi

\usepackage{amsmath}
\usepackage{amssymb}
\usepackage{color}
\usepackage[usenames,dvipsnames,svgnames,table]{xcolor}
\usepackage{booktabs}       
\usepackage[ruled,section]{algorithm}
\usepackage[noend]{algorithmic}
\usepackage{array}
\ifCLASSOPTIONcompsoc
  \usepackage[caption=false,font=normalsize,labelfont=sf,textfont=sf]{subfig}
\else
  \usepackage[caption=false,font=footnotesize]{subfig}
\fi

\usepackage{epstopdf}
\usepackage{url}
\usepackage{wrapfig}
\usepackage{tikz}
\usepackage{pgfplots}
\usetikzlibrary{arrows,positioning,automata,bayesnet}
\usepgfplotslibrary{external} 
\allowdisplaybreaks
\allowbreak

\DeclareMathOperator*{\argmin}{argmin}

\pgfplotsset{
    compat=newest,
    every axis/.append style={
        legend image post style={xscale=0.5}
    }
}

\begin{document}
\title{Learning Methods for Dynamic Topic Modeling in Automated Behaviour Analysis}
\author{Olga~Isupova,
        Danil~Kuzin,
        and~Lyudmila~Mihaylova,~\IEEEmembership{Senior Member,~IEEE}
\thanks{O.Isupova, D.Kuzin, L.Mihaylova are with the Department
of Automatic Control and Systems Engineering, The University of Sheffield, Sheffield,
UK e-mail: o.isupova@sheffield.ac.uk, dkuzin1@sheffield.ac.uk, l.s.mihaylova@sheffield.ac.uk}}


\maketitle

\begin{abstract}
Semi-supervised and unsupervised systems provide operators with
invaluable support and can tremendously reduce the operators' load.
In the light of the necessity to process large volumes of video data
and provide autonomous decisions, this work proposes new learning
algorithms for activity analysis in video. The activities and behaviours are described by a dynamic topic model.
Two novel learning algorithms based on the
expectation maximisation approach and variational Bayes inference
are proposed. Theoretical derivations of the posterior estimates of model parameters are given. The designed learning algorithms are compared with the Gibbs sampling inference scheme introduced earlier in the literature.
A detailed comparison of the learning algorithms is presented on
real video data. We also propose an anomaly localisation procedure, elegantly embedded in the topic modeling framework. 
It is shown that the developed learning algorithms can achieve $95\%$
success rate.
The proposed framework can be applied to a number of areas,
including transportation systems, security and surveillance.  
\end{abstract}


\section{Introduction}

\IEEEPARstart{B}{ehaviour} analysis is an important area in intelligent video surveillance, where abnormal behaviour detection is a difficult problem. One of the challenges in this field is informality of the problem formulation. Due to the broad scope of applications and desired objectives there is no unique way, in which normal or abnormal behaviour can be described.  In general, the objective is to detect unusual events and inform in due course a human operator about them.

This paper considers a probabilistic framework for anomaly detection, where less probable events are labelled as abnormal. We propose two learning algorithms and an anomaly localisation procedure for spatial detection of abnormal behaviours. 
  
\subsection{Related work}
There is a wealth of methods for abnormal behaviour detection, for example, pattern-based methods \cite{Raghavendra2011, Yen13, Ouivirach13}. These methods extract explicit patterns from data and use them as behaviour templates for decision making. In~\cite{Raghavendra2011} the sum of the visual features of a reference frame is treated as a normal behaviour template. Another common approach for representing normal templates is using clusters of visual features~\cite{Yen13, Ouivirach13}. Visual features can range from raw intensity values of pixels to complex features that exploit the data nature~\cite{Zhou2016}. 

In the testing stage new observations are compared with the extracted patterns. The comparison is based on some similarity measure between observations, e.g., the Jensen-Shannon divergence in~\cite{Su2014} or the $Z$-score value in~\cite{Yen13, Ouivirach13}. If the distance between the new observation and any of the normal patterns is larger than a threshold, then the observation is classified as abnormal.

Abnormal behaviour detection can be considered as a classification problem. It is difficult in advance to collect and label all kind of abnormalities. Therefore, only one class label can be expected and one-class classifiers are applied to abnormal behaviour detection: e.g., a one-class Support Vector Machine~\cite{Cheng2013}, a support vector data description algorithm~\cite{Liu2010}, a neural network approach~\cite{Maddalena2013}, a level set method~\cite{Osher1988} for normal data boundary determination~\cite{Ding2015}. 

Another class of methods rely on the estimation of probability distributions of the visual data. These estimated distributions are then used in the decision making process. Different kinds of probability estimation algorithms are proposed in the literature, e.g., based on non-parametric sample histograms~\cite{Adam2008}, Gaussian distribution modelling~\cite{Basharat2008}. Spatio-temporal motion data dependency is modelled as a coupled Hidden Markov Model in~\cite{Kratz2009}. Auto-regressive process modelling based on self-organised maps is proposed in~\cite{Brighenti2011}.   

An efficient approach is to seek for feature sets that tend to appear together. These feature sets form typical activities or behaviours in the scene. Topic modeling~\cite{Hofmann99, Blei03LDA} is an approach to find such kinds of statistical regularities in a form of probability distributions. The approach can be applied for abnormal behaviour detection, e.g.,~\cite{Mehran09, Li2008, Varadarajan2009}. A number of variations of the conventional topic models for abnormal behaviour detection have been recently proposed: clustering of activity distributions~\cite{Wang09}, modelling temporal dependencies among activities ~\cite{Hospedales2011}, a continuous model for an object velocity~\cite{Jeong14}.  

Within the probabilistic modelling approach~\cite{Jeong14, Li2008, Mehran09, Basharat2008, Wang09, Kratz2009} the decision about abnormality is mainly made by computing likelihood of a new observation. The comparison of the different abnormality measures based on the likelihood estimation is provided in~\cite{Varadarajan2009}. 

Topic modeling is originally developed for text mining~\cite{Hofmann99, Blei03LDA}. It aims to find latent variables called \emph{``topics''} given the collection of unlabelled text \emph{documents} consisted of \emph{words}. In probabilistic topic modeling documents are represented as a mixture of topics, where each topic is assumed to be a distribution over words.

There are two main types of topic models: Probabilistic Latent Semantic Analysis (PLSA)~\cite{Hofmann99} and Latent Dirichlet Allocation (LDA)~\cite{Blei03LDA}. The former considers the problem from the frequentist perspective while the later studies it within the Bayesian approach. 
The main learning techniques proposed for these models include maximum likelihood estimation via the Expectation-Maximisation (EM) algorithm~\cite{Hofmann99}, variational Bayes inference~\cite{Blei03LDA}, Gibbs sampling~\cite{Griffiths2004} and Maximum a Posteriori (MAP) estimation~\cite{Chien2008}. 

\subsection{Contributions}
In this paper inspired by ideas from~\cite{Hospedales2011} we propose an unsupervised learning framework based on a Markov Clustering Topic Model for behaviour analysis and anomaly detection. It groups possible topic mixtures of visual documents and forms a Markov chain for the groups. 

The key contributions of this work consist in developing new learning algorithms, namely MAP estimation using the EM-algorithm and variational Bayes inference for the  Markov Clustering Topic Model (MCTM), and in proposing an anomaly localisation procedure that follows concepts of probabilistic topic modeling. We derive the likelihood expressions as a normality measure of newly observed data.  The developed learning algorithms are compared with the Gibbs sampling scheme proposed in~\cite{Hospedales2011}.
A comprehensive analysis of the algorithms is presented over real video sequences. The empirical results show that the proposed methods provide more accurate results than the Gibbs sampling scheme in terms of anomaly detection performance. 

Our preliminary results with the EM-algorithm for behaviour analysis are published in~\cite{Isupova2015}. In contrast to~\cite{Isupova2015} we now consider a fully Bayesian framework, where we  propose the EM-algorithm for MAP estimates rather than the maximum likelihood ones. We also propose here a novel learning algorithm based on variational Bayes inference and a novel anomaly localisation procedure. The experiments are performed on more challenging datasets in comparison to~\cite{Isupova2015}.  

The rest of the paper is organised as follows. Section~\ref{sec:features} describes the overall structure of visual documents and visual words. Section~\ref{sec:model} introduces the dynamic topic model. The new learning algorithms are presented in Section~\ref{sec:inference}, where the proposed MAP estimation via the EM-algorithm and variational Bayes algorithm are introduced first and then the Gibbs sampling scheme is reviewed. The methods are given with a detailed discussion about their similarities and differences. The anomaly detection procedure is presented in Section~\ref{sec:abnormality}. The learning algorithms are evaluated with real data in Section~\ref{sec:experiments} and Section~\ref{sec:conlusion} concludes the paper.

\begin{figure}[!t]
\centering
\includegraphics[width = 0.95\columnwidth]{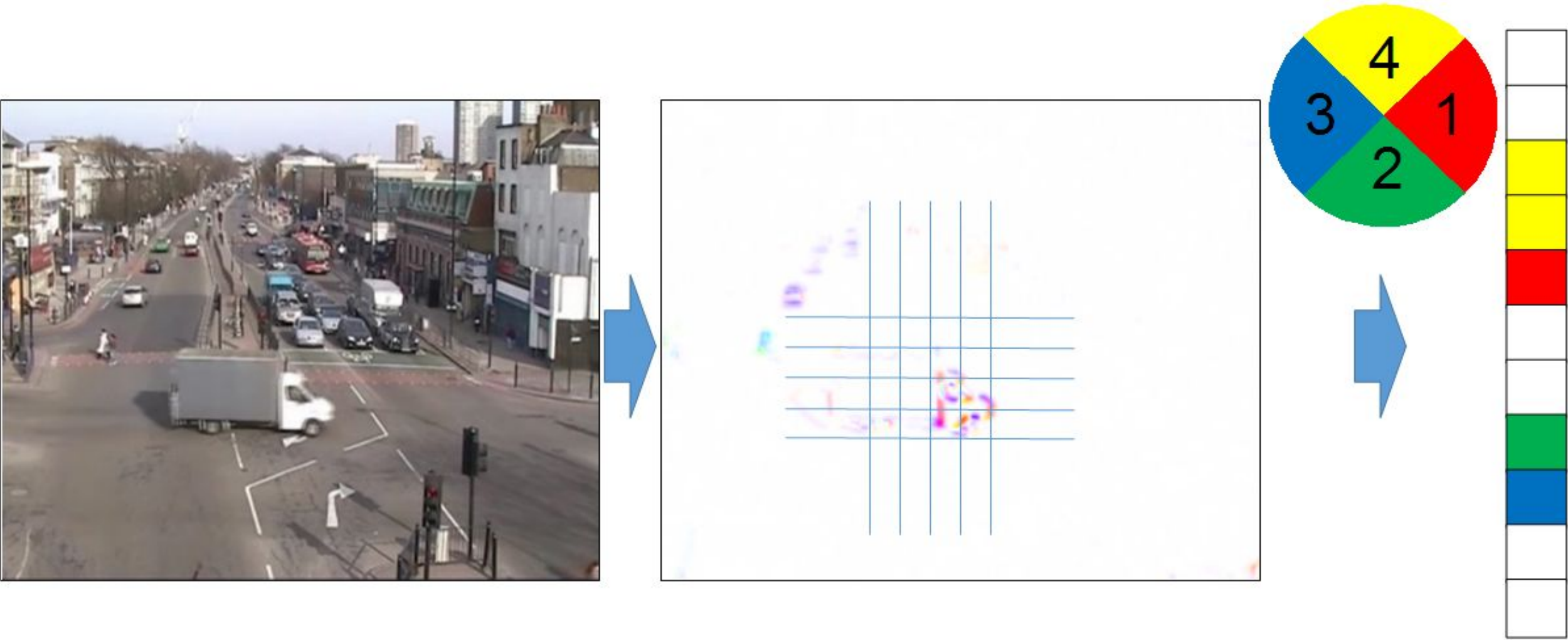}%
\caption{Structure of the visual feature extraction: from an input
frame (on the left) a map of local motions is calculated
(in the centre). The motion is quantised into four directions to get the feature
representation (on the right).}
\label{fig:feature_extraction}
\end{figure}

\section{Video analytics within the topic modeling framework}
\label{sec:features}
Video analytics tasks can be formulated within
the framework of topics modeling. This requires a definition of visual documents and visual words, e.g., as in~\cite{Wang09, Hospedales2011}. The whole video sequence is divided into non-overlapping short clips. These clips are treated as visual documents. Each frame is divided next into grid cells of pixels. Motion detection is applied to each of the cells. The cells where motion is detected are called moving cells. For each of the moving cells the motion direction is determined. This direction is further quantised into four dominant ones - up, left, down, right (see Figure~\ref{fig:feature_extraction}). The position of the moving cell and the quantised direction of its motion form a visual word. 

Each of the visual documents is then represented as a sequence of visual words' identifiers, where identifiers are obtained by some ordering of a set of unique words. This discrete representation of the input data can be processed by topic modeling methods. 

\section{The Markov Clustering Topic Model for behavioural analysis}
\label{sec:model}

\subsection{Motivation}
In topic modeling there are two main kinds of distributions --- the distributions over words, which correspond to topics, and the distributions over topics, which characterise the documents. The relationship between documents and words is then represented via latent low-dimensional entities called topics. Having only an unlabelled collection of documents, topic modeling methods restore a hidden structure of data, i.e., the distributions over words and the distributions over topics. 

Consider a set of distributions over topics and a topic distribution for each document is chosen from this set. 
If the cardinality of the set of distributions over topics is less than the number of documents, then documents are clustered into groups such that documents have the same topic distribution within a group. A unique distribution over topics is called a \textit{behaviour} in this work. Therefore, each document corresponds to one behaviour. In topic modeling a document is fully described by a corresponding distribution over topics, which means in this case a document is fully described by a corresponding behaviour.

There are a number of applications where we can observe documents clustered into groups with the same distribution over topics.
Let us consider some examples from video analytics where a visual word corresponds to a motion within a tiny cell. As topics represent words that statistically often appear together, in video analytics applications topics define some motion patterns in local areas. 

Let us consider a road junction regulated by traffic lights. A general motion on the junction is the same with the same traffic light regime. Therefore, the documents associated to the same traffic light regimes have the same distributions over topics, i.e., they correspond to the same behaviours.

Another example is a video stream generated by a video surveillance camera from a train station. Here it is also possible to distinguish several types of general motion within the camera scene: getting off and on a train and waiting for it. These types of motion correspond to behaviours, where the different visual documents showing different instances of the same behaviour have very similar motion structures, i.e., the same topic distribution. 

Each action in real life lasts for some time, e.g.,
a traffic light regime stays the same and people get on and off a train for several seconds. Moreover, often these different types of motion or behaviours follow a cycle and their changes occur in some order. These insights motivate to model a sequence of behaviours as a Markov chain, so that the behaviours remain the same during some documents and change in a predefined order. The model that has these described properties is called a Markov Clustering Topic Model~(MCTM) in~\cite{Hospedales2011}. The next section formally formulates the model.

\subsection{Model formulation}
This section starts from the introduction of the main notations used through the paper. Denote by $\mathcal{X}$ the vocabulary of all visual words, by $\mathcal{Y}$ the set of all topics and by $\mathcal{Z}$ the set of all behaviours, $x$, $y$ and $z$ are used for elements from these sets, respectively. When an additional element of a set is required it is denoted with a prime, e.g., $z'$ is another element from $\mathcal{Z}$.  

Let $\mathbf{x}_t = \{x_{i, t}\}_{i = 1}^{N_t}$ be a set of words for the document~$t$, where $N_t$ is the length of the document $t$. Let $\mathbf{x}_{1:T_{tr}} = \{\mathbf{x}_t\}_{t = 1}^{T_{tr}}$ denote a set of all words for the whole dataset, where $T_{tr}$ is the number of documents in the dataset. Similarly, denote by $\mathbf{y}_t = \{y_{i, t}\}_{i = 1}^{N_t}$ and $\mathbf{y}_{1:T_{tr}} = \{\mathbf{y}_t\}_{t = 1}^{T_{tr}}$ a set of topics for the document $t$ and a set of all topics for the whole dataset, respectively. Let $\mathbf{z}_{1:T_{tr}} = \{z_t\}_{t = 1}^{T_{tr}}$ be a set of all behaviours for all documents.

Note that $x$, $y$ and $z$ without subscript denote possible values for a word, topic and behaviour from $\mathcal{X}$, $\mathcal{Y}$ and $\mathcal{Z}$, respectively, while the symbols with subscript denote word, topic and behaviour assignments in particular places in a dataset.

Here, $\boldsymbol{\Phi}$ is a matrix corresponding to the distributions over words given the topics, $\boldsymbol{\Theta}$ is a matrix corresponding to the distributions over topics given behaviours. For a Markov chain of behaviours a vector $\boldsymbol{\pi}$ for a behaviour distribution for the first document and a matrix $\boldsymbol{\Xi}$ for transition probability distributions between the behaviours are introduced: 
\begin{align*}
\boldsymbol{\Phi} &= \{\phi_{x, y}\}_{x \in \mathcal{X}, y \in \mathcal{Y}}, &\phi_{x, y} &= p(x | y), &\boldsymbol{\phi}_y &= \{\phi_{x, y}\}_{x \in \mathcal{X}};\\
\boldsymbol{\Theta} &= \{\theta_{y, z}\}_{y \in \mathcal{Y}, z \in \mathcal{Z}}, &\theta_{y, z} &= p(y | z), &\boldsymbol{\theta_z} &= \{\theta_{y, z}\}_{y \in \mathcal{Y}};\\
\boldsymbol{\pi} &= \{\pi_z\}_{z \in \mathcal{Z}}, &\pi_z &= p(z); \\
\boldsymbol{\Xi} &= \{\xi_{z', z}\}_{z' \in \mathcal{Z}, z \in \mathcal{Z}}, &\xi_{z', z} &= p(z' | z), &\boldsymbol{\xi}_z &= \{\xi_{z', z}\}_{z' \in \mathcal{Z}},
\end{align*}
where the matrices $\boldsymbol{\Phi}$, $\boldsymbol{\Theta}$ and $\boldsymbol{\Xi}$ and the vector $\boldsymbol{\pi}$ are formed as follows. An element of a matrix on the $i$-th row and $j$-th column is a probability of the $i$-th element given the $j$-th one, e.g., $\phi_{x, y}$ is a probability of the word $x$ in the topic~$y$. The columns of the matrices are then distributions for corresponding elements, e.g., $\boldsymbol\theta_z$ is a distribution over topics for the behaviour $z$. Elements of the vector $\boldsymbol\pi$ are probabilities of behaviours to be chosen by the first document. All these distributions are categorical. 

The introduced distributions form a set 
\begin{equation}
\boldsymbol{\Omega} = \{\boldsymbol{\Phi}, \boldsymbol{\Theta}, \boldsymbol{\pi}, \boldsymbol{\Xi}\}
\end{equation}
of model parameters and they are estimated during a learning procedure.

Prior distributions are imposed to all the  parameters. Conjugate Dirichlet distributions are used:
\begin{align*}
\boldsymbol{\phi}_y &\sim Dir(\boldsymbol{\phi}_y | \boldsymbol{\beta}), &\forall y \in \mathcal{Y};\\
\boldsymbol{\theta}_z  &\sim Dir(\boldsymbol{\theta}_z | \boldsymbol{\alpha}), &\forall z \in \mathcal{Z};\\
\boldsymbol{\pi} &\sim Dir(\boldsymbol{\pi} | \boldsymbol{\eta}); \\
\boldsymbol{\xi}_z &\sim Dir(\boldsymbol{\xi}_z | \boldsymbol{\gamma}), &\forall z \in \mathcal{Z},
\end{align*} 
where $Dir(\cdot)$ is a Dirichlet distribution and $\boldsymbol{\beta}$, $\boldsymbol{\alpha}$, $\boldsymbol{\eta}$ and $\boldsymbol{\gamma}$ are the corresponding hyperparameters. As topics and behaviours are not known a priori and will be specified via the learning procedure, it is impossible to distinguish two topics or two behaviours in advance. This is the reason why all the prior distributions are the same for all topics and all behaviours.

\begin{figure}[!t]
\begin{center}
\includegraphics{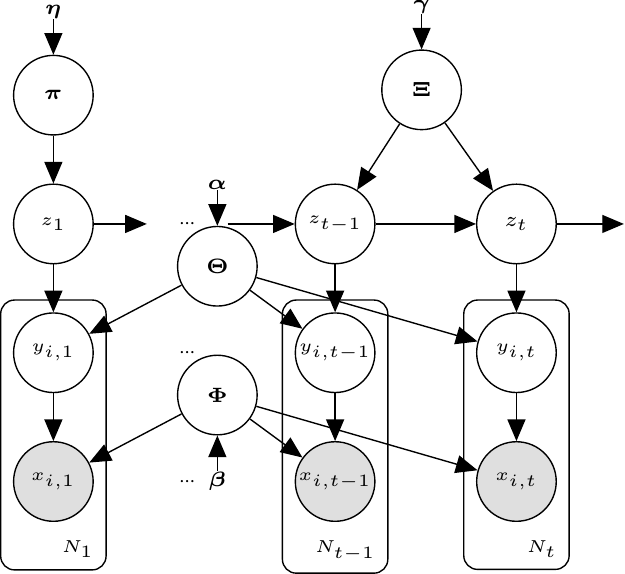}
\caption{Graphical representation of the Markov Clustering Topic Model.}
\label{fig:graph_model}
\end{center}
\end{figure}

The generative process for the model is as follows. All the parameters are drawn from the corresponding prior Dirichlet distributions. At each time moment $t$ a behaviour $z_t$ is chosen first for a visual document. The behaviour is sampled using the matrix $\boldsymbol{\Xi}$ according to the behaviour chosen for the previous document. For the first document the behaviour is sampled using the vector $\boldsymbol\pi$. 

Once the behaviour is selected, the procedure of choosing visual words repeats for the number of times equal to the length of the current document $N_t$. The procedure consists of two steps --- sampling a topic $y_{i, t}$ using the matrix $\boldsymbol\Theta$ according to the chosen behaviour $z_t$ followed by sampling a word $x_{i, t}$ using the matrix $\boldsymbol\Phi$ according to the chosen topic $y_{i, t}$ for each token $i \in \{1, \dotsc, N_t\}$, where a token is a particular place inside a document where a word is assigned. The generative process is summarised in Algorithm~\ref{alg:generative_em}. The graphical model, showing the relationships between the variables, can be found in Figure~\ref{fig:graph_model}.

\begin{algorithm}[t]
\caption{The generative process for the MCTM}
\label{alg:generative_em}
\begin{algorithmic}[1]
    \REQUIRE
        The number of clips -- $T_{tr}$, the length of each clip -- $N_t$ $\forall t = \{1, \dotsc, T_{tr}\}$, the hyperparameters -- $\boldsymbol{\beta}$, $\boldsymbol{\alpha}$, $\boldsymbol{\eta}$, $\boldsymbol{\gamma}$;
    \ENSURE
        The dataset $\textbf{x}_{1:T_{tr}} = \{x_{1, 1}, \dotsc, x_{i, t}, \dotsc, x_{N_{T_{tr}}, T_{tr}}\}$;
	\FORALL{$y \in \mathcal{Y}$}
		\STATE draw a word distribution for the topic $y$: $$\boldsymbol\phi_y \sim Dir(\boldsymbol\phi_y | \boldsymbol\beta);$$
	\ENDFOR
	\FORALL{$z \in \mathcal{Z}$}
		\STATE draw a topic distribution for behaviour $z$: $$\boldsymbol\theta_z \sim Dir(\boldsymbol\theta_z | \boldsymbol\alpha);$$
		\STATE draw a transition distribution for behaviour $z$: $$\boldsymbol\xi_z \sim Dir(\boldsymbol\xi_z | \boldsymbol\gamma);$$
	\ENDFOR
	\STATE draw a behaviour probability distribution for the initial document $$\boldsymbol\pi \sim Dir(\boldsymbol\phi | \boldsymbol\eta);$$
	  
    \FORALL{$t \in \{1, \dotsc, T_{tr}\}$}
        \IF{$t = 1$}
            \STATE draw a behaviour for the document from the initial distribution: $z_t \sim Cat(z_t | \boldsymbol{\pi})$\footnotemark;
        \ELSE
            \STATE draw a behaviour for the document based on the behaviour of the previous document: $z_t \sim Cat(z_t | \boldsymbol{\xi}_{z_{t-1}})$;
            \label{alg_step:gen_behaviour}
        \ENDIF
        \FORALL{$i \in \{1, \dotsc, N_t\}$}
            \STATE draw a topic for the token $i$ based on the chosen behaviour: $y_{i, t} \sim Cat(y_{i, t} | \boldsymbol{\theta}_{z_t}$);
            \label{alg_step:gen_topic}
            \STATE draw a visual word for the token $i$ based on the chosen topic: $x_{i, t} \sim Cat(x_{i, t} | \boldsymbol{\phi}_{y_{i, t}})$;
            \label{alg_step:gen_word}
        \ENDFOR
    \ENDFOR
\end{algorithmic}
\end{algorithm}

\footnotetext{Here, $Cat(\cdot | \mathbf{v})$ denotes a categorical distribution, where components of a vector $\mathbf{v}$ are probabilities of a discrete random variable to take one of possible values.}

The full likelihood of the observed variables $\mathbf{x}_{1:T_{tr}}$, the hidden variables $\mathbf{y}_{1:T_{tr}}$ and $\mathbf{z}_{1:T_{tr}}$ and the set of parameters~$\boldsymbol\Omega$ can be written then as follows:

\begin{align}
&p(\mathbf{x}_{1:T_{tr}}, \mathbf{y}_{1:T_{tr}}, \mathbf{z}_{1:T_{tr}}, \boldsymbol\Omega | \boldsymbol\beta, \boldsymbol\alpha, \boldsymbol\eta, \boldsymbol\gamma) = \nonumber\\
& \underbrace{p(\boldsymbol\pi | \boldsymbol\eta) \, p(\boldsymbol\Xi | \boldsymbol\gamma) \, p(\boldsymbol\Theta | \boldsymbol\alpha) \, p(\boldsymbol\Phi | \boldsymbol\beta)}_{\text{Priors}} \times \nonumber\\
\label{eq:full_likelihood}
& \underbrace{p(z_1 | \boldsymbol\pi) \left[\prod\limits_{t = 2}^{T_{tr}} p(z_t | z_{t-1}, \boldsymbol\Xi) \right] \prod\limits_{t = 1}^{T_{tr}} \prod\limits_{i = 1}^{N_t} p(x_{i, t} | y_{i, t}, \boldsymbol\Phi) p(y_{i, t} | z_t, \boldsymbol\Theta)}_{\text{Likelihood}}
\end{align}

In~\cite{Hospedales2011} Gibbs sampling is implemented for parameters learning in the MCTM. We propose two new learning algorithms: based on an EM-algorithm for the MAP estimates of the parameters and based on variational Bayes inference to estimate posterior distributions of the parameters. We introduce the proposed learning algorithms below and briefly review the Gibbs sampling scheme.

\section{Parameters learning}
\label{sec:inference}
\subsection{Learning: EM-algorithm scheme}
\label{sec:em}

We propose a learning algorithm for MAP estimates of the parameters based on the Expectation-Maximisation algorithm~\cite{Dempster77}. The algorithm consists of repeating E and M-steps. Conventionally, the EM-algorithm is applied to get maximum likelihood estimates. In that case the M-step is: 
\begin{equation}
\mathcal{Q}(\boldsymbol\Omega, \boldsymbol\Omega^{\text{old}}) \longrightarrow \max\limits_{\boldsymbol\Omega},
\end{equation}
where $\boldsymbol\Omega^{\text{old}}$ denotes the set of parameters obtained at the previous iteration and $\mathcal{Q}(\boldsymbol\Omega, \boldsymbol\Omega^{\text{old}})$ is the expected logarithm of the full likelihood function of the observed and hidden variables:
\begin{multline}
\mathcal{Q}(\boldsymbol\Omega, \boldsymbol\Omega^{\text{old}}) = \\
\mathbb{E}_{p(\mathbf{y}_{1:T_{tr}}, \mathbf{z}_{1:T_{tr}} | \mathbf{x}_{1:T_{tr}}, \boldsymbol\Omega^{\text{old}})} \log p(\mathbf{x}_{1:T_{tr}}, \mathbf{y}_{1:T_{tr}}, \mathbf{z}_{1:T_{tr}} | \boldsymbol\Omega).
\end{multline}
The subscript of the expectation sign means the distribution, with respect to which the expectation is calculated. During the E-step the posterior distribution of the hidden variables is estimated given the current estimates of the parameters. 

In this paper the EM-algorithm is applied to get MAP estimates instead of traditional maximum likelihood ones. The M-step is modified in this case as:
\begin{equation}
\label{eq:M_step_functional}
\mathcal{Q}(\boldsymbol\Omega, \boldsymbol\Omega^{\text{old}}) + \log p(\boldsymbol\Omega | \boldsymbol\beta, \boldsymbol\alpha, \boldsymbol\eta, \boldsymbol\gamma) \longrightarrow \max\limits_{\boldsymbol\Omega},
\end{equation}
where $p(\boldsymbol\Omega | \boldsymbol\beta, \boldsymbol\alpha, \boldsymbol\eta, \boldsymbol\gamma)$ is the prior distribution of the parameters.

As the hidden variables are discrete, the expectation converts to a sum of all possible values for the whole set of the hidden variables $\{\mathbf{y}_{1:T_{tr}}, \mathbf{z}_{1:T_{tr}}\}$. The substitution of the likelihood expression from (\ref{eq:full_likelihood}) into (\ref{eq:M_step_functional}) allows to marginalise some hidden variables from the sum. The remaining distributions that are required for computing the $\mathcal{Q}$-function are as follows:
\begin{itemize}
\item $p(z_1 = z | \mathbf{x}_{1:T_{tr}}, \boldsymbol\Omega^{\text{old}})$ --- the posterior distribution of a behaviour for the first document;
\item $p(z_t = z', z_{t-1} = z | \mathbf{x}_{1:T_{tr}}, \boldsymbol\Omega^{\text{old}})$ --- the posterior distribution of two behaviours for successive documents;
\item $p(y_{i, t} = y | \mathbf{x}_{1:T_{tr}}, \boldsymbol\Omega^{\text{old}})$ --- the posterior distribution of a topic assignment for a given token;
\item $p(y_{i, t} = y, z_t = z | \mathbf{x}_{1:T_{tr}}, \boldsymbol\Omega^{\text{old}})$ --- the joint posterior distribution of a topic and behaviour assignments for a given token.
\end{itemize}

With the fixed current values for these posterior distributions the estimates of the parameters that maximise the required functional of the M-step (\ref{eq:M_step_functional}) can be computed as:
\begin{align}
\label{eq:M:phi}
\widehat{\phi}_{x, y}^{\, \text{EM}} &= \dfrac{\left(\beta_x + \hat{n}_{x, y}^{\,\text{EM}} - 1\right)_+}{\sum\limits_{x' \in \mathcal{X}} \left(\beta_{x'} + \hat{n}_{x', y}^{\,\text{EM}} - 1\right)_{+}}, &\forall x \in \mathcal{X}, y \in \mathcal{Y};\\
\label{eq:M:theta}
\widehat{\theta}_{y, z}^{\, \text{EM}} &= \dfrac{\left(\alpha_y + \hat{n}_{y, z}^{\,\text{EM}} - 1\right)_+}{\sum\limits_{y' \in \mathcal{Y}} \left(\alpha_{y'} + \hat{n}_{y', z}^{\,\text{EM}} - 1\right)_+}, &\forall y \in \mathcal{Y}, z \in \mathcal{Z};\\
\label{eq:M_psi_k,l}
\widehat{\xi}_{z', z}^{\, \text{EM}} &= \dfrac{\left(\gamma_{z'} + \hat{n}_{z', z}^{\, \text{EM}} - 1\right)_+}{\sum\limits_{\check{z} \in \mathcal{Z}} \left(\gamma_{\check{z}} + \hat{n}_{\check{z}, z}^{\, \text{EM}} - 1\right)_{+}}, &\forall z', z \in \mathcal{Z};\\
\label{eq:M:pi}
\widehat{\pi}_z^{\, \text{EM}} &= \dfrac{\left(\eta_z + \hat{n}_{z}^{\, \text{EM}} - 1\right)_+}{\sum\limits_{z' \in \mathcal{Z}} \left(\eta_{z'} + \hat{n}_{z'}^{\, \text{EM}} - 1\right)_+}, &\forall z \in \mathcal{Z},
\end{align}  
where $(a)_+ \stackrel{\text{def}}{=} \max(a, 0)$~\cite{Vorontsov2014ARTMArticle}; $\beta_x$, $\alpha_y$ and $\gamma_{z'}$ are the elements of the hyperparameter vectors $\boldsymbol\beta$, $\boldsymbol\alpha$ and $\boldsymbol\gamma$, respectively; and $\hat{n}_{x, y}^{\,\text{EM}} = \sum\limits_{t = 1}^{T_{tr}} \sum\limits_{i = 1}^{N_t} p(y_{i, t} = y | \mathbf{x}_{1:T_{tr}}, \boldsymbol\Omega^{\text{old}}) \mathbb{I}(x_{i, t} = x)$ is the expected number of times, when the word $x$ is associated to the topic $y$, where $\mathbb{I}(\cdot)$ is the indicator function; $\hat{n}_{y, z}^{\,\text{EM}} = \sum\limits_{t = 1}^{T_{tr}} \sum\limits_{i = 1}^{N_t} p(y_{i,t} = y, z_t = z | \mathbf{x}_{1:T_{tr}}, \boldsymbol\Omega^{\text{old}})$ is the expected number of times, when the topic $y$ is associated to the behaviour $z$; $\hat{n}_{z}^{\, \text{EM}} = p(z_1 = z | \mathbf{x}_{1:T_{tr}}, \boldsymbol\Omega^{\text{old}})$ is the ``expected number of times'', when the behaviour $z$ is associated to the first document, in this case the ``expected number'' is just a probability, the notation is used for the similarity  with the rest of the parameters; $\hat{n}_{z', z}^{\, \text{EM}} = \sum\limits_{t = 2}^{T_{tr}} p(z_t = z', z_{t - 1} = z| \mathbf{x}_{1:T_{tr}}, \boldsymbol\Omega^{\text{old}})$ is the expected number of times, when the behaviour $z$ is followed by the behaviour $z'$.

During the E-step with the fixed current estimates of the parameters $\boldsymbol\Omega^{\text{old}}$, the updated values for the posterior distributions of the  hidden variables should be computed. The derivation of the updated formulae for these distributions is similar to the Baum-Welch forward-backward algorithm~\cite{Murphy2012}, where the EM-algorithm is applied to the maximum likelihood estimates for a Hidden Markov Model (HMM). This similarity appears because the generative model can be viewed as extension of a HMM.

For effective computation of the required posterior distributions the additional variables $\acute{\alpha}_z(t)$ and $\acute{\beta}_z(t)$ are introduced. A dynamic programming technique is applied for computation of these variables. Having the updated values for $\acute{\alpha}_z(t)$ and $\acute{\beta}_z(t)$ one can update the required posterior distributions of the hidden variables. The E-step is then formulated as follows (for simplification of notation the superscript ``old'' for the parameters variables is omitted inside the formulae):
 
\begin{align}
\label{eq:E:alpha}
&\begin{cases}
\begin{aligned}
&\acute{\alpha}_z(t) = \prod\limits_{i = 1}^{N_t} \sum\limits_{y \in \mathcal{Y}} \phi_{x_{i, t}, y} \, \theta_{y, z} \times\\
&\quad\sum\limits_{z'\in \mathcal{Z}} \acute{\alpha}_{z'}(t-1) \xi_{z, \tilde{z}},  \,\text{if}\, t \geq 2; 
\end{aligned}\\
\acute{\alpha}_{z}(1) = \pi_z \prod\limits_{i = 1}^{N_1} \sum\limits_{y\in \mathcal{Y}} \phi_{x_{i, 1}, y} \, \theta_{y, z};
\end{cases}\\
\label{eq:E:beta}
&\begin{cases}
\begin{aligned}
&\acute{\beta}_{z}(t) = \sum\limits_{z' \in \mathcal{Z}} \acute{\beta}_{z'}(t+1) \xi_{z', z} \times \\
&\quad \prod\limits_{i = 1}^{N_{t+1}} \sum\limits_{y \in \mathcal{Y}} \phi_{x_{i, t+1}, y} \, \theta_{y, z'} , \,\text{if}\, t < T_{tr}; 
\end{aligned}\\
\acute{\beta}_{z}(T_{tr}) = 1;
\end{cases}\\
\label{eq:E:normalisation_const}
&K = \sum\limits_{z \in \mathcal{Z}} \acute{\alpha}_{z}(1) \acute{\beta}_{z}(1);\\
\label{eq:E:z_t}
&p(z_1 | \textbf{x}_{1:T_{tr}}, \boldsymbol\Omega^{\text{old}}) = \dfrac{\acute{\alpha}_{z_1}(1) \acute{\beta}_{z_1}(1)}{K}; \\
\label{eq:E:z_t, z_t-1}
&\begin{aligned}
&p(z_t, z_{t-1} | \textbf{x}_{1:T_{tr}}, \boldsymbol\Omega^{\text{old}}) = \dfrac{\acute{\alpha}_{z_{t-1}}(t-1) \acute{\beta}_{z_t}(t) \xi_{z_t, z_{t-1}}}{K} \times\\
&\quad \prod\limits_{i = 1}^{N_t} \sum\limits_{y \in \mathcal{Y}} \phi_{x_{i, t}, y} \theta_{y, z_t};
\end{aligned}\\
\label{eq:E:y_i,t, z_t}
&\begin{cases}
\begin{aligned}
&p(y_{i, t}, z_t | \textbf{x}_{1:T_{tr}}, \boldsymbol\Omega^{\text{old}})
= \dfrac{\phi_{x_{i, t}, y_{i, t}} \theta_{y_{i, t}, z_t} \acute{\beta}_{z_t}(t)}{K} \times\\
&\quad\sum\limits_{z' \in \mathcal{Z}} \acute{\alpha}_{z'}(t-1) \xi_{z_t, z'} \prod\limits_{\substack{j = 1 \\ j \neq i}}^{N_t} \sum\limits_{y' \in \mathcal{Y}} {\phi_{x_{j, t}, y'} \theta_{y', z_t}},\,\text{if}\, t \geq 2;
\end{aligned} \\
\begin{aligned}
&p(y_{i, 1}, z_1 | \textbf{x}_{1:T_{tr}}, \boldsymbol\Omega^{\text{old}}) 
= \dfrac{\phi_{x_{i, 1}, y_{i, 1}} \theta_{y_{i, 1}, z_1} \acute{\beta}_{z_1}(1)}{K} \times\\
&\quad \pi_{z_1} \prod\limits_{\substack{j = 1 \\ j \neq i}}^{N_1} \sum\limits_{y' \in \mathcal{Y}} {\phi_{x_{j, 1}, y'} \theta_{y', z_1}};
\end{aligned}
\end{cases} \\
\label{eq:E:y_i,t}
&\begin{aligned}
p(y_{i, t} | \textbf{x}_{1:T_{tr}}, \boldsymbol\Omega^{\text{old}}) &= \sum\limits_{z \in \mathcal{Z}} p(y_{i, t}, z | \textbf{x}_{1:T_{tr}}, \boldsymbol\Omega^{\text{old}}),
\end{aligned}
\end{align}
where $K$ is a normalisation constant for all the posterior distributions of the hidden variables. 

Starting with some random initialisation of the parameter estimates, the EM-algorithm iterates the E and M-steps until convergence. The obtained estimates of the parameters are used for further analysis.

\subsection{Learning: Variational Bayes scheme}
\label{sec:vb}

We also propose a learning algorithm based on the variational Bayes (VB) approach~\cite{Jordan1999} to find approximated posterior distributions for both the hidden variables and the parameters. 

In the VB inference scheme the true posterior distribution, in this case the distribution of the parameters and the hidden variables $p(\mathbf{y}_{1:T_{tr}}, \mathbf{z}_{1:T_{tr}}, \boldsymbol\Omega | \mathbf{x}_{1:T_{tr}}, \boldsymbol\eta, \boldsymbol\gamma, \boldsymbol\alpha, \boldsymbol\beta)$, is approximated with a factorised distribution --- $q(\mathbf{y}_{1:T_{tr}}, \mathbf{z}_{1:T_{tr}}, \boldsymbol\Omega)$. The approximation is made to minimise the Kullback-Leibler divergence between the factorised distribution and true one. We factorise the distribution in order to separate the hidden variables and the parameters:
\begin{multline}
\label{eq:vb_factorisation}
\hat{q}(\mathbf{y}_{1:T_{tr}}, \mathbf{z}_{1:T_{tr}}, \boldsymbol\Omega) = \hat{q}(\mathbf{y}_{1:T_{tr}}, \mathbf{z}_{1:T_{tr}}) \hat{q}(\boldsymbol\Omega) \stackrel{\text{def}}{=} \\
\argmin \mathrm{KL} \left(q(\mathbf{y}_{1:T_{tr}}, \mathbf{z}_{1:T_{tr}}) q(\boldsymbol\Omega) || \right.\\
\left.p(\mathbf{y}_{1:T_{tr}}, \mathbf{z}_{1:T_{tr}}, \boldsymbol\Omega | \mathbf{x}_{1:T_{tr}}, \boldsymbol\eta, \boldsymbol\gamma, \boldsymbol\alpha, \boldsymbol\beta)\right),
\end{multline}
where $\mathrm{KL}$ denotes the Kullback-Leibler divergence. The minimisation of the Kullback-Leibler divergence is equivalent to the maximisation of the evidence lower bound (ELBO). The maximisation is done by coordinate ascent~\cite{Jordan1999}.

During the update of the parameters the approximated distribution $q(\boldsymbol\Omega)$ is further factorised:
\begin{equation}
\label{eq:q_param_factorisation}
q(\boldsymbol\Omega) = q(\boldsymbol\pi) q(\boldsymbol\Xi) q(\boldsymbol\Theta) q(\boldsymbol\Phi).
\end{equation}
Note that this factorisation of approximated parameter distributions is a corollary of our model and not an assumption. 

The iterative process of updating the approximated distributions of the parameters and the hidden variables can be formulated as an EM-like algorithm, where during the E-step the approximated distributions of the hidden variables are updated and during the M-step the approximated distributions of the parameters are updated. 

The M-like step is as follows:
\begin{align}
\label{eq:VB:beta}
&\begin{cases}
q(\boldsymbol\Phi) = \prod\limits_{y \in \mathcal{Y}} Dir\left(\boldsymbol\phi_y; \tilde{\boldsymbol\beta}_y\right),\\
\tilde{\beta}_{x, y} = \beta_x + \hat{n}_{x, y}^{\, \text{VB}}, &\forall x \in \mathcal{X}, y \in \mathcal{Y};
\end{cases}\\
&\begin{cases}
q(\boldsymbol\Theta) = \prod\limits_{z \in \mathcal{Z}} Dir(\boldsymbol\theta_z; \tilde{\boldsymbol\alpha}_z),\\
\tilde{\alpha}_{y, z} = \alpha_y + \hat{n}_{y, z}^{\, \text{VB}}, &\forall y \in \mathcal{Y}, z \in \mathcal{Z};
\end{cases}\\
&\begin{cases}
q(\boldsymbol\pi) = Dir(\boldsymbol\pi; \tilde{\boldsymbol\eta}),\\
\tilde{\eta}_z = \eta_z + \hat{n}_z^{\,\text{VB}}, &\forall z \in \mathcal{Z};
\end{cases}\\
\label{eq:VB:gamma}
&\begin{cases}
q(\boldsymbol\Xi) = \prod\limits_{z \in \mathcal{Z}} Dir(\boldsymbol\xi_{z}; \tilde{\boldsymbol\gamma}_z),\\
\tilde{\gamma}_{z', z} = \gamma_{z'} + \hat{n}_{z', z}^{\, \text{VB}}, &\forall z', z \in \mathcal{Z},
\end{cases}
\end{align} 
where $\tilde{\boldsymbol\beta}_y$, $\tilde{\boldsymbol\alpha}_z$, $\tilde{\boldsymbol\eta}$ and $\tilde{\boldsymbol\gamma}_z$ are updated hyperparameters of the corresponding posterior Dirichlet distributions; and $\hat{n}_{x, y}^{\, \text{VB}} = \sum\limits_{t = 1}^{T_{tr}} \sum\limits_{i = 1}^{N_t} \mathbb{I}(x_{i, t} = x) q(y_{i, t} = y)$ is the expected number of times, when the word $x$ is associated with the topic $y$. Here and below the expected number is computed with respect to the approximated posterior distributions of the hidden variables; $\hat{n}_{y, z}^{\, \text{VB}} = \sum\limits_{t = 1}^{T_{tr}} \sum\limits_{i = 1}^{N_t} q(y_{i, t} = y, z_t = z)$ is the expected number of times, when the topic $y$ is associated with the behaviour $z$; $\hat{n}_z^{\,\text{VB}} = q(z_1 = z)$ is the ``expected number'' of times, when the behaviour $z$ is associated to the first document; $\hat{n}_{z', z}^{\, \text{VB}} = \sum\limits_{t = 2}^{T_{tr}} q(z_t = z', z_{t-1} = z)$ is the expected number of times, when the behaviour $z$ is followed by the behaviour~$z'$.

The following additional variables are introduced for the E-like step:
\begin{align}
\label{eq:VB:introduced_pi}
\tilde{\pi}_z &= \exp\left(\psi\left(\tilde{\eta}_z\right) - \psi\left(\sum\limits_{z' \in \mathcal{Z}} \tilde{\eta}_{z'}\right)\right)\\
\tilde{\xi}_{\tilde{z}, z} &= \exp\left(\psi\left(\tilde{\gamma}_{\tilde{z}, z}\right) - \psi\left(\sum\limits_{z' \in \mathcal{Z}} \tilde{\gamma}_{z', z}\right)\right);\\
\tilde{\phi}_{x, y} &= \exp\left(\psi\left(\tilde{\beta}_{x, y} \right) - \psi\left(\sum\limits_{x' \in \mathcal{X}} \tilde{\beta}_{x', y} \right)\right);\\
\label{eq:VB:introduced_theta}
\tilde{\theta}_{y, z} &= \exp\left(\psi\left(\tilde{\alpha}_{y, z}\right) - \psi\left(\sum\limits_{y' \in \mathcal{Y}}\tilde{\alpha}_{y', z}\right)\right),
\end{align}
where $\psi(\cdot)$ is the digamma function.

Using these additional notations, the E-like step is formulated the same as the E-step of the EM-algorithm, replacing everywhere the estimates of the parameters with the corresponding tilde introduced notation and true posterior distributions of the hidden variables with the corresponding approximated ones in (\ref{eq:E:alpha}) -- (\ref{eq:E:y_i,t}).

The point estimates of the parameters can be obtained by expected values of the posterior approximated distributions. An expected value for a Dirichlet distribution (a posterior distribution for all the parameters) is a normalised vector of hyperparameters. Using the expressions for the hyperparameters from (\ref{eq:VB:beta}) -- (\ref{eq:VB:gamma}), the final parameters estimates can be obtained by:
\begin{align}
\label{eq:VB:phi}
\widehat{\phi}_{x, y}^{\,\text{VB}} &= \dfrac{\beta_x + \hat{n}_{x, y}^{\, \text{VB}}}{\sum\limits_{x' \in \mathcal{X}} \left(\beta_{x'} + \hat{n}_{x', y}^{\, \text{VB}}\right)}, &\forall x \in \mathcal{X}, y \in \mathcal{Y};\\
\label{eq:VB:theta}
\widehat{\theta}_{y, z}^{\,\text{VB}} &= \dfrac{\alpha_y + \hat{n}_{y, z}^{\, \text{VB}}}{\sum\limits_{y' \in \mathcal{Y}} \left(\alpha_{y'} + \hat{n}_{y', z}^{\, \text{VB}}\right)}, &\forall y \in \mathcal{Y}, z \in \mathcal{Z};\\
\label{eq:VB:xi}
\widehat{\xi}_{z', z}^{\,\text{VB}} &= \dfrac{\gamma_{z'} + \hat{n}_{z', z}^{\, \text{VB}}}{\sum\limits_{\check{z} \in \mathcal{Z}} \left(\gamma_{\check{z}} + \hat{n}_{\check{z}, z}^{\, \text{VB}}\right)}, &\forall z', z \in \mathcal{Z};\\
\label{eq:VB:pi}
\widehat{\pi}_{z}^{\,\text{VB}} &= \dfrac{\eta_z + \hat{n}_z^{\,\text{VB}}}{\sum\limits_{z' \in \mathcal{Z}} \left(\eta_{z'} + \hat{n}_{z'}^{\,\text{VB}}\right)}, &\forall z \in \mathcal{Z}.
\end{align}  

\subsection{Learning: Gibbs sampling algorithm}
\label{sec:gibbs}

In~\cite{Hospedales2011} the collapsed version of Gibbs sampling (GS) is used for parameter learning in the MCTM. The Markov chain is built to sample only the hidden variables $y_{i, t}$ and $z_t$, while the parameters $\boldsymbol{\Phi}$, $\boldsymbol{\Theta}$ and $\boldsymbol{\Xi}$ are integrated out (note that the distribution for the initial behaviour choice $\boldsymbol\pi$ is not considered in~\cite{Hospedales2011}).

During the burn-in stage the hidden topic and behaviour assignments to each token in the dataset are drawn from the conditional distributions given all the remaining variables. Following the Markov Chain Monte Carlo framework it would draw samples from the posterior distribution $p(\mathbf{y}_{1:T_{tr}}, \mathbf{z}_{1:T_{tr}} | \mathbf{x}_{1:T_{tr}}, \boldsymbol\beta, \boldsymbol\alpha, \boldsymbol\eta, \boldsymbol\gamma)$. From the whole sample for $\{\mathbf{y}_{1:T_{tr}}, \mathbf{z}_{1:T_{tr}}\}$ the parameters can be estimated by~\cite{Griffiths2004}:
\begin{align}
\label{eq:GS:phi}
\widehat{\phi}_{x, y}^{\,\text{GS}} &= \dfrac{\hat{n}_{x, y}^{\,\text{GS}} + \beta_x}{\sum\limits_{x' \in \mathcal{X}} \left(\hat{n}_{x', y}^{\,\text{GS}} + \beta_{x'}\right)}, &\forall x \in \mathcal{X}, y \in \mathcal{Y};\\
\widehat{\theta}_{y, z}^{\,\text{GS}} &= \dfrac{\hat{n}_{y, z}^{\,\text{GS}} + \alpha_y}{\sum\limits_{y' \in \mathcal{Y}} \left(\hat{n}_{y', z}^{\,\text{GS}} + \alpha_{y'} \right)}, &\forall y \in \mathcal{Y}, z \in \mathcal{Z};\\
\label{eq:GS:xi}
\widehat{\xi}_{z', z}^{\,\text{GS}} &= \dfrac{\hat{n}_{z', z}^{\,\text{GS}} + \gamma_{z'}}{\sum\limits_{\check{z} \in \mathcal{Z}} \left(\hat{n}_{\check{z}, z}^{\,\text{GS}} + \gamma_{\check{z}} \right)}, &\forall z', z \in \mathcal{Z},
\end{align}
where $\hat{n}_{x, y}^{\,\text{GS}}$ is the count for the number of times when the word $x$ is associated with the topic $y$, $\hat{n}_{y, z}^{\,\text{GS}}$ is the count for the topic $y$ and the behaviour $z$ pair, $\hat{n}_{z', z}^{\,\text{GS}}$ is the count for the number of times when the behaviour $z$ is followed by the behaviour~$z'$.    

\subsection{Similarities and differences of the learning algorithms}
\label{sec:comparison}

The point parameter estimates for all three learning algorithms (\ref{eq:M:phi}) -- (\ref{eq:M:pi}), (\ref{eq:VB:phi}) -- (\ref{eq:VB:pi}) and (\ref{eq:GS:phi}) -- (\ref{eq:GS:xi}) have a similar form. The EM-algorithm estimates differ up to the hyperparameters reassignment --- adding one to all the hyperparameters in the VB or GS algorithms ends up with the same final equations for the parameters estimates in the EM-algorithm. We explore this in the experimental part. This ``-1'' term in the EM-algorithm formulae (\ref{eq:M:phi}) -- (\ref{eq:M_psi_k,l}) occurs because it uses modes of the posterior distributions while the point estimates obtained by the VB and GS algorithms are means of the corresponding posterior distributions. For a Dirichlet distribution, which is a posterior distribution for all the parameters, mode and mean expressions differ by this ``-1'' term.

The main differences of the methods consist in the ways the counts $n_{x, y}$, $n_{y, z}$ and $n_{z', z}$ are estimated. In the GS algorithm they are calculated by a single sample from the posterior distribution of the hidden variables $p(\mathbf{y}_{1:T_{tr}}, \mathbf{z}_{1:T_{tr}} | \mathbf{x}_{1:T_{tr}}, \boldsymbol\beta, \boldsymbol\alpha, \boldsymbol\gamma)$. In the EM-algorithm the counts are computed as expected numbers of the corresponding events with respect to the posterior distributions of the hidden variables. In the VB algorithm the counts are computed in the same way as in the EM-algorithm up to replacing the true posterior distributions with the approximated ones. 

Our observations for the dynamic topic model confirm the comparison results for the vanilla PLSA and LDA models provided in~\cite{Asuncion2009}.   

\section{Anomaly detection}
\label{sec:abnormality}
This paper presents on-line anomaly detection with the MCTM in video streams. The decision making procedure is divided into two stages. At a learning stage the parameters are estimated using $T_{tr}$ visual documents by one of the learning algorithms, presented in Section~\ref{sec:inference}. After that during a testing stage a decision about abnormality of new upcoming testing documents is made comparing a marginal likelihood of each document with a threshold. The likelihood is computed using the parameters obtained during the learning stage. The threshold is a parameter of the method and can be set empirically, for example, to label 2\% of the testing data as abnormal. This paper presents a comparison of the algorithms (Section \ref{sec:experiments}) using the measure independent of threshold value selection. 

We also propose an anomaly localisation procedure during the testing stage for those visual documents that are labelled as abnormal. This procedure is designed to provide spatial information about anomalies, while documents labelled as abnormal provide temporal detection. The following sections introduce both the anomaly detection procedure on a document level and the anomaly localisation procedure within a video frame. 

\subsection{Abnormal documents detection}
The marginal likelihood of a new visual document $\mathbf{x}_{t+1}$ given all the previous data $\mathbf{x}_{1:t}$ can be used as a normality measure of the document~\cite{Hospedales2011}:
\begin{multline}
\label{eq:online_likelihood_integral}
p(\mathbf{x}_{t+1} | \mathbf{x}_{1:t}) = \\
\iiint p(\mathbf{x}_{t+1} | \mathbf{x}_{1:t}, \boldsymbol{\Phi}, \boldsymbol{\Theta}, \boldsymbol{\Xi}) p(\boldsymbol{\Phi}, \boldsymbol{\Theta}, \boldsymbol{\Xi} | \mathbf{x}_{1:t}) \mathrm{d}\boldsymbol\Phi \mathrm{d}\boldsymbol\Theta \mathrm{d} \boldsymbol{\Xi}.
\end{multline} 

If the likelihood value is small it means that the current document cannot be fitted to the learnt behaviours and topics, which represent typical motion patterns. Therefore, this is an indication for an abnormal event in this document. The decision about abnormality of a document is then made by comparing the marginal likelihood of the document with the threshold.   

In real world applications it is essential to detect anomalies as soon as possible. Hence an approximation of the integral in (\ref{eq:online_likelihood_integral}) is used for efficient computation. The first approximation is based on the assumption that the training dataset is representative for parameter learning, which means that the posterior probability of the parameters would not change if there is more observed data: 
\begin{equation}
p(\boldsymbol{\Phi}, \boldsymbol{\Theta}, \boldsymbol{\Xi} | \mathbf{x}_{1:t}) \approx p(\boldsymbol{\Phi}, \boldsymbol{\Theta}, \boldsymbol{\Xi} | \mathbf{x}_{1:Tr}) \quad \forall t \geq T_{tr}.
\end{equation}

The marginal likelihood can be then approximated as
\begin{multline}
\label{eq:online_likelihood_integral_train}
\iiint p(\mathbf{x}_{t+1} | \mathbf{x}_{1:t}, \boldsymbol{\Phi}, \boldsymbol{\Theta}, \boldsymbol{\Xi}) p(\boldsymbol{\Phi}, \boldsymbol{\Theta}, \boldsymbol{\Xi} | \mathbf{x}_{1:t}) \mathrm{d}\boldsymbol\Phi \mathrm{d}\boldsymbol\Theta \mathrm{d} \boldsymbol{\Xi} \approx \\
\iiint p(\mathbf{x}_{t+1} | \mathbf{x}_{1:t}, \boldsymbol{\Phi}, \boldsymbol{\Theta}, \boldsymbol{\Xi}) p(\boldsymbol{\Phi}, \boldsymbol{\Theta}, \boldsymbol{\Xi} | \mathbf{x}_{1:T_{tr}}) \mathrm{d}\boldsymbol\Phi \mathrm{d}\boldsymbol\Theta \mathrm{d} \boldsymbol{\Xi}.
\end{multline} 

Depending on the algorithm used for learning the integral in (\ref{eq:online_likelihood_integral_train}) can be further approximated in different ways. We consider two types of approximation.

\subsubsection{Plug-in approximation}
The point estimates of the parameters can be plug-in in the integral (\ref{eq:online_likelihood_integral_train}) for approximation:
\begin{multline}
\label{eq:online_likelihood_plugin}
\iiint p(\mathbf{x}_{t+1} | \mathbf{x}_{1:t}, \boldsymbol{\Phi}, \boldsymbol{\Theta}, \boldsymbol{\Xi}) p(\boldsymbol{\Phi}, \boldsymbol{\Theta}, \boldsymbol{\Xi} | \mathbf{x}_{1:Tr}) \mathrm{d}\boldsymbol\Phi \mathrm{d}\boldsymbol\Theta \mathrm{d} \boldsymbol{\Xi} \approx\\
\iiint p(\mathbf{x}_{t+1} | \mathbf{x}_{1:t}, \boldsymbol{\Phi}, \boldsymbol{\Theta}, \boldsymbol{\Xi}) \delta_{\hat{\boldsymbol{\Phi}}}(\boldsymbol{\Phi}) \delta_{\hat{\boldsymbol{\Theta}}}(\boldsymbol{\Theta}), \delta_{\hat{\boldsymbol{\Xi}}}(\boldsymbol{\Xi}) \mathrm{d}\boldsymbol\Phi \mathrm{d}\boldsymbol\Theta \mathrm{d} \boldsymbol{\Xi} = \\
p(\mathbf{x}_{t+1} | \mathbf{x}_{1:t}, \hat{\boldsymbol{\Phi}}, \hat{\boldsymbol{\Theta}}, \hat{\boldsymbol{\Xi}}),
\end{multline} 
where $\delta_{a}(\cdot)$ is the delta-function with the centre in $a$; $\hat{\boldsymbol{\Phi}}$, $\hat{\boldsymbol{\Theta}}$, $\hat{\boldsymbol{\Xi}}$ are point estimates of the parameters, which can be computed by any of the considered learning algorithms using (\ref{eq:M:phi}) -- (\ref{eq:M_psi_k,l}), (\ref{eq:VB:phi}) -- (\ref{eq:VB:xi}) or (\ref{eq:GS:phi}) -- (\ref{eq:GS:xi}). 

The product and sum rules, the conditional independence equations from the generative model are then applied and the final formula for the plug-in approximation is as follows:
\begin{multline}
\label{eq:online_likelihood_final}
p(\mathbf{x}_{t+1} | \mathbf{x}_{1:t}) \approx p(\mathbf{x}_{t+1} | \mathbf{x}_{1:t}, \hat{\boldsymbol{\Phi}}, \hat{\boldsymbol{\Theta}}, \hat{\boldsymbol{\Xi}}) =\\
\sum\limits_{z_t}\sum\limits_{z_{t+1}} \left[ p(\mathbf{x}_{t+1} | z_{t+1}, \hat{\boldsymbol{\Phi}}, \hat{\boldsymbol{\Theta}}) \times \right.\\
\left. p(z_{t+1} | z_t,  \hat{\boldsymbol{\Xi}}) p(z_t | \mathbf{x}_{1:t}, \hat{\boldsymbol{\Phi}}, \hat{\boldsymbol{\Theta}}, \hat{\boldsymbol{\Xi}})\right],
\end{multline}
where the predictive probability of the behaviour for the current document, given the observed data up to the current document, can be computed via the recursive formula:
\begin{multline}
\label{eq:predictive_behaviour}
p(z_{t} | \mathbf{x}_{1:t}, \hat{\boldsymbol{\Phi}}, \hat{\boldsymbol{\Theta}}, \hat{\boldsymbol{\Xi}}) = \\
\sum_{z_{t-1}} \dfrac{p(\mathbf{x}_{t} | z_{t}, \hat{\boldsymbol{\Phi}}, \hat{\boldsymbol{\Theta}}) p(z_{t} | z_{t-1}, \hat{\boldsymbol{\Xi}}) p(z_{t-1} | \mathbf{x}_{1:t-1}, \hat{\boldsymbol{\Phi}}, \hat{\boldsymbol{\Theta}}, \hat{\boldsymbol{\Xi}})}{p(\mathbf{x}_{t} | \mathbf{x}_{1:t-1}, \hat{\boldsymbol{\Phi}}, \hat{\boldsymbol{\Theta}}, \hat{\boldsymbol{\Xi}})}.
\end{multline}

The point estimates can be computed for all three learning algorithms, therefore a normality measure based on the plug-in approximation of the marginal likelihood is applicable for all of them.  

\subsubsection{Monte Carlo approximation}
If samples $\{\boldsymbol{\Phi}^{s}, \boldsymbol{\Theta}^{s}, \boldsymbol{\Xi}^{s}\}$ from the posterior distribution  $p(\boldsymbol{\Phi}, \boldsymbol{\Theta}, \boldsymbol{\Xi} | \mathbf{x}_{1:T_{tr}})$ of the parameters can be obtained, the integral (\ref{eq:online_likelihood_integral_train}) is further approximated by the Monte Carlo method:
\begin{multline}
\iiint p(\mathbf{x}_{t+1} | \mathbf{x}_{1:t}, \boldsymbol{\Phi}, \boldsymbol{\Theta}, \boldsymbol{\Xi}) p(\boldsymbol{\Phi}, \boldsymbol{\Theta}, \boldsymbol{\Xi} | \mathbf{x}_{1:T_{tr}}) \mathrm{d}\boldsymbol\Phi \mathrm{d}\boldsymbol\Theta \mathrm{d} \boldsymbol{\Xi} \approx\\
\dfrac{1}{S} \sum\limits_{s = 1}^{S} p(\mathbf{x}_{t+1} | \mathbf{x}_{1:t}, \boldsymbol{\Phi}^{s}, \boldsymbol{\Theta}^{s}, \boldsymbol{\Xi}^{s}),
\end{multline}
where $S$ is the number of samples. These samples can be obtained (i) from the approximated posterior distributions $q(\boldsymbol{\Phi})$, $q(\boldsymbol{\Theta})$, and $q(\boldsymbol{\Xi})$ of the parameters, computed by the VB learning algorithm, or (ii) from the independent samples of the GS scheme. For the conditional likelihood $p(\mathbf{x}_{t+1} | \mathbf{x}_{1:t}, \boldsymbol{\Phi}^{s}, \boldsymbol{\Theta}^{s}, \boldsymbol{\Xi}^{s})$ the formula~(\ref{eq:online_likelihood_final}) is valid.

Note that for the approximated posterior distribution of the parameters, i.e., the output of the VB learning algorithm, the integral (\ref{eq:online_likelihood_integral_train}) can be resolved analytically, but it would be computationally infeasible. This is the reason why the Monte Carlo approximation is used in this case.

Finally, in order to compare documents of different lengths the normalised likelihood is used as a normality measure $s$:
\begin{equation}
s(\mathbf{x}_{t+1}) = \dfrac{1}{N_{t+1}} p(\mathbf{x}_{t+1} | \mathbf{x}_{1:t}).
\end{equation}

\subsection{Localisation of anomalies}
\label{sec:localisation}
The topic modeling approach allows to compute a likelihood function not only of the whole document but of an individual word within the document too. Recall that the visual word contains the information about a location in the frame. We propose to use the location information from the least probable words (e.g., 10 words with the least likelihood values) to localise anomalies in the frame. Note, we do not require anything additional to a topic model, e.g., modelling regional information explicitly as in~\cite{Haines2010} or comparing a test document with training ones as in~\cite{Pathak2015}. Instead, the proposed anomaly localisation procedure is general and can be applied in any topic modeling based method, where spatial information is encoded to visual words. 

The marginal likelihood of a word can be computed in a similar way to the likelihood of the whole document. For the point estimates of the parameters and plug-in approximation of the integral it is:
\begin{equation}
p(x_{i, t+1} | \mathbf{x}_{1:t}) \approx
p(x_{i, t+1} | \mathbf{x}_{1:t}, \hat{\boldsymbol{\Phi}}, \hat{\boldsymbol{\Theta}}, \hat{\boldsymbol{\Xi}}).
\end{equation}
For the samples from the posterior distributions of the parameters and the Monte Carlo integral approximation it is:
\begin{equation}
p(x_{i, t+1} | \mathbf{x}_{1:t}) \approx
\dfrac{1}{S} \sum\limits_{s = 1}^{S} p(x_{i, t+1} | \mathbf{x}_{1:t}, \boldsymbol{\Phi}^{s}, \boldsymbol{\Theta}^{s}, \boldsymbol{\Xi}^{s}).
\end{equation}

\section{Performance validation}
\label{sec:experiments}

We compare the two proposed learning algorithms, based on EM and VB, with the GS algorithm, proposed in~\cite{Hospedales2011}, on two real datasets.  

\begin{figure}[!t]
\centering
\subfloat[]{\includegraphics[width=0.4\columnwidth]{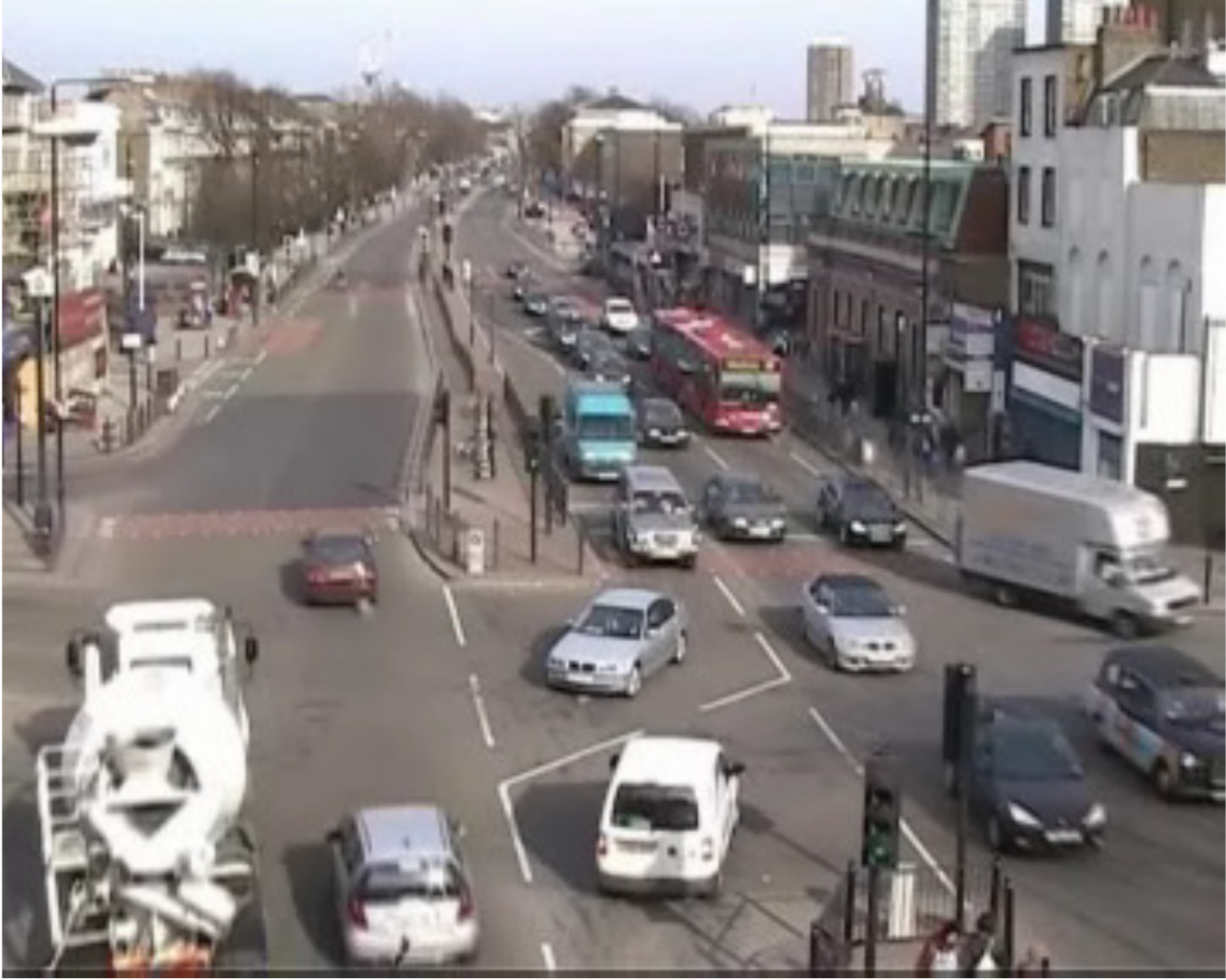}}
\hfil
\subfloat[]{\includegraphics[width=0.4\columnwidth]{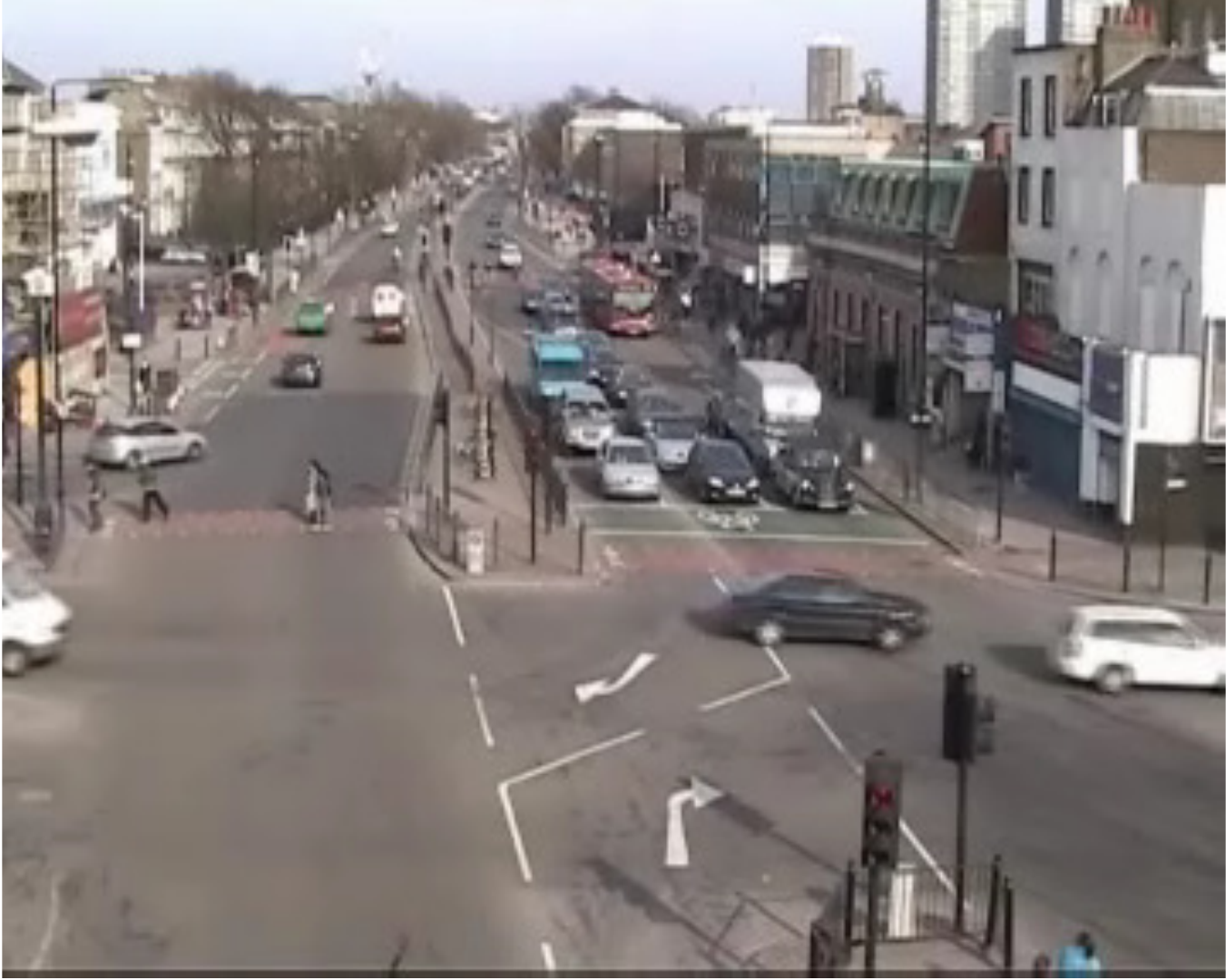}}
\hfil
\subfloat[]{\includegraphics[width=0.4\columnwidth]{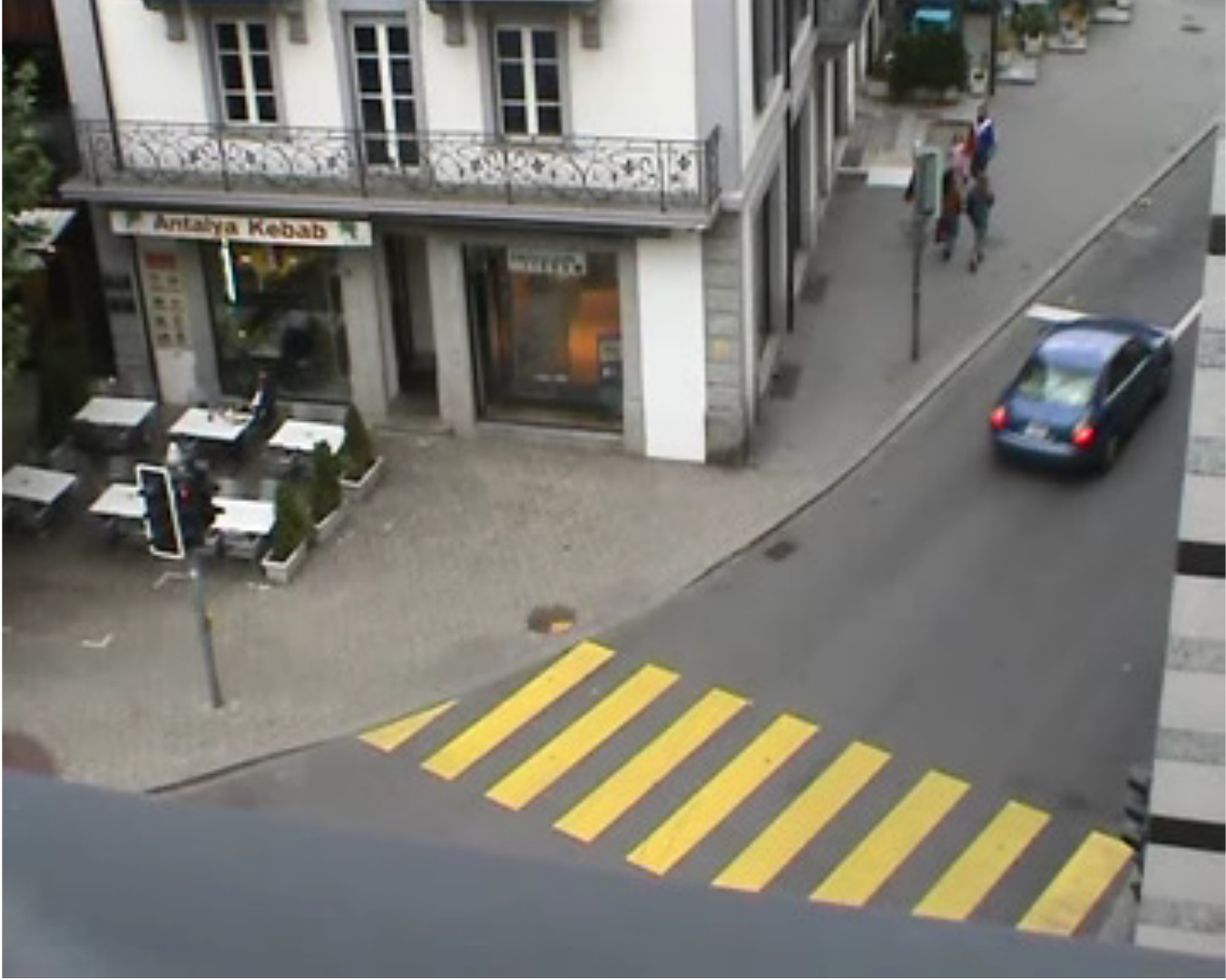}}
\hfil
\subfloat[]{\includegraphics[width=0.4\columnwidth]{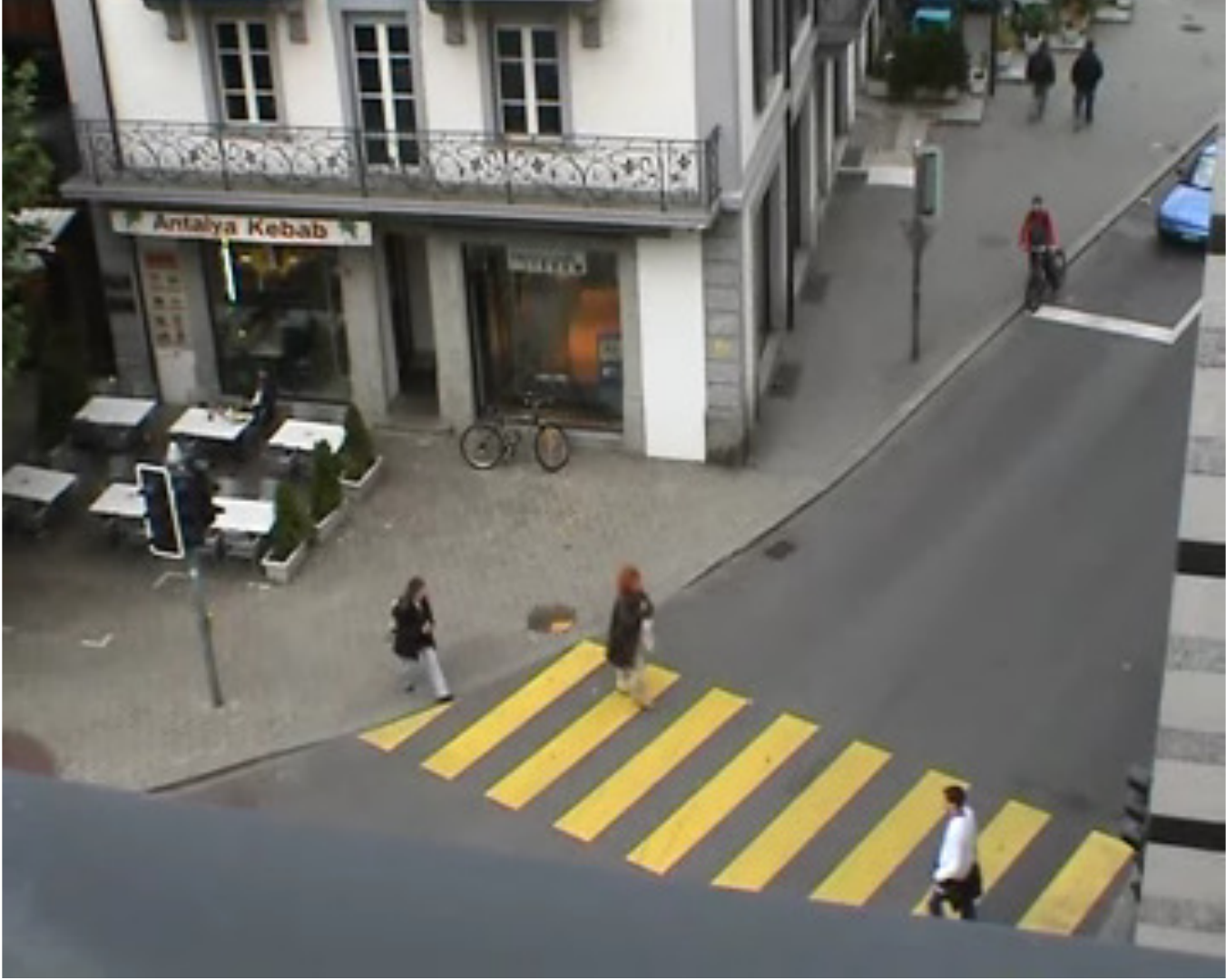}}
\caption{Sample frames of the real datasets. The top row presents two sample frames from the QMUL data, the bottom row presents two sample frames from the Idiap data.}
\label{fig:sample_frames}
\end{figure}

\subsection{Setup}
The performance of the algorithms is compared on the QMUL street intersection data~\cite{Hospedales2011} and Idiap traffic junction data \cite{Varadarajan2009}. Both datasets are $45$-minutes video sequences, captured busy traffic road junctions, where we use a $5$-minute video sequence as a training dataset and others as a testing one. The documents that have less than $20$ visual words are discarded from consideration. In practice these documents can be classified to be normal by default as there is no enough information to make a decision. The frame size for both datasets is $288 \times 360$. Sample frames are presented in Figure~\ref{fig:sample_frames}. 

The size of grid cells is set to $8 \times 8$ pixels for spatial quantisation of the local motion for visual word determination. Non-overlapping clips with a one second  length are treated as visual documents. 

\begin{figure}[!t]
\includegraphics[width = 0.95\columnwidth]{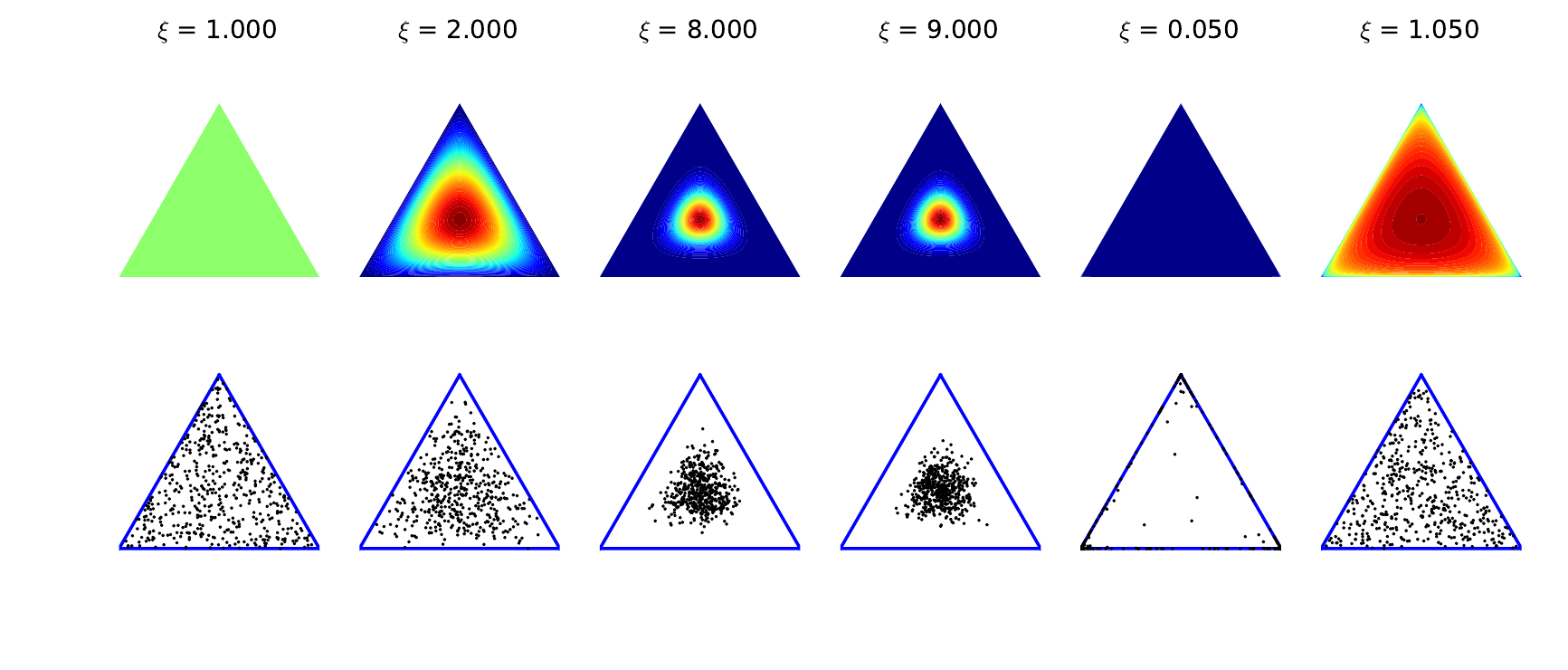}%
\caption{Dirichlet distributions with different symmetric parameters $\xi$. For the representation purposes the three-dimensional space is used. On the top row the colours correspond to the Dirichlet probability density function values in the area. On the bottom row there are samples generated from the corresponding density functions. The sample size is $500$.}
\label{fig:dirichlet_pdf}
\end{figure} 

\begin{figure*}[!t]
\centering
\subfloat[Behaviour 1]{\includegraphics[width=0.24\textwidth]{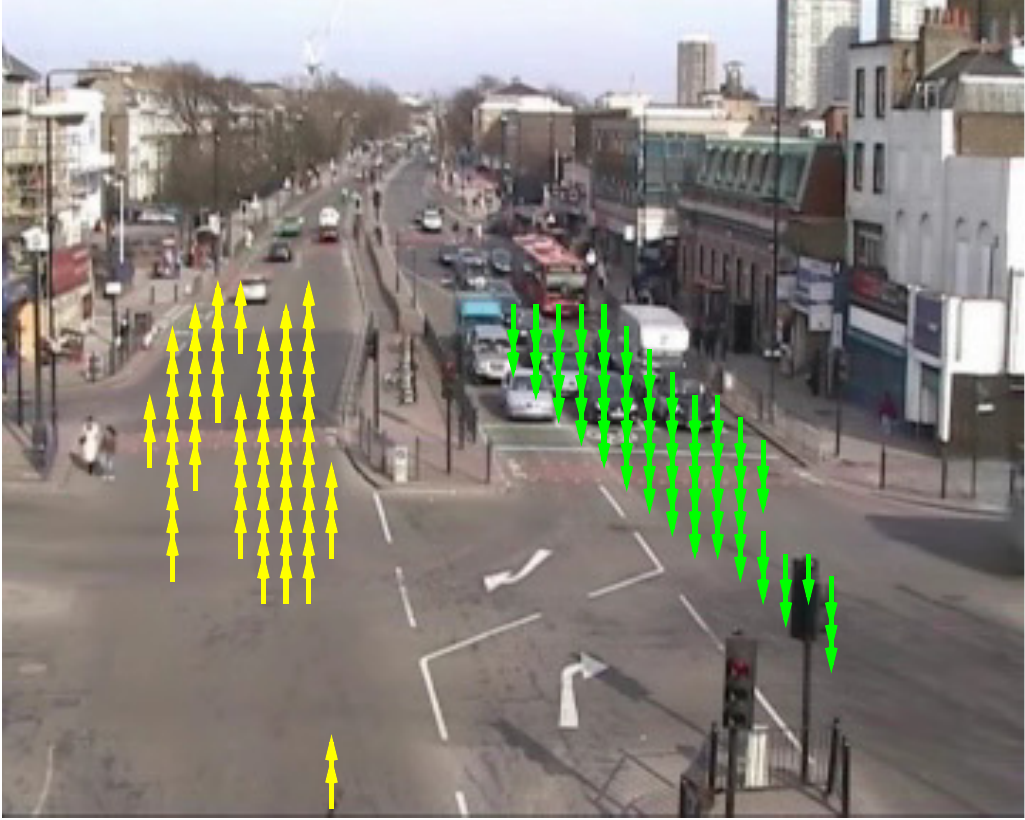}
\label{fig:qmul_behav_1}}%
\hfil
\subfloat[Behaviour 2]{\includegraphics[width=0.24\textwidth]{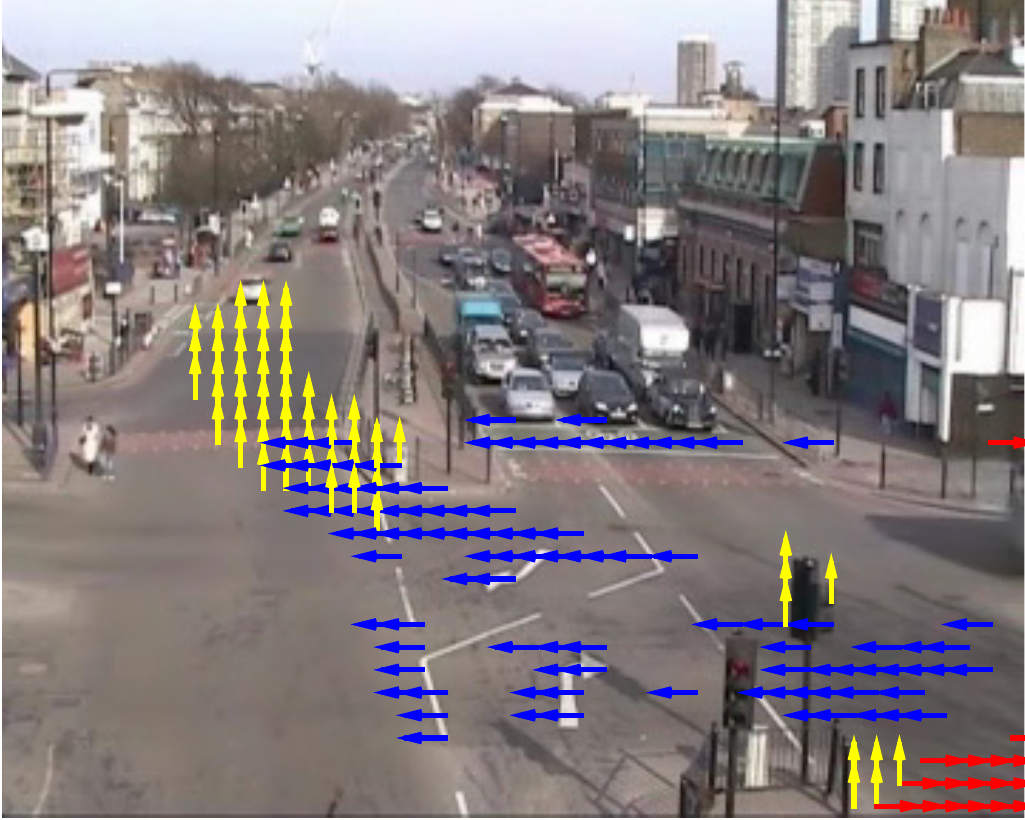}
\label{fig:qmul_behav_2}}%
\hfil
\subfloat[Behaviour 3]{\includegraphics[width=0.24\textwidth]{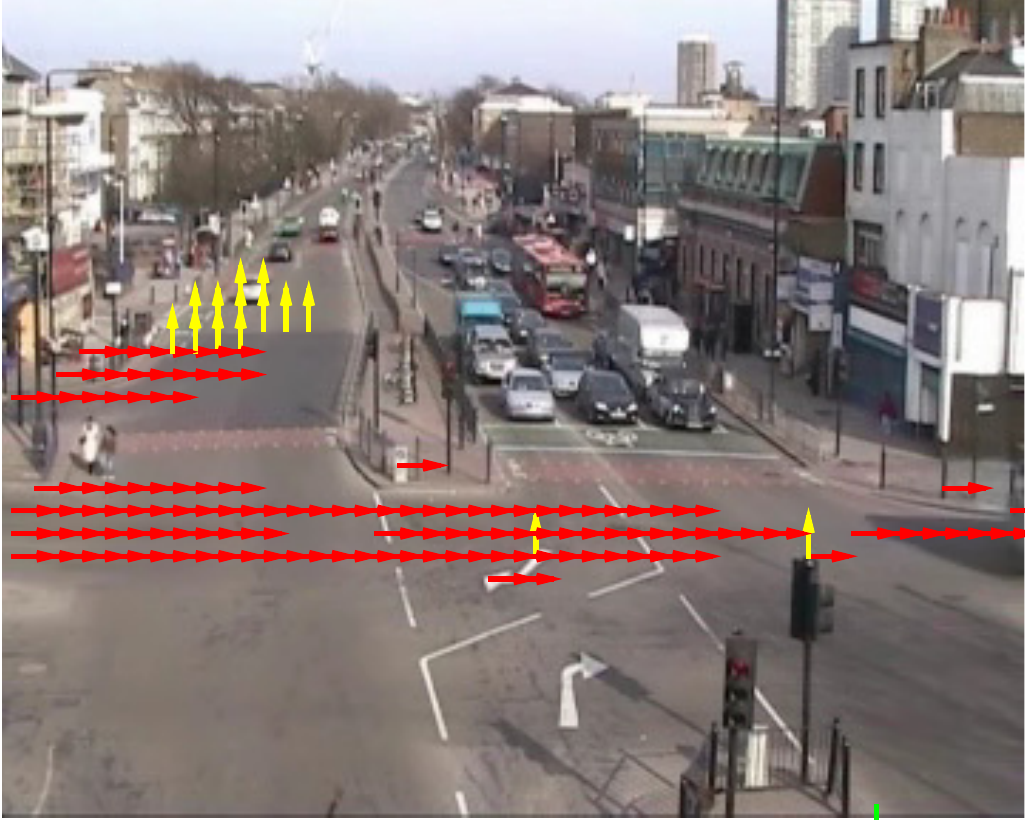}
\label{fig:qmul_behav_3}}%
\hfil
\subfloat[Behaviour 4]{\includegraphics[width=0.24\textwidth]{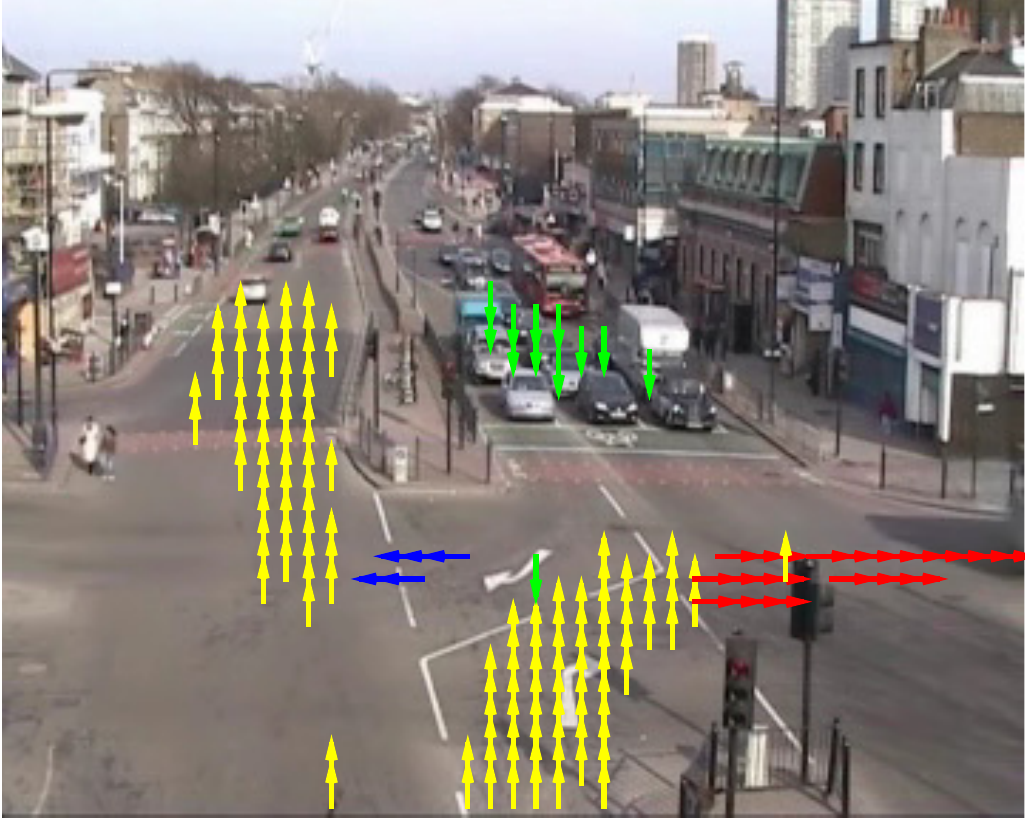}
\label{fig:qmul_behav_4}}%
\caption{Behaviours learnt by the EM learning algorithm for the QMUL data. The arrows represent the visual words: the location and direction of the motion. The first behaviour \protect\subref{fig:qmul_behav_1} corresponds to the vertical traffic flow, the second \protect\subref{fig:qmul_behav_2} and the third \protect\subref{fig:qmul_behav_3} behaviours correspond to the left and right traffic flow, respectively. The fourth \protect\subref{fig:qmul_behav_4} behaviour correspond to turns that follow the vertical traffic flow.}
\label{fig:qmul_behaviours}
\end{figure*} 

\begin{figure*}[!t]
\centering
\subfloat[Behaviour 1]{\includegraphics[width=0.24\textwidth]{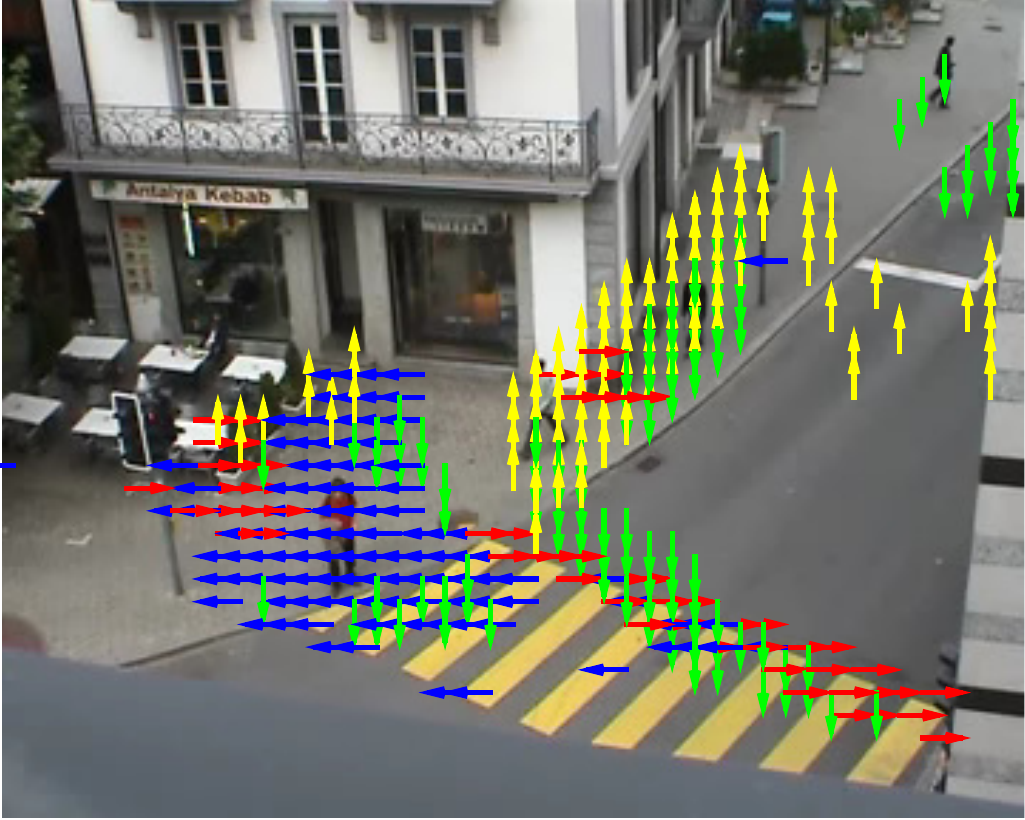}
\label{fig:idiap_behav_1}}%
\hfil
\subfloat[Behaviour 2]{\includegraphics[width=0.24\textwidth]{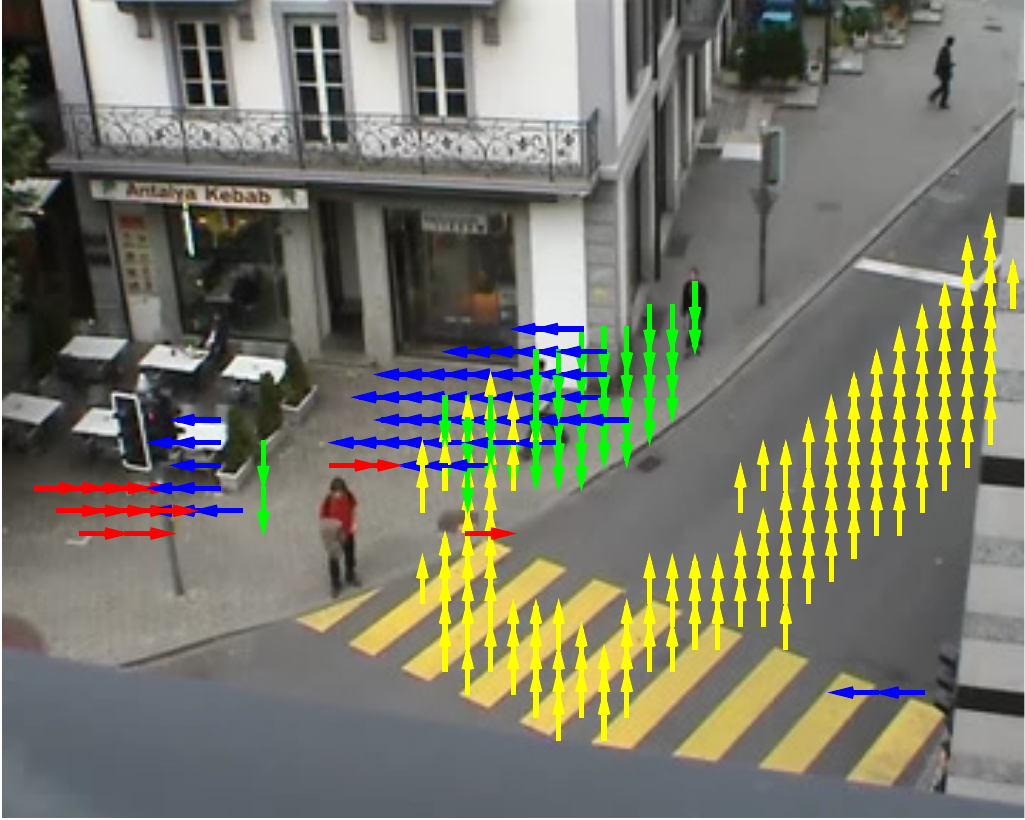}
\label{fig:idiap_behav_2}}%
\hfil
\subfloat[Behaviour 3]{\includegraphics[width=0.24\textwidth]{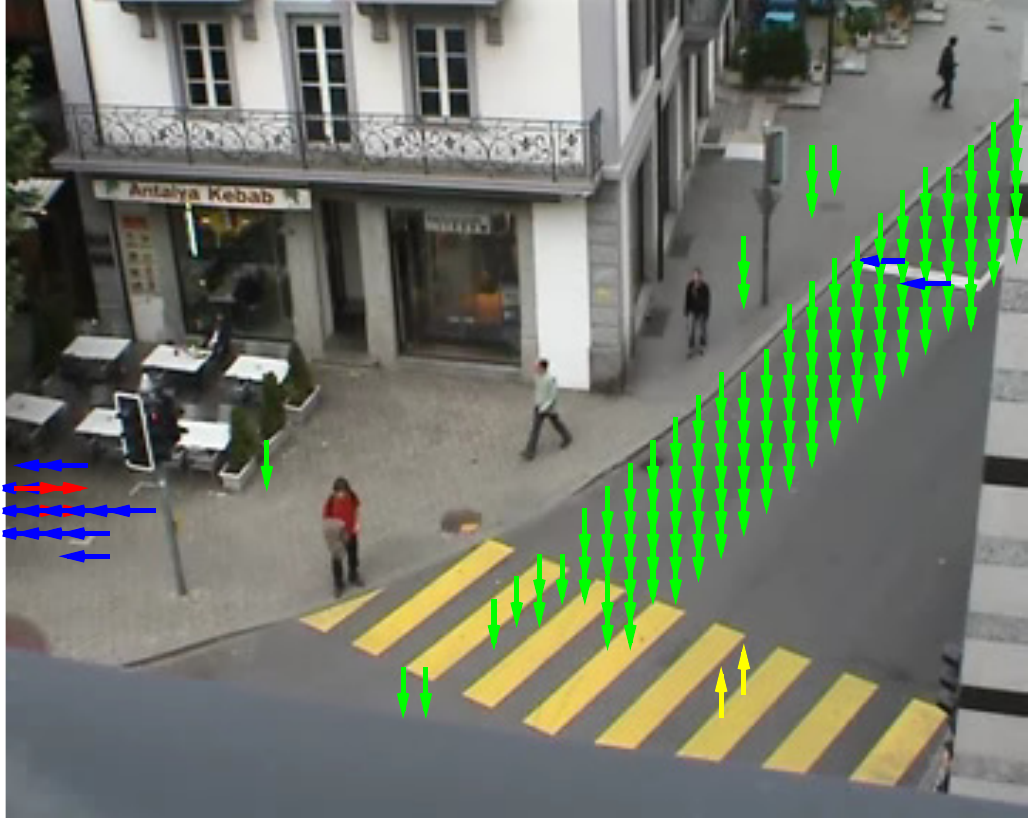}
\label{fig:idiap_behav_3}}%
\caption{Behaviours learnt by the EM learning algorithm for the Idiap data. The arrows represent the visual words: the location and direction of the motion. The first behaviour \protect\subref{fig:idiap_behav_1} corresponds to the pedestrian motion, the second \protect\subref{fig:idiap_behav_2} and the third \protect\subref{fig:idiap_behav_3} behaviours correspond to the upward and downward traffic flows, respectively.}
\label{fig:idiap_behaviours}
\end{figure*} 

We also study the influence of the hyperparameters on the learning algorithms. In all the experiments we use the symmetric hyperparameters: $\boldsymbol\alpha = \{\alpha, \dotsc, \alpha\}$, $\boldsymbol\beta = \{\beta, \dotsc, \beta\}$, $\boldsymbol\gamma = \{\gamma, \dotsc, \gamma\}$ and $\boldsymbol\eta = \{\eta, \dotsc, \eta\}$. The three groups of the hyperparameters settings are compared: $\{\alpha = 1, \beta = 1, \gamma = 1, \eta = 1\}$ (referred as ``prior type 1''), $\{\alpha = 8 , \beta = 0.05, \gamma = 1, \eta = 1\}$ (``prior type H'') and $\{\alpha = 9, \beta = 1.05, \gamma = 2, \eta = 2\}$ (``prior type H+1''). Note that the first group corresponds to the case when in the EM-algorithm learning scheme the prior components are cancelled out, i.e., the MAP estimates in this case are equal to the maximum likelihood ones. The equations for the point estimates in the EM learning algorithm with the prior type H+1 of the hyperparameters settings are equal to the equations for the point estimates in the VB and GS learning algorithms with the prior type H of the settings. The corresponding Dirichlet distributions with all used parameters are presented in Figure~\ref{fig:dirichlet_pdf}. 

Note that parameter learning is an ill-posed problem in topic modeling~\cite{Vorontsov2014ARTMArticle}. This means there is no unique solution for parameter estimates. We use $20$ Monte Carlo runs for all the learning algorithms with different random initialisations resulting with different solutions. The mean results among these runs are presented below for comparison. 

All three algorithms are run with three different groups of hyperparameters settings. The number of topics and behaviours is set to $8$ and $4$, respectively, for the QMUL dataset, $10$ and $3$ are used for the corresponding values for the Idiap dataset. The EM and VB algorithms are run for $100$ iterations. The GS algorithm is run for $500$ burn-in iterations and independent samples are taken with a $100$ iterations delay after the burn-in period. 

\subsection{Performance measure}

Anomaly detection performance of the algorithms depends on threshold selection. To make a fair comparison of the different learning algorithms we use a performance measure, which is independent of threshold selection.

In binary classification the following measures~\cite{Murphy2012} are used: $\text{TP}$ --- true positive, a number of documents, which are correctly detected as positive (abnormal in our case); $\text{TN}$ --- true negative, a number of documents, which are correctly detected as negative (normal in our case); $\text{FP}$ --- false positive, a number of documents, which are incorrectly detected as positive, when they are negative; $\text{FN}$ --- false negative, a number of documents, which are incorrectly detected as negative, when they are positive; $\text{precision} = \dfrac{\text{TP}}{\text{TP} + \text{FP}}$ --- a fraction of correct detections among all documents labelled as abnormal by an algorithm; $\text{recall} = \dfrac{\text{TP}}{\text{TP} + \text{FN}}$ --- a fraction of correct detections among all truly abnormal documents.
   
The area under the precision-recall curve is used as a performance measure in this paper. This measure is more informative for detection of rare events than the popular area under the receiver operating characteristic (ROC) curve~\cite{Murphy2012}.

\subsection{Parameter learning}
We visualise the learnt behaviours for the qualitative assessment of the proposed framework (Figures~\ref{fig:qmul_behaviours} and~\ref{fig:idiap_behaviours}). For illustrative purposes we consider one run of the EM learning algorithm with the prior type H+1 of the hyperparameters settings. 

The behaviours learnt for the QMUL data are shown in Figure~\ref{fig:qmul_behaviours} (for visualisation words representing $50\%$ of probability mass of a behaviour are used). One can notice that the algorithm correctly recognises the motion patterns in the data. The general motion of the scene follows a cycle: a vertical traffic flow (the first behaviour in Figure~\ref{fig:qmul_behav_1}), when cars move downward and upward on the road; left and right turns (the fourth behaviour in Figure~\ref{fig:qmul_behav_4}): some cars moving on the ``vertical'' road turn to the perpendicular road at the end of the vertical traffic flow; a left traffic flow (the second behaviour in Figure~\ref{fig:qmul_behav_2}), when cars move from right to left on the ``horizontal'' road; and a right traffic flow (the third behaviour in Figure~\ref{fig:qmul_behav_3}), when cars move from left to right on the ``horizontal'' road. Note that the ordering numbers of behaviours correspond to their internal representation in the algorithm. The transition probability matrix $\boldsymbol{\Xi}$ is used to recognise the correct behaviours order in the data.    

Figure~\ref{fig:idiap_behaviours} presents the behaviours learnt for the Idiap data. In this case the learnt behaviours have also a clear semantic meaning. The scene motion follows a cycle: a pedestrian flow (the first behaviour in Figure \ref{fig:idiap_behav_1}), when cars stop in front of the stop line and pedestrians cross the road; a downward traffic flow (the third behaviour in Figure \ref{fig:idiap_behav_3}), when cars move downward along the road; an upward traffic flow (the second behaviour in Figure \ref{fig:idiap_behav_2}), when cars from left and right sides move upward on the road.  

\subsection{Anomaly detection}
\begin{figure}[!t]
\centering
\subfloat[Car moving on the opposite lane]{\includegraphics[width=0.4\columnwidth]{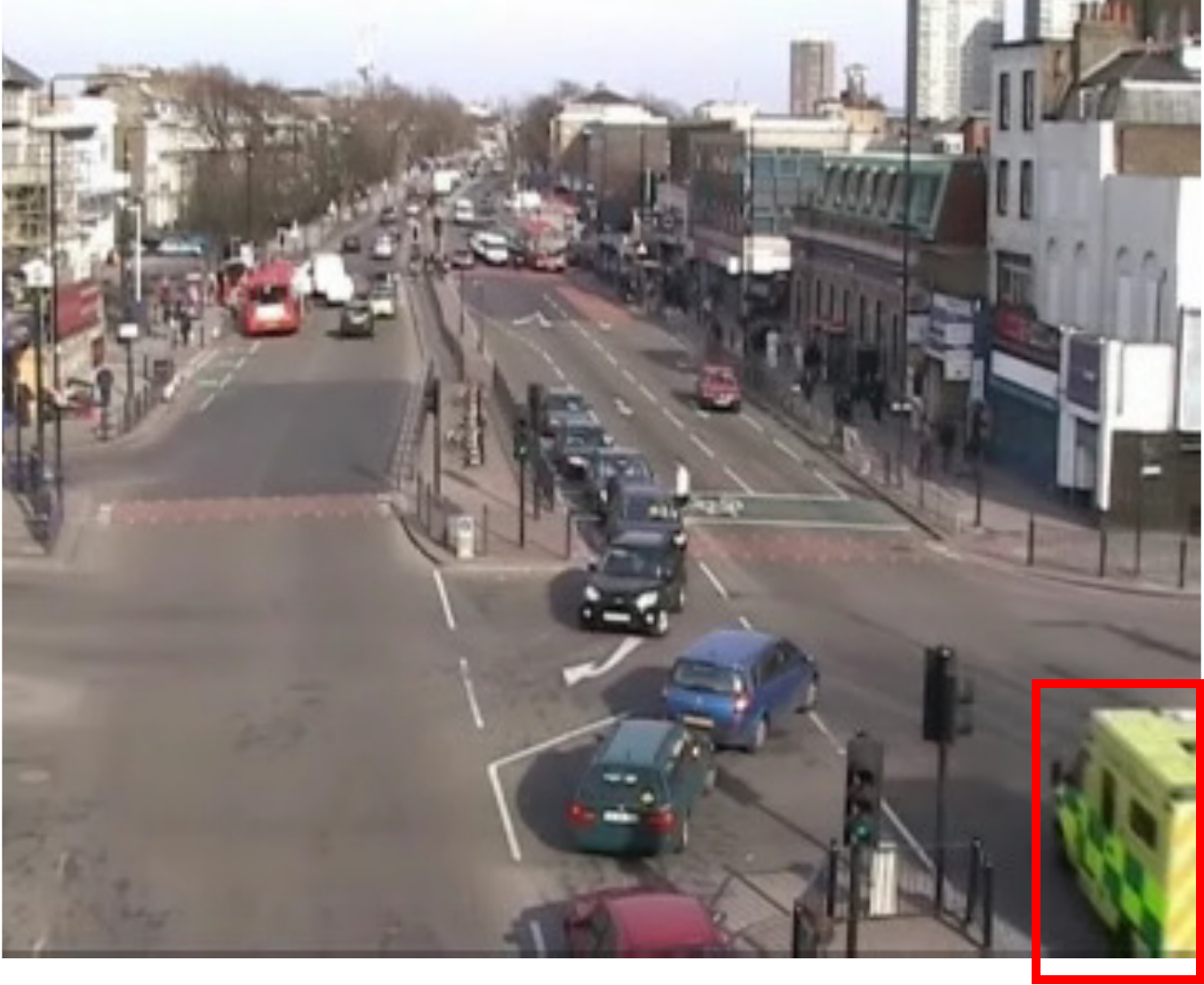}}
\hfil
\subfloat[Disruption of the traffic flow]{\includegraphics[width=0.4\columnwidth]{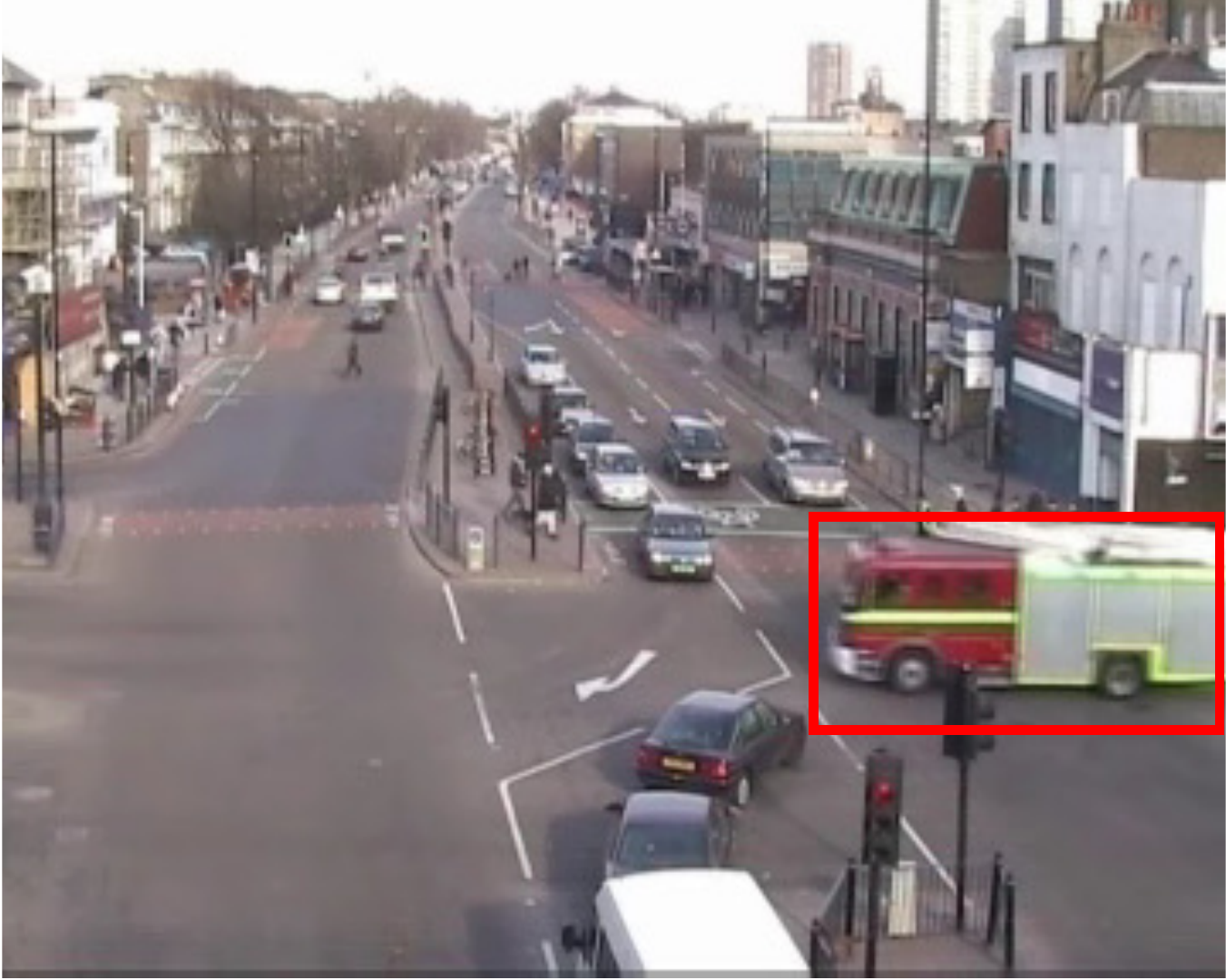}}
\hfil
\subfloat[Jaywalking]{\includegraphics[width=0.4\columnwidth]{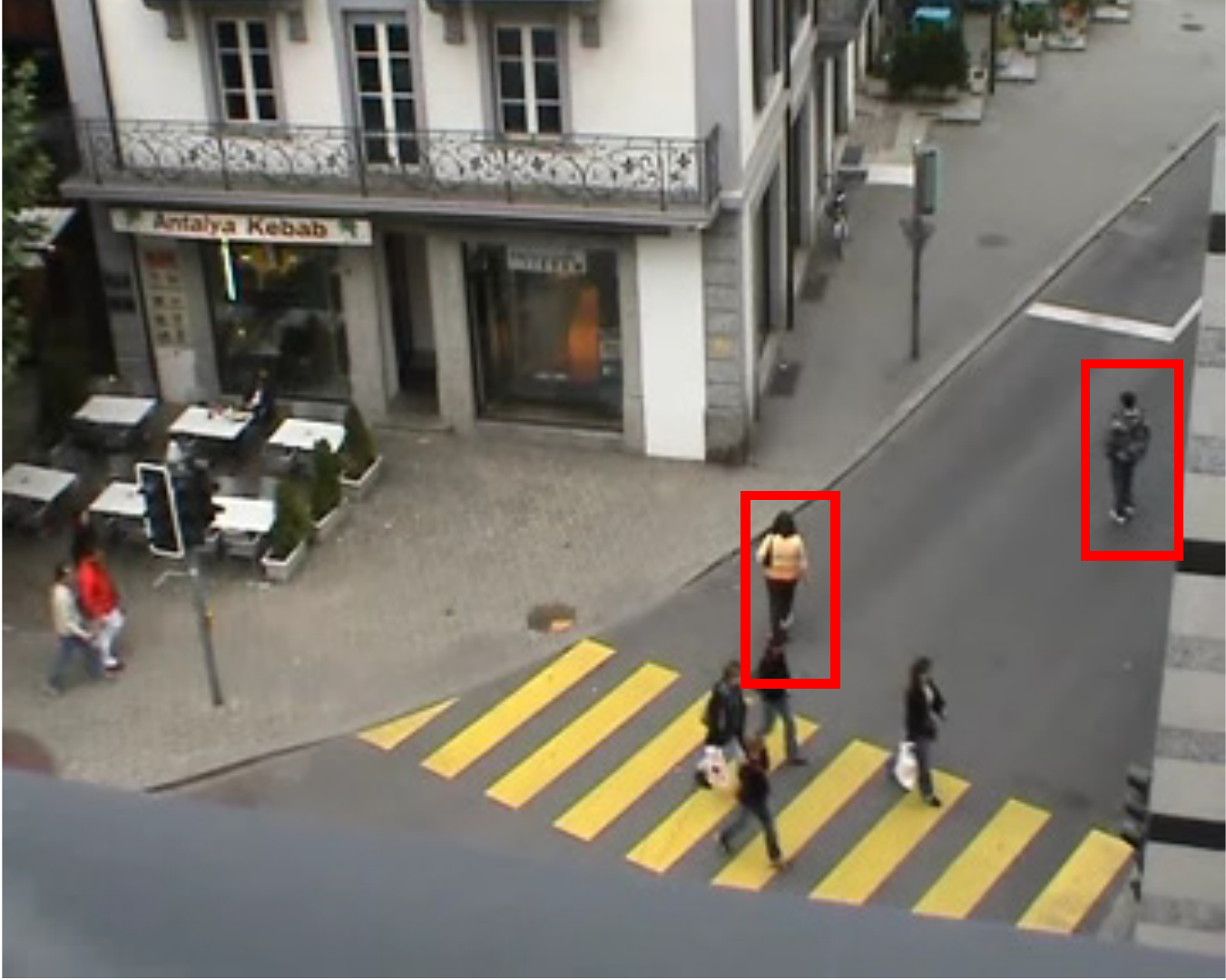}}
\hfil
\subfloat[Car moving on the sidewalk]{\includegraphics[width=0.4\columnwidth]{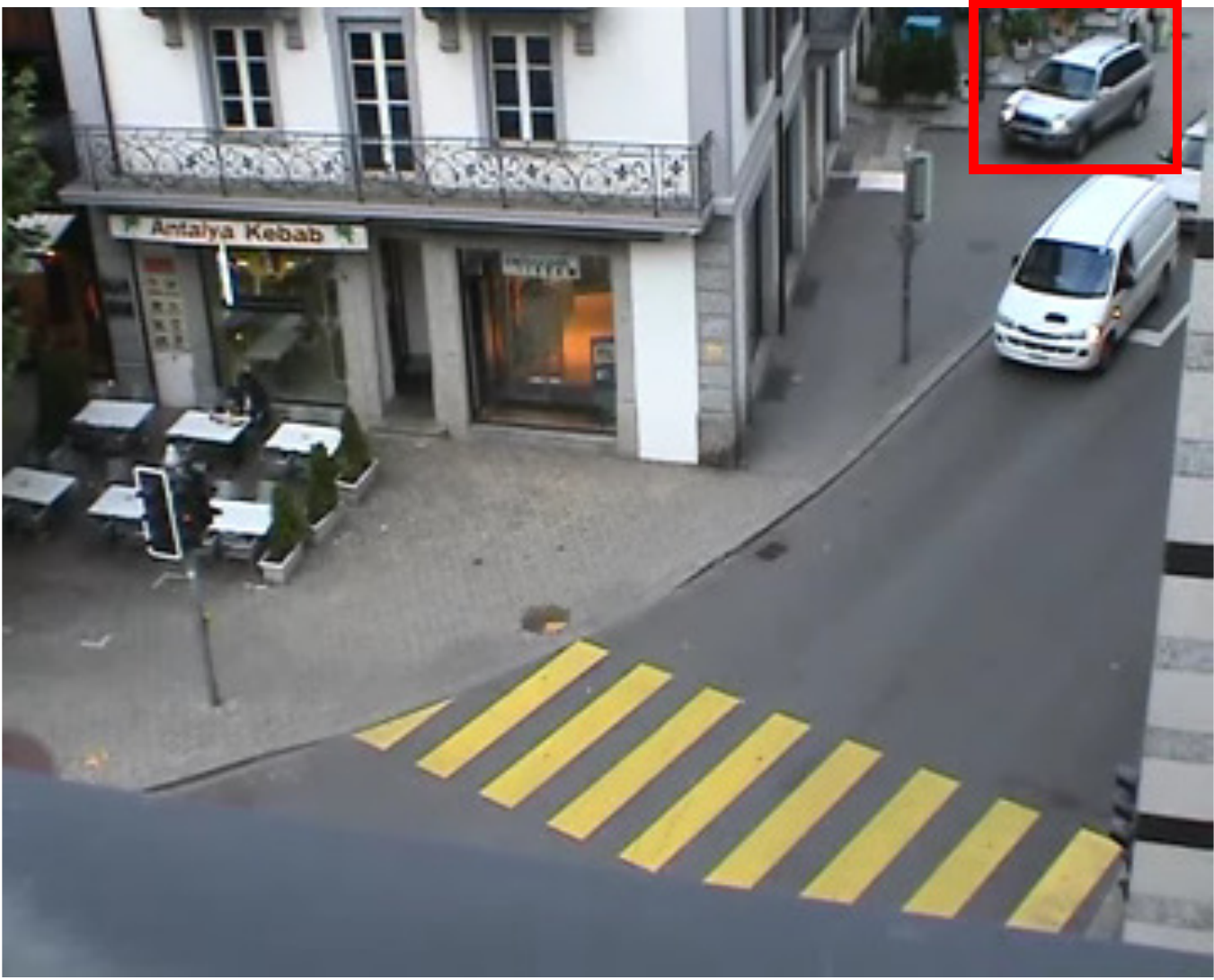}}
\caption{Examples of abnormal events}
\label{fig:sample_abnormalities}
\end{figure}

In this section the anomaly detection performance achieved by all three learning algorithms is compared. The datasets contain the number of abnormal events, such as jaywalking, car moving on the opposite lane, disruption of the traffic flow (see examples in Figure~\ref{fig:sample_abnormalities}). 

For the EM learning algorithm the plug-in approximation of the marginal likelihood is used for anomaly detection. For both the VB and GS learning algorithms both the plug-in and Monte Carlo approximations of the likelihood are used. Note that for the GS algorithm samples are obtained during the learning stage, $5$ and $100$ independent samples are taken. For the VB learning algorithm samples are obtained after the learning stage from the posterior distributions, parameters of which are learnt. This means that the number of samples that are used for anomaly detection does not influence on the computational cost of learning. We test the Monte Carlo approximation of the marginal likelihood with $5$ and $100$ samples for the VB learning algorithm. 

As a result, we have $21$ methods to compare: obtained by three learning algorithms, three different groups of hyperparameters settings, one type of marginal likelihood approximation for the EM learning algorithm, two types of marginal likelihood approximation for the VB and GS learning algorithms, where two Monte Carlo approximations are used with $5$ and $100$ samples. The list of methods references can be found in Table~\ref{tab:methods_references}. 

Note that we achieve a very fast decision making performance in our framework. Indeed, anomaly detection is made for approximately $0.0044$ sec per visual document by the plug-in approximation of the marginal likelihood, for $0.0177$ sec per document by the Monte Carlo approximation with $5$ samples and for $0.3331$ sec per document by the Monte Carlo approximation with $100$ samples\footnote{The computational time is provided for a laptop computer with i7-4702HQ CPU with 2.20GHz, 16 GB RAM using Matlab R2015a implementation.}.

\begin{table}[t]
  \caption{Methods references}
  \label{tab:methods_references}
  \centering
  \begin{tabular}{p{2cm} p{0.9cm} p{1.2cm} p{1.6cm} p{0.8cm}}
    \toprule
    Reference     & Learning algorithm     & Hyper-parameters settings & Marginal likelihood approximation & Number of posterior samples \\
    \midrule
    EM 1 p & EM & type 1 & Plug-in & ---\\
	EM H p & EM & type H & Plug-in & ---\\
	EM H+1 p & EM & type H+1 & Plug-in & ---\\
	\midrule
	VB 1 p & VB & type 1 & Plug-in & ---\\
	VB 1 mc 5 & VB & type 1 & Monte Carlo & 5\\
	VB 1 mc 100 & VB & type 1 & Monte Carlo & 100\\
	VB H p & VB & type H & Plug-in & ---\\
	VB H mc 5 & VB & type H & Monte Carlo & 5\\
	VB H mc 100 & VB & type H & Monte Carlo & 100\\
	VB H+1 p & VB & type H+1 & Plug-in & ---\\
	VB H+1 mc 5 & VB & type H+1 & Monte Carlo & 5\\
	VB H+1 mc 100 & VB & type H+1 & Monte Carlo & 100\\
	\midrule
	GS 1 p & GS & type 1 & Plug-in & ---\\
	GS 1 mc 5 & GS & type 1 & Monte Carlo & 5\\
	GS 1 mc 100 & GS & type 1 & Monte Carlo & 100\\	
	GS H p & GS & type H & Plug-in & ---\\
	GS H mc 5 & GS & type H & Monte Carlo & 5\\
	GS H mc 100 & GS & type H & Monte Carlo & 100\\
	GS H+1 p & GS & type H+1 & Plug-in & ---\\
	GS H+1 mc 5 & GS & type H+1 & Monte Carlo & 5\\
    GS H+1 mc 100 & GS & type H+1 & Monte Carlo & 100\\
    \bottomrule
  \end{tabular}
\end{table}

\begin{figure}[!t]
	\centering
	\subfloat[QMUL data results]{
		\includegraphics{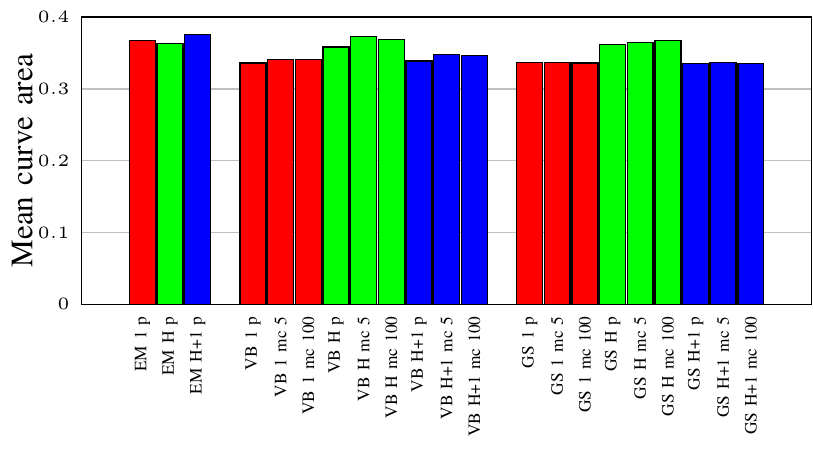}
		\label{fig:qmul_results}
	}%
	\hfil
	\subfloat[Idiap data results]{
		\includegraphics{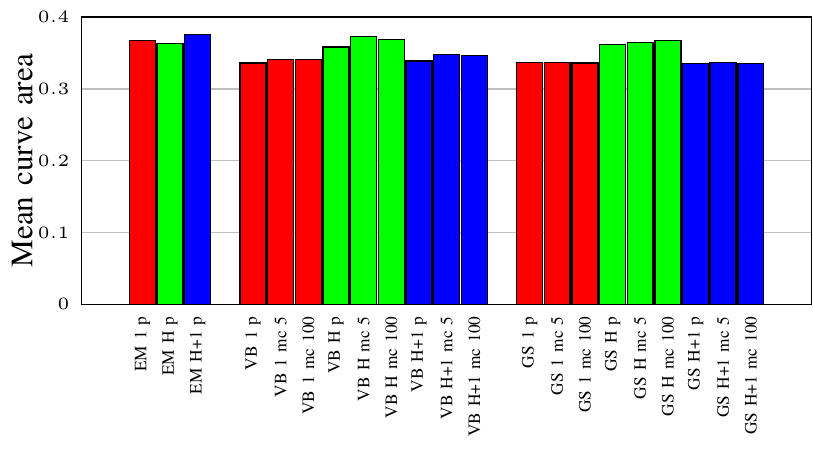}
		\label{fig:idiap_results}
	}%
	\caption{Results of anomaly detection. \protect\subref{fig:qmul_results} are the mean areas under precision-recall curves for the QMUL data. \protect\subref{fig:idiap_results} are the mean areas under precision-recall curves for the Idiap data.} 
	\label{fig:mean_results}
\end{figure}

The mean areas under precision-recall curves for anomaly detection for all $21$ compared methods can be found in Figure~\ref{fig:mean_results}. Below we examine the results with respect to hyperparameters sensitivity, an influence of the likelihood approximation on the final performance, we also compare the learning algorithms and discuss anomaly localisation results.   

\begin{figure}[!t]
\centering
\subfloat[VB 1 p curves]{\includegraphics[width=0.485\columnwidth]{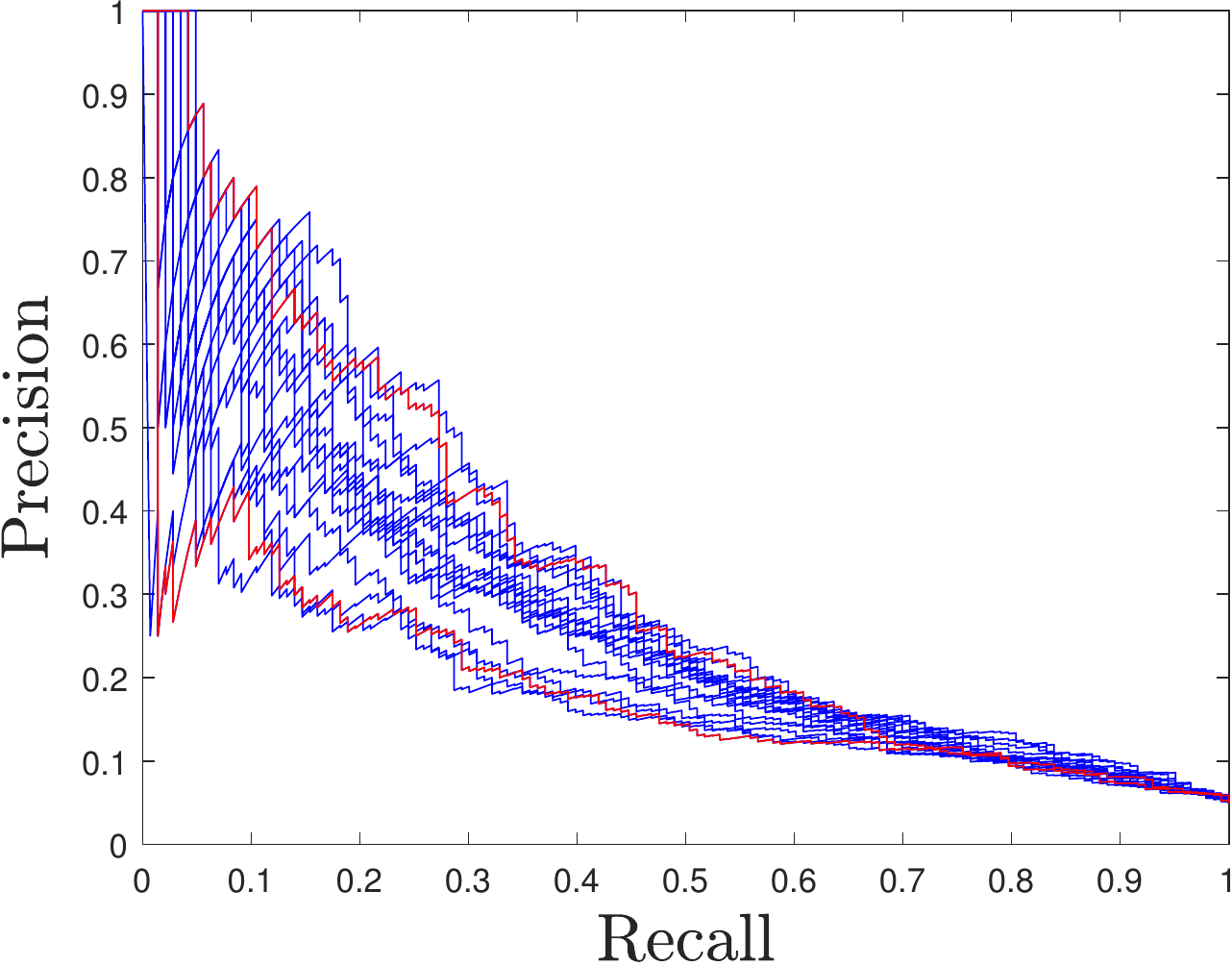}
\label{fig:hyperparam_sen_vb_1}}%
\hfil
\subfloat[VB H p curves]{\includegraphics[width=0.485\columnwidth]{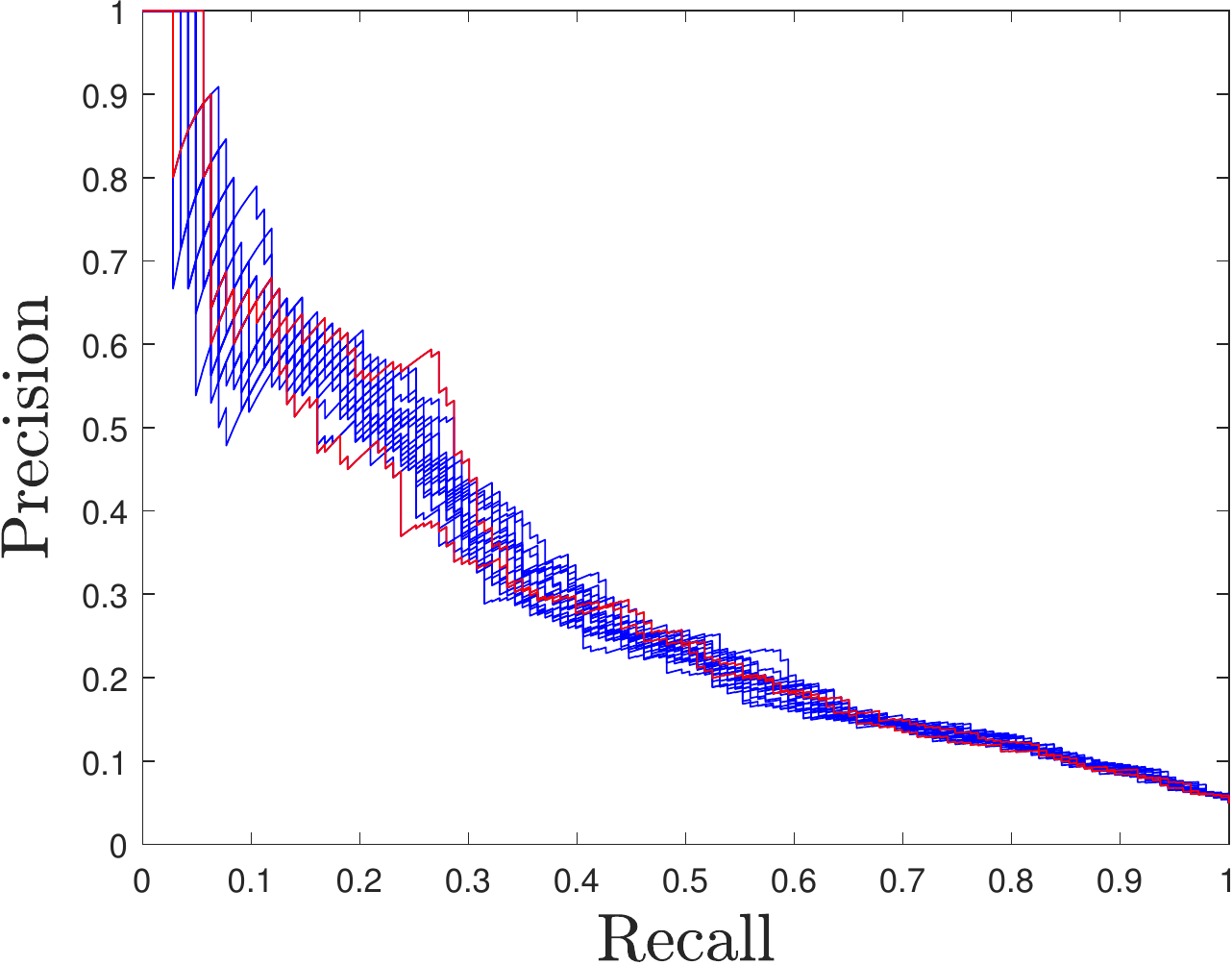}
\label{fig:hyperparam_sen_vb_2}}%
\hfil
\subfloat[EM 1 p curves]{\includegraphics[width=0.485\columnwidth]{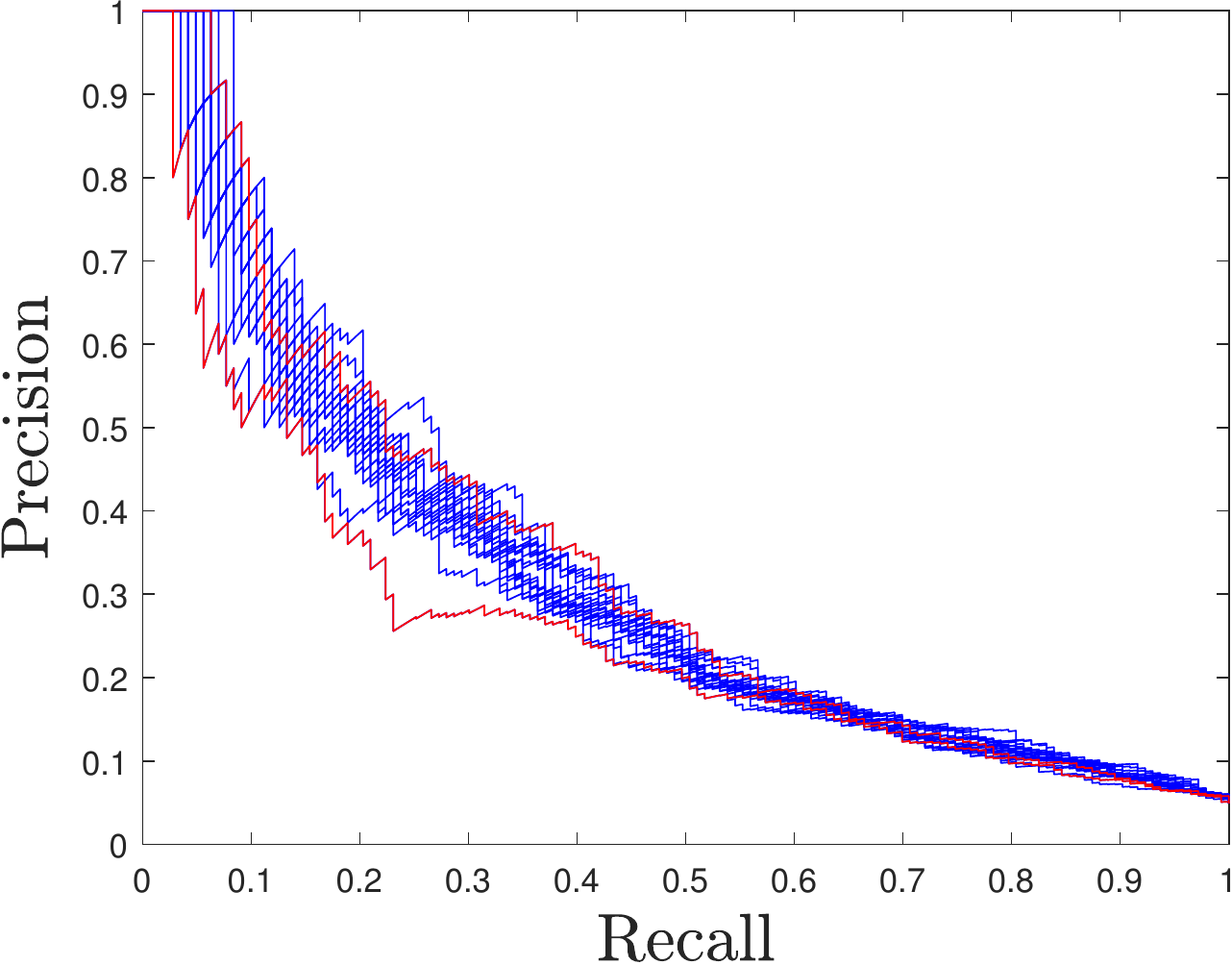}
\label{fig:hyperparam_sen_em_1}}%
\hfil
\subfloat[EM H p curves]{\includegraphics[width=0.485\columnwidth]{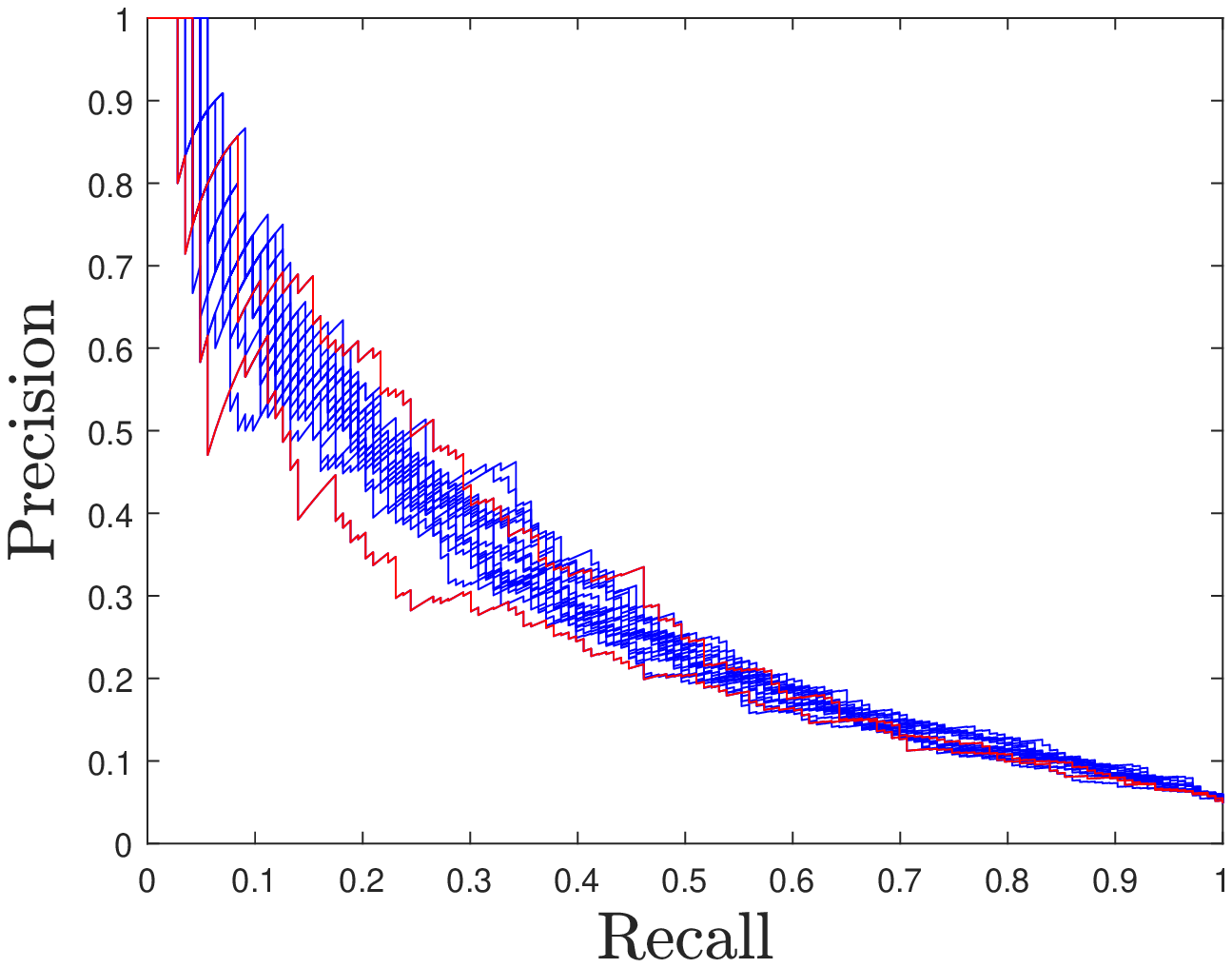}
\label{fig:hyperparam_sen_em_2}}%
\caption{Hyperparameters sensitivity of the precision-recall curves. The top row corresponds to all the independent runs of the VB learning algorithm with the prior type $1$ \protect\subref{fig:hyperparam_sen_vb_1} and the prior type H \protect\subref{fig:hyperparam_sen_vb_2}. The bottom row corresponds to all the independent runs of the EM learning algorithm with the prior type $1$ \protect\subref{fig:hyperparam_sen_em_1} and the prior type H \protect\subref{fig:hyperparam_sen_em_2}. The red colour highlights the curves with the maximum and minimum areas under curves.}
\label{fig:hyperparam_sensitivity}
\end{figure} 

\subsubsection{Hyperparameters sensitivity} 
This section presents sensitivity analysis of the anomaly detection methods with respect to changes of the hyperparameters.

The analysis of the mean areas under curves (Figure \ref{fig:mean_results}) suggests that the  hyperparameters almost do not influence on the results of the EM learning algorithm, while there is a significant dependence between hyperparameters changes and results of the VB and GS learning algorithms. These conclusions are confirmed by examination of the individual runs of the algorithms. For example, Figure \ref{fig:hyperparam_sensitivity} presents the precision-recall curves for all~$20$ runs with different initialisations of~$4$ methods for the QMUL data: the VB learning algorithm using the plug-in approximation of the marginal likelihood with the prior types~1 and~H of the hyperparameters settings and the EM learning algorithm with the same prior groups of the hyperparameters settings. One can notice that the variance of the curves for the VB learning algorithm with the prior type~1 is larger than the corresponding variance with the prior type H, while the similar variances for the EM learning algorithm are very close to each other. 
 
Note that the results of the EM learning algorithm with the prior type~1 do not significantly differ from the results with the other priors, despite of the fact that the prior type~1 actually cancels out the prior influence on the parameters estimates and equates the MAP and maximum likelihood estimates. We can conclude that the choice of the hyperparameters settings is not a problem for the EM learning algorithm and we can even simplify the derivations considering only the maximum likelihood estimates without the prior influence. 

The VB and GS learning algorithms require a proper choice of the hyperparameters settings as they can significantly change the anomaly detection performance. This choice can be performed empirically or with the type II maximum likelihood approach~\cite{Murphy2012}.

\subsubsection{Marginal likelihood approximation influence}
In this section the influence of the type of the marginal likelihood approximation on the anomaly detection results is studied. 

The average results for both datasets (Figure \ref{fig:mean_results}) demonstrate that the type of the marginal likelihood approximation does not influence remarkably on anomaly detection performance. As the plug-in approximation requires less computational resources both in terms of time and memory (as there is no need to sample and store posterior samples and average among them) this type of approximation is recommended to be used for anomaly detection in the proposed framework.

\subsubsection{Learning algorithms comparison}
This section compares the anomaly detection performance obtained by three learning algorithms. 

The best results in terms of a mean area under a precision-recall curve are obtained by the EM learning algorithm, the worst results are obtained by the GS learning algorithm (Figure \ref{fig:mean_results} and Table \ref{tab:mean_area}). In Table \ref{tab:mean_area} for each learning algorithm the group of hyperparameters settings and the type of marginal likelihood approximation is chosen to have the maximum of the mean area under curves, where a mean is taken over independent runs of the same method and maximum is taken among different settings for the same learning algorithm. 

\begin{table}[!t]
\caption{Mean area under precision-recall curves}
\label{tab:mean_area}
\centering
\begin{tabular}{p{2cm} p{1cm} p{1cm} p{1cm}}
\toprule
Dataset & EM & VB & GS \\
\midrule
QMUL & 0.3166 & 0.3155 & 0.2970\\
Idiap & 0.3759 & 0.3729 & 0.3673\\
\bottomrule
\end{tabular}
\end{table} 

\begin{figure}[!t]
\centering
\subfloat[QMUL data --- best results]{\includegraphics[width = 0.485\columnwidth]{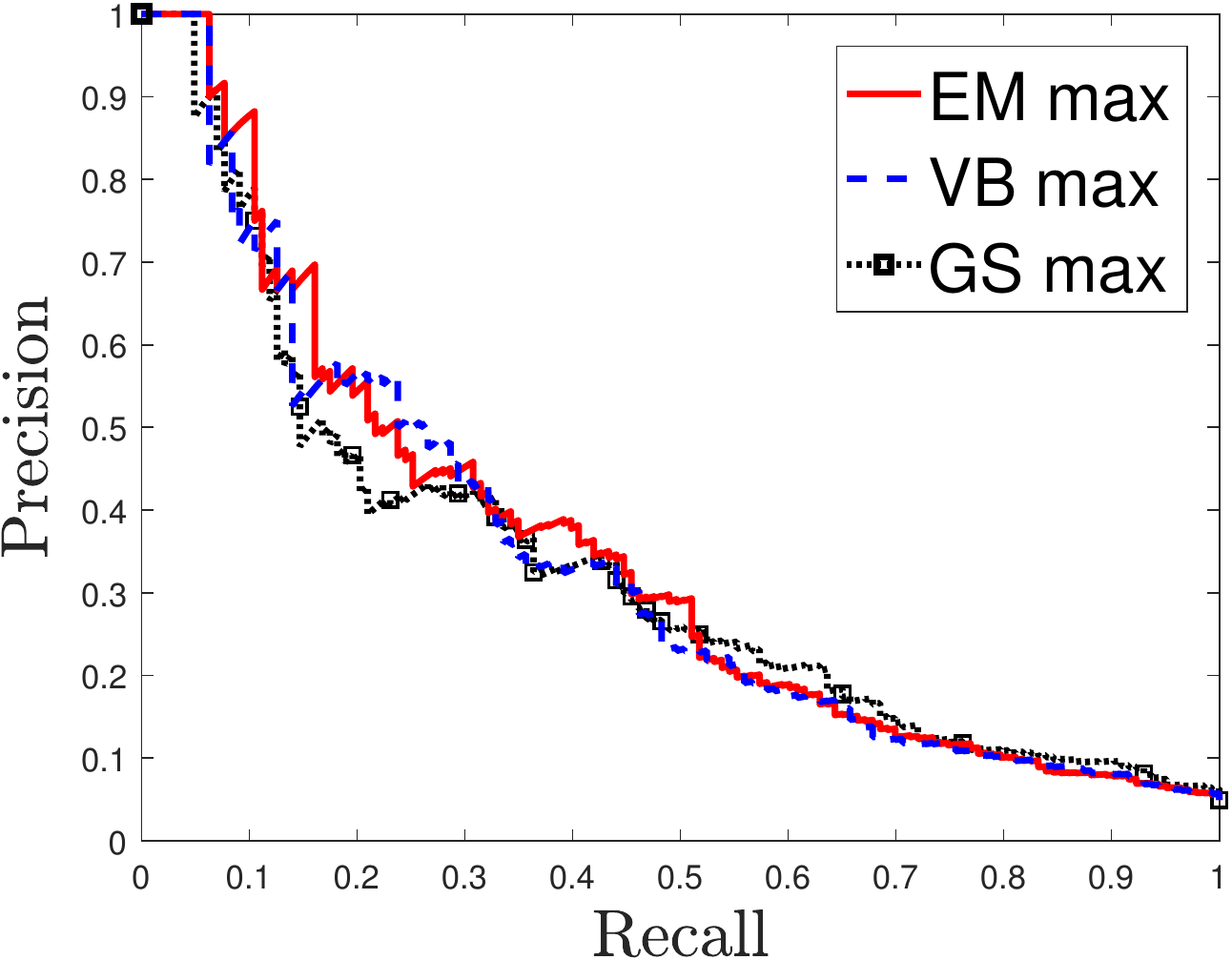}
\label{fig:qmul_best_curves}}%
\hfil
\subfloat[QMUL data --- worst results]{\includegraphics[width = 0.485\columnwidth]{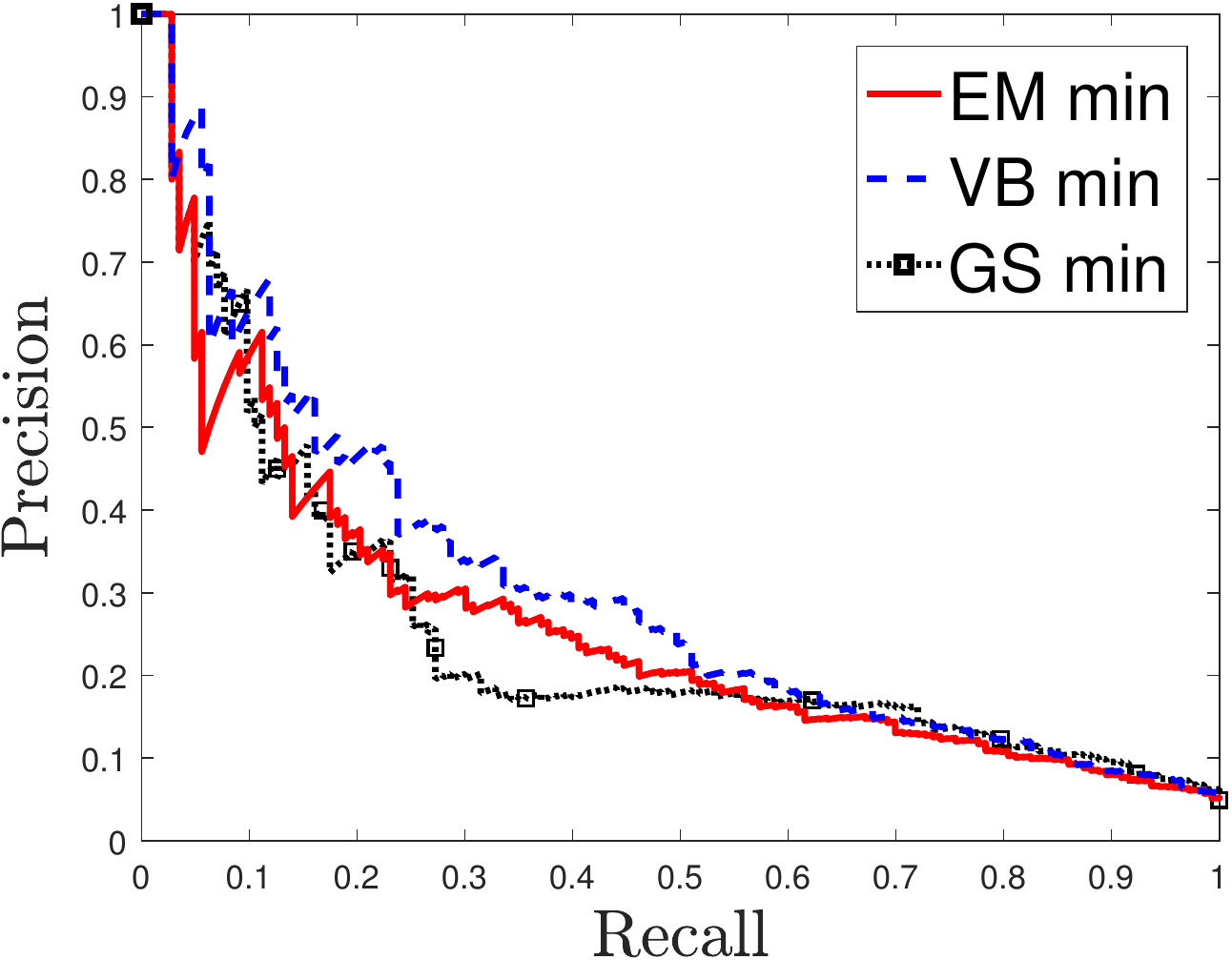}
\label{fig:qmul_worst_curves}}%
\hfil
\subfloat[Idiap data --- best results]{\includegraphics[width = 0.485\columnwidth]{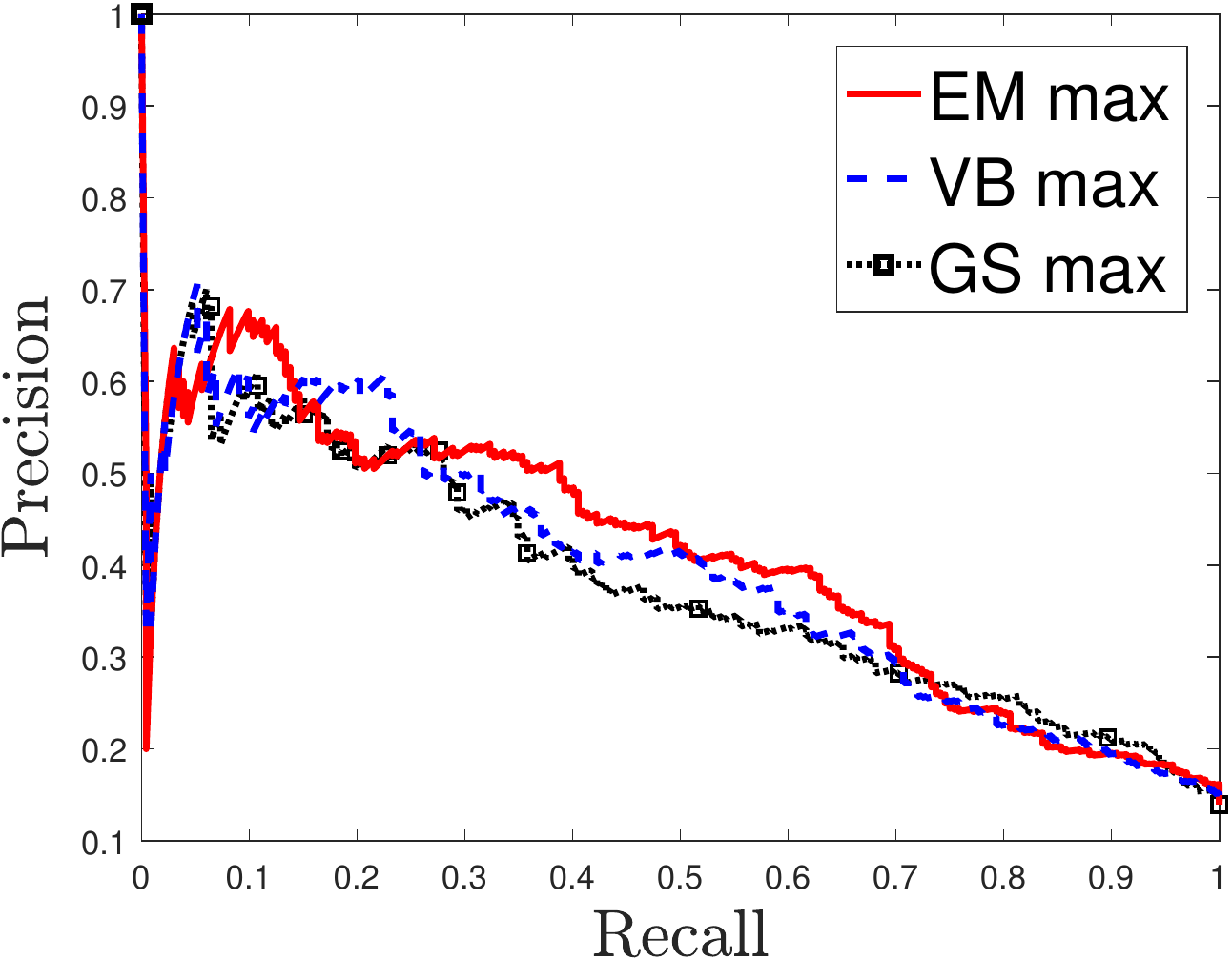}
\label{fig:idiap_best_curves}}%
\hfil
\subfloat[Idiap data --- worst results]{\includegraphics[width = 0.485\columnwidth]{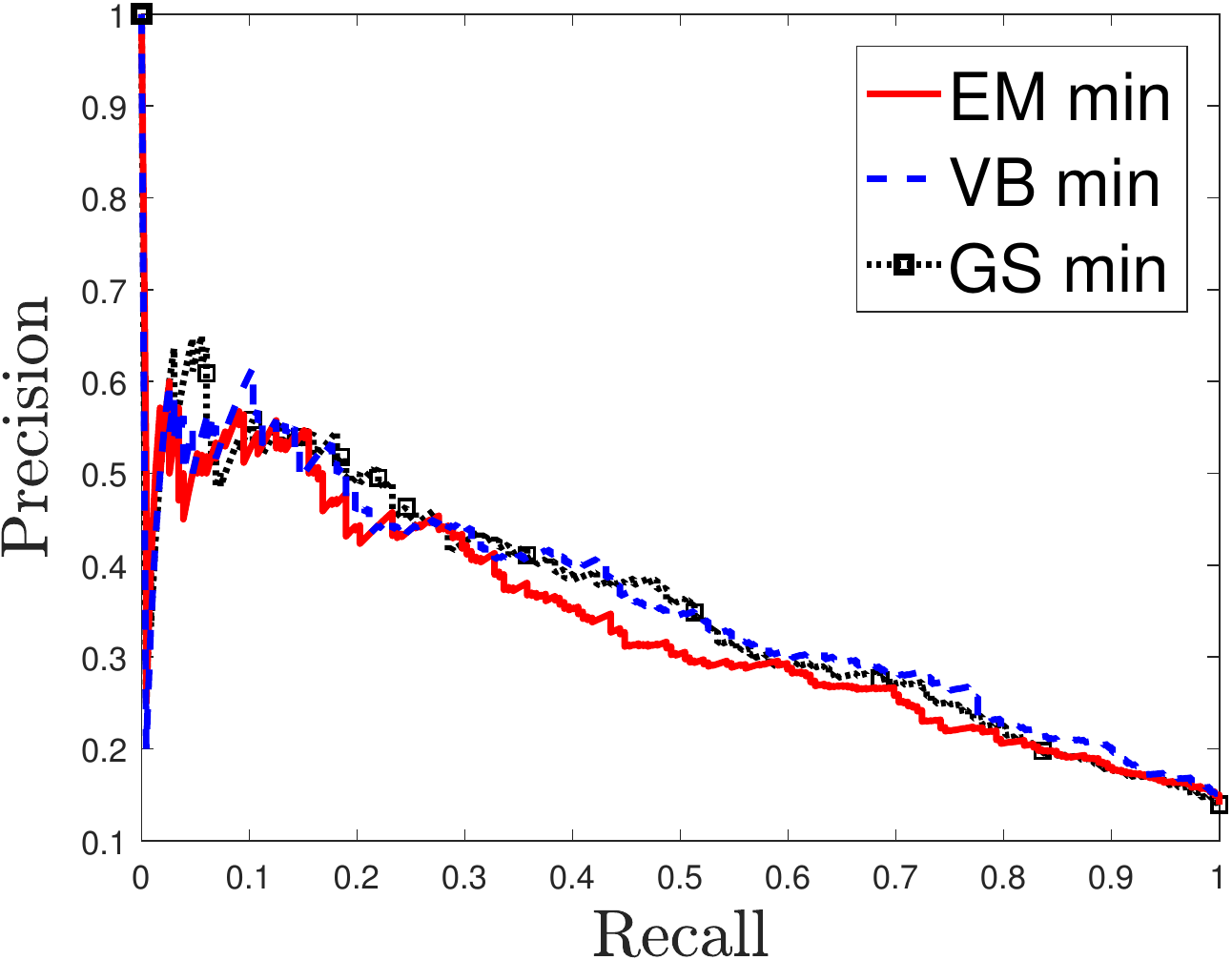}
\label{fig:idiap_worst_curves}}%
\caption{Precision-recall curves with the maximum and minimum areas under curves for the three learning algorithms (maximum and minimum is among all the runs with different initialisations for all groups of hyperparameters settings and all types of marginal likelihood approximations). \protect\subref{fig:qmul_best_curves} presents the ``best'' curves for the QMUL data, i.e., the curves with the maximum area under a curve. \protect\subref{fig:qmul_worst_curves} presents the ``worst'' curves for the QMUL data, i.e., the curves with the minimum area under a curve. \protect\subref{fig:idiap_best_curves} presents the ``best'' curves for the Idiap data, \protect\subref{fig:idiap_worst_curves} --- the ``worst'' curves for the Idiap data.}
\label{fig:best_worst_curves}
\end{figure}
Figure \ref{fig:best_worst_curves} presents the best and the worst precision-recall curves (in terms of the area under them) for the individual runs of the learning algorithms. The figure shows that among the individual runs the EM learning algorithm also demonstrates the most accurate results. Although, the minimum area under the precision-recall curve for the EM learning algorithm is less than the area under the corresponding curve for the VB algorithm. It means that the variance among the individual curves for the EM learning algorithm is larger in comparison with the VB learning algorithm. 

The variance of the precision-recall curves for both VB and GS learning algorithms is relatively small. However, the VB learning algorithm has the curves higher than the curves obtained by the GS learning algorithm. It can be confirmed by examination of the best and worst precision-recall curves (Figure \ref{fig:best_worst_curves}) and the mean values of the area under curves (Figure \ref{fig:mean_results} and Table \ref{tab:mean_area}). 

We also present the results of classification accuracy, i.e., the fraction of the correctly classified documents, for anomaly detection, which can be achieved with some fixed threshold. The best classification accuracy for the EM learning algorithm in both datasets can be found in Table~\ref{tab:accuracy}.

\begin{table}[!t]
\caption{Best classification accuracy for the EM learning algorithm}
\label{tab:accuracy}
\centering
\begin{tabular}{p{2cm} p{1.2cm}}
\toprule
Dataset & Accuracy\\
\midrule
QMUL & 0.9544 \\
Idiap & 0.8891 \\
\bottomrule
\end{tabular}
\end{table}

\subsubsection{Anomaly localisation}
\begin{figure}[!t]
\centering
\subfloat[]{\includegraphics[width=0.4\columnwidth]{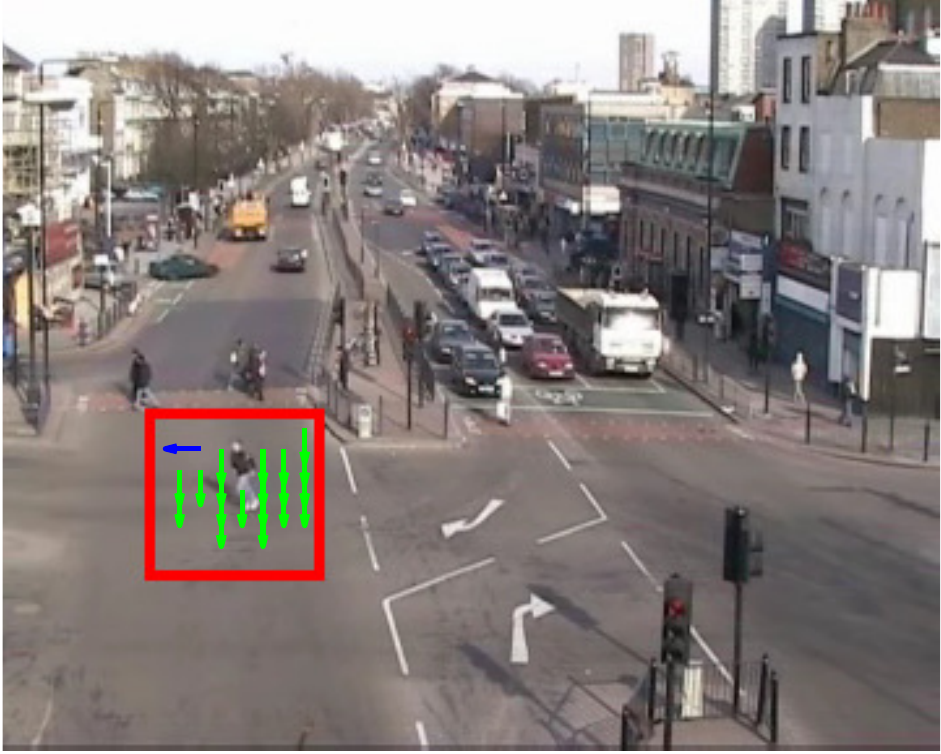}
\label{fig:qmul_loc_1}}%
\hfil
\subfloat[]{\includegraphics[width=0.4\columnwidth]{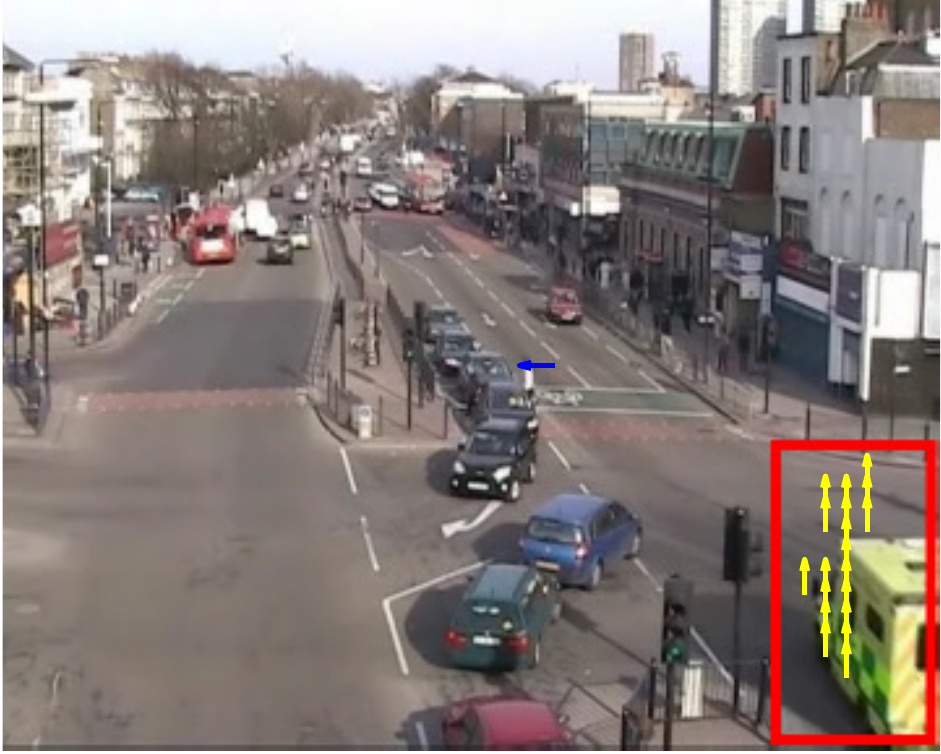}
\label{fig:qmul_loc_2}}%
\hfil
\subfloat[]{\includegraphics[width=0.4\columnwidth]{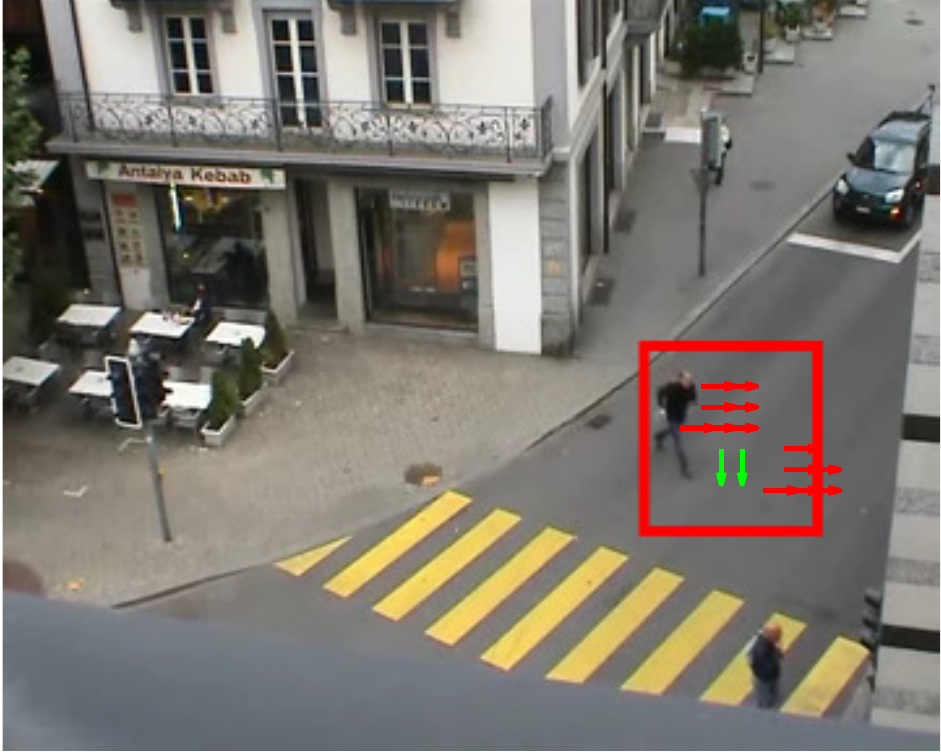}
\label{fig:idiap_loc_1}}%
\hfil
\subfloat[]{\includegraphics[width=0.4\columnwidth]{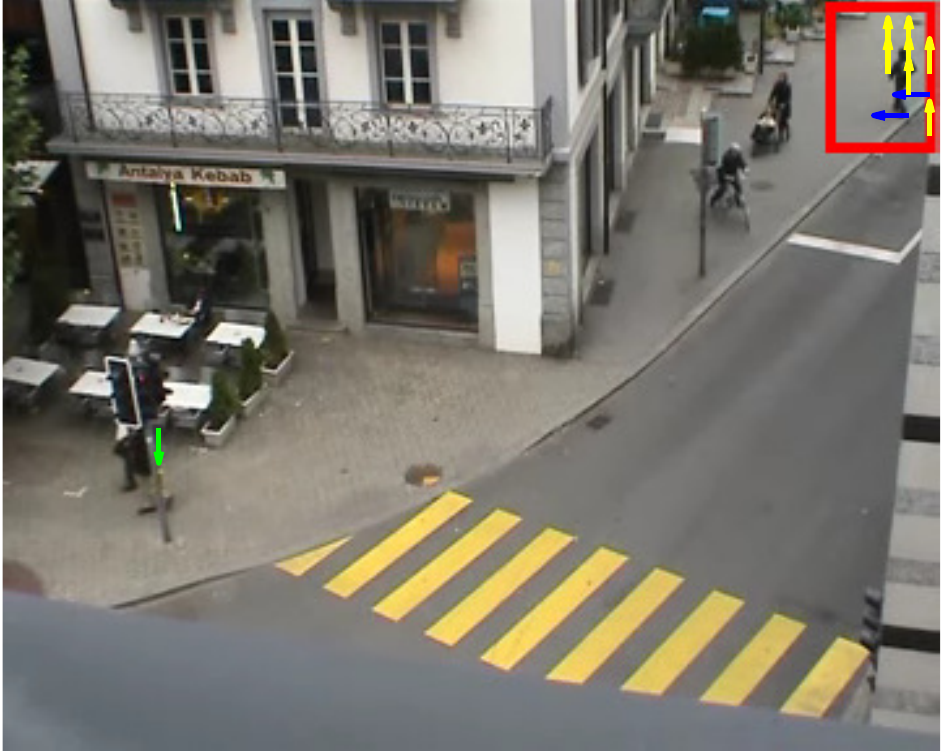}
\label{fig:idiap_loc_2}}%
\caption{Examples of anomalies localisation. The red rectangle is the manual localisation. The arrows represent the visual words with the smallest marginal likelihood, the locations of the arrows are the results of the algorithmic anomaly localisation.}
\label{fig:abnormality_localisation}
\end{figure}

We apply the proposed method for anomaly localisation, presented in Section~\ref{sec:localisation}, and get promising results. We demonstrate the localisation results for the EM learning algorithm with the prior type H+1 on both datasets in Figure~\ref{fig:abnormality_localisation}. The red rectangle is manually set to locate the abnormal events within the frame, the arrows correspond to the visual words with the smallest marginal likelihood computed by the algorithm. It can be seen that the abnormal events correctly localised by the proposed method.

\begin{figure}
\centering
\includegraphics[width=0.9\columnwidth]{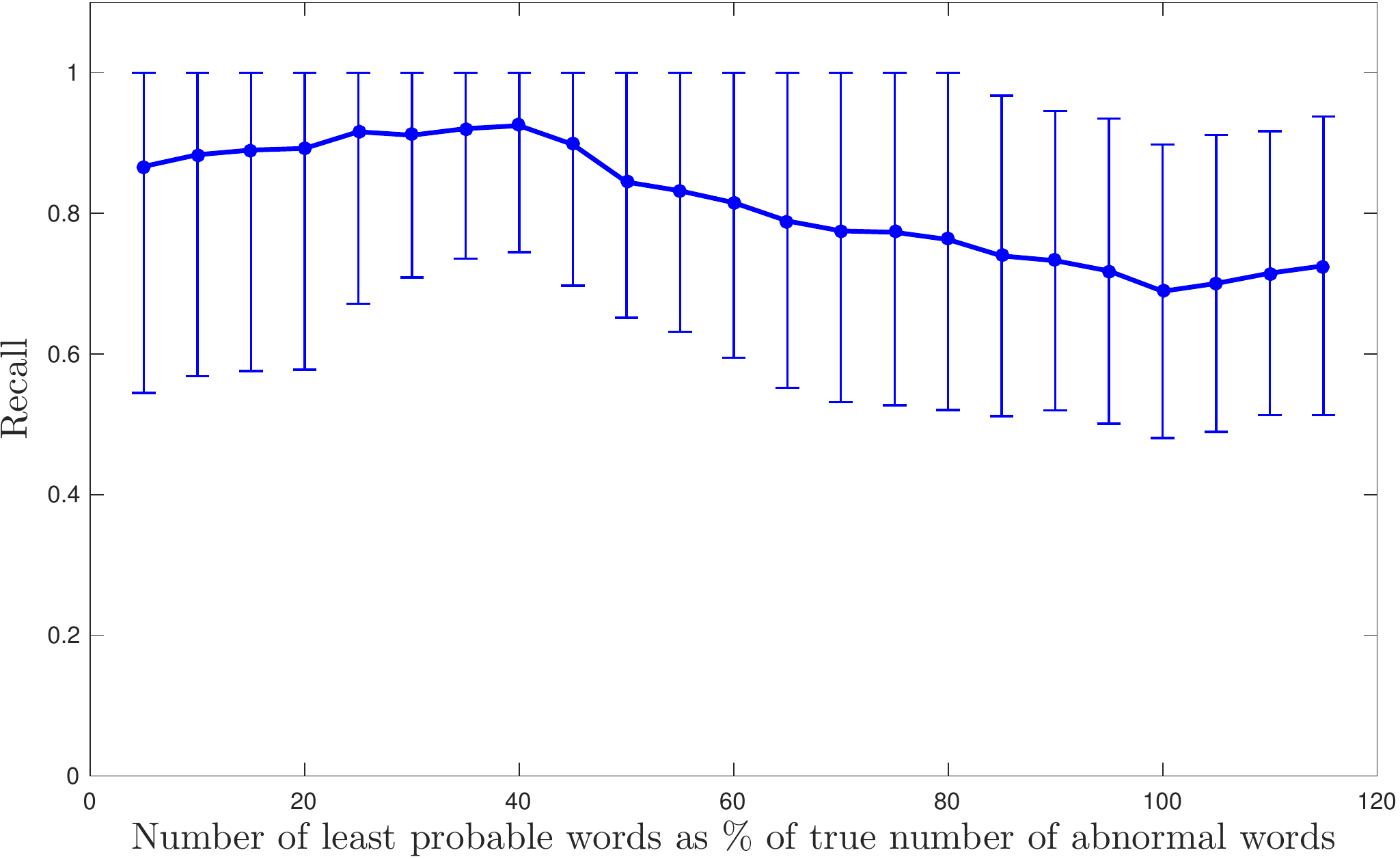}
\caption{Recall results of the proposed anomaly localisation procedure}
\label{fig:quantitative_localisation}
\end{figure}

For quantitative evaluation we analyse $10$ abnormal events ($5$ from each dataset). For each clip for a given number $N_{\text{top}}$ of the least probable words, we measure the recall: $\text{recall} = \dfrac{\text{TP}}{N_{\text{an}}}$, where $N_{\text{an}}$ is the maximum possible number of abnormal words among $N_{\text{top}}$, i.e., $N_{\text{an}} = N_{\text{top}}$ if $N_{\text{top}} \leq N_{\text{total an}}$, where $N_{\text{total an}}$ is the total number of abnormal words, and $N_{\text{an}} = N_{\text{total an}}$ if $N_{\text{top}} > N_{\text{total an}}$. Figure~\ref{fig:quantitative_localisation}  presents the mean results for all events. One can notice, for example, that when the localisation procedure can possibly detect $45\%$ of the total number of abnormal words, it correctly finds $\approx 90\%$ of them.

\section{Conclusions}
\label{sec:conlusion}
This paper presents two learning algorithms for the dynamic
topic model for behaviour analysis in video:  the EM-algorithm
is developed for the MAP estimates of the model parameters and a variational Bayes inference algorithm is developed for calculating the posterior distributions of them. A detailed comparison of these proposed learning algorithms with the Gibbs sampling based algorithm developed in~\cite{Hospedales2011} is presented. The differences and the similarities of the theoretical aspects for all three learning algorithms are well emphasised. The empirical comparison is performed for abnormal behaviour detection using two unlabelled real video datasets. Both proposed learning algorithms demonstrate more accurate results than the algorithm proposed in~\cite{Hospedales2011} in terms of anomaly detection performance. 

The EM learning algorithm demonstrates the best results in terms of the mean values of the performance measure, obtained by the independent runs of the algorithm with different random initialisations. Although, it is noticed that the variance among the precision-recall curves of the individual runs is relatively high. The variational Bayes learning algorithm shows the smaller variance among the precision-recall curves than the EM-algorithm. The results show that the VB algorithm answers are more robust to different initialisation values. However, it is shown that the results of the algorithm are significantly influenced by the choice of the hyperparameters. The hyperparameters require additional tuning before the algorithm can be applied to data. Note that the results of the EM learning algorithm only slightly depend on the choice of the hyperparameters settings. Moreover, the hyperparameters can be even set in such a way as the EM algorithm is applied to obtain the maximum likelihood estimates instead of the maximum a posteriori ones. Both proposed learning algorithms --- EM and VB --- provide more accurate results in comparison to the Gibbs sampling based algorithm. 

We also demonstrate that consideration of marginal likelihoods of visual words rather than visual documents can provide satisfactory results about locations of anomalies within a frame. In our best knowledge the proposed localisation procedure is the first general approach in probabilistic topic modeling that requires only presence of spatial information encoded in visual words. 

\appendices
\section{EM-algorithm derivations}
This Appendix presents the details of the proposed EM learning algorithm derivation. The objective function in the EM-algorithm is:
\begin{align}
&\mathcal{Q}(\boldsymbol\Omega, \boldsymbol\Omega^{\text{old}}) + \log p(\boldsymbol\Omega | \boldsymbol\beta, \boldsymbol\alpha, \boldsymbol\eta, \boldsymbol\gamma) = \nonumber\\
&\sum_{\mathbf{y}_{1:T_{tr}}} \sum_{\mathbf{z}_{1:T_{tr}}} \left( p(\mathbf{y}_{1:T_{tr}}, \mathbf{z}_{1:T_{tr}} | \mathbf{x}_{1:T_{tr}}, \boldsymbol\Omega^{Old}) \times \right.\nonumber\\
&\left. \log{p(\mathbf{x}_{1:T_{tr}}, \mathbf{y}_{1:T_{tr}}, \mathbf{z}_{1:T_{tr}} | \boldsymbol\Omega, \boldsymbol\alpha, \boldsymbol\beta, \boldsymbol\gamma, \boldsymbol\eta)} \vphantom{p(\mathbf{y}_{1:T_{tr}}, \mathbf{z}_{1:T_{tr}} | \mathbf{x}_{1:T_{tr}}, \boldsymbol\Omega^{Old})} \right) + \nonumber\\
&+ \log p(\boldsymbol\Omega | \boldsymbol\beta, \boldsymbol\alpha, \boldsymbol\eta, \boldsymbol\gamma) = \nonumber\\
&= Const + \sum_{z_1 \in \mathcal{Z}} \left( \log{\pi_{z_1}} \, p(z_1 | \mathbf{x}_{1:T_{tr}}, \boldsymbol\Omega^{Old}) \right) + \nonumber\\ 
&\sum_{t = 2}^{T_{tr}} \sum_{z_t \in \mathcal{Z}} \sum_{z_{t-1} \in \mathcal{Z}} \left( \log{\xi_{z_t, z_{t-1}}} \, p(z_t, z_{t-1} | \mathbf{x}_{1:T_{tr}}, \boldsymbol\Omega^{Old}) \right) + \nonumber\\
&\sum_{t = 1}^{T_{tr}} \sum_{i = 1}^{N_t} \sum_{y_{i, t} \in \mathcal{Y}} \left( \log{\phi_{x_{i, t}, y_{i, t}}} \, p(y_{i, t} | \mathbf{x}_{1:T_{tr}}, \boldsymbol\Omega^{Old}) \right) + \nonumber\\
&+ \sum_{t = 1}^{T_{tr}} \sum_{i = 1}^{N_t} \sum_{z_t \in \mathcal{Z}} \sum_{y_{i, t} \in \mathcal{Y}} \left( \log{\theta_{y_{i, t}, z_t}}\, p(y_{i, t}, z_t | \mathbf{x}_{1:T_{tr}}, \boldsymbol\Omega^{Old}) \right) + \nonumber\\
&\sum\limits_{z \in \mathcal{Z}} (\eta_z - 1) \log\pi_{z} + \sum\limits_{z \in \mathcal{Z}}\sum\limits_{z' \in \mathcal{Z}} (\gamma_z - 1) \log\xi_{z, z'} + \nonumber\\
\label{eq:em_maximised_function} 
&\sum\limits_{z \in \mathcal{Z}} \sum\limits_{y \in \mathcal{Y}} (\alpha_y - 1) \log\theta_{y, z} + \sum\limits_{y \in \mathcal{Y}} \sum\limits_{x \in \mathcal{X}} (\beta_x - 1) \log\phi_{x, y}
\end{align}

On the M-step the function (\ref{eq:em_maximised_function}) is maximised with respect to the parameters $\boldsymbol\Omega$ with fixed values for $p(z_1  | \mathbf{x}_{1:T_{tr}}, \boldsymbol\Omega^{Old})$, $p(z_t, z_{t-1} | \mathbf{x}_{1:T_{tr}}, \boldsymbol\Omega^{Old})$, $p(y_{i, t} | \mathbf{x}_{1:T_{tr}}, \boldsymbol\Omega^{Old})$, $p(y_{i, t}, z_t | \mathbf{x}_{1:T_{tr}}, \boldsymbol\Omega^{Old})$. The optimisation problem can be solved separately for each parameter, which leads to the equations (\ref{eq:M:phi}) -- (\ref{eq:M_psi_k,l}).

On the E-step for the efficient implementation the forward-backward steps are developed for the auxiliary variables $\acute{\alpha}_z(t)$ and $\acute{\beta}_z(t)$:
\begin{multline}
\label{eq:alpha_def}
\acute{\alpha}_z(t) \stackrel{\text{def}}{=} p(\mathbf{x}_1, \dotsc, \mathbf{x}_t, z_t = z | \boldsymbol\Omega^{Old}) = \\
\sum\limits_{\mathbf{z}_{1:t-1}} \pi^{Old}_{z_1} \left[ \prod_{\acute{t} = 2}^{t-1} \xi^{Old}_{z_{\acute{t}}, z_{\acute{t}-1}} \right] \left[\prod_{\acute{t} = 1}^{t-1} \prod_{\vphantom{\acute{t}} i = 1}^{N_{\acute{t}}} \sum\limits_{\vphantom{\acute{t}} y \in \mathbf{Y}} \phi^{Old}_{x_{i, \acute{t}}, y} \theta^{Old}_{y, z_{\acute{t}}}\right] \times \\
\xi^{Old}_{z_t = k, z_{t-1}} \prod\limits_{i = 1}^{N_t} \sum\limits_{y \in \mathcal{Y}} \phi^{Old}_{x_{i, t}, y} \theta^{Old}_{y, z_t = z}.
\end{multline}  
Reorganisation of the terms in (\ref{eq:alpha_def}) leads to the recursive expressions (\ref{eq:E:alpha}).

Similarly for $\acute{\beta}_z(t)$:
\begin{multline}
\label{eq:beta_def}
\acute{\beta}_k(t) \stackrel{\text{def}}{=} p(\mathbf{x}_{t+1}, \dotsc, \mathbf{x}_{T_{tr}} | z_t = z, \boldsymbol\Omega^{Old}) = \\
\sum\limits_{\mathbf{z}_{t+1 : T_{tr}}} \xi^{Old}_{z_{t+1}, z_t = z} \left[\prod\limits_{\acute{t} = t+2}^{T_{tr}} \xi^{Old}_{z_{\acute{t}}, z_{\acute{t}-1}} \right] \prod\limits_{\acute{t} = t+1}^{T_{tr}} \prod\limits_{\vphantom{\acute{t}} i = 1}^{N_{\acute{t}}} \sum\limits_{\vphantom{\acute{t}} y \in \mathcal{Y}} \phi^{Old}_{x_{i, \acute{t}}, y} \theta^{Old}_{y, z_{\acute{t}}}.
\end{multline}
The recursive formula (\ref{eq:E:beta}) is obtained by interchanging the terms in (\ref{eq:beta_def}).

The required posterior of the hidden variables terms $p(z_1  | \mathbf{x}_{1:T_{tr}}, \boldsymbol\Omega^{Old})$, $p(z_t, z_{t-1} | \mathbf{x}_{1:T_{tr}}, \boldsymbol\Omega^{Old})$, $p(y_{i, t} | \mathbf{x}_{1:T_{tr}}, \boldsymbol\Omega^{Old})$, $p(y_{i, t}, z_t | \mathbf{x}_{1:T_{tr}}, \boldsymbol\Omega^{Old})$ are then expressed via the axillary variables $\acute{\alpha}_z(t)$ and $\acute{\beta}_z(t)$, which leads to (\ref{eq:E:z_t}) -- (\ref{eq:E:y_i,t}). 

\section{VB algorithm derivations}
This Appendix presents the details of the proposed variational Bayes inference derivation. We have separated the parameters and the hidden variables. Let us consider the update formula of the variational Bayes inference scheme~\cite{Murphy2012} for the parameters:
\begin{align}
\label{eq:q_param_full}
&\log q(\boldsymbol\Omega) = Const + \nonumber\\
&\mathbb{E}_{q(\mathbf{y}_{1:T_{tr}}, \mathbf{z}_{1:T_{tr}})} \log p(\mathbf{x}_{1:T_{tr}}, \mathbf{y}_{1:T_{tr}}, \mathbf{z}_{1:T_{tr}}, \boldsymbol\Omega | \boldsymbol\eta, \boldsymbol\gamma, \boldsymbol\alpha, \boldsymbol\beta)= \nonumber\\
&Const + \mathbb{E}_{q(\mathbf{y}_{1:T_{tr}}, \mathbf{z}_{1:T_{tr}})} \left( \sum\limits_{z \in \mathcal{Z}} \left(\eta_z - 1 \right) \log \pi_z + \right.\nonumber\\
&\sum\limits_{z \in \mathcal{Z}} \sum\limits_{\tilde{z} \in \mathcal{Z}} \left(\gamma_{\tilde{z}} - 1\right) \log \xi_{\tilde{z}, z} + \sum\limits_{z \in \mathcal{Z}} \sum\limits_{y \in \mathcal{Y}} \left(\alpha_y - 1\right) \log \theta_{y, z} + \nonumber\\
&\sum\limits_{y \in \mathcal{Y}} \sum\limits_{x \in \mathcal{X}} \left(\beta_x - 1 \right) \log \phi_{x, y} + \sum\limits_{z \in \mathcal{Z}} \mathbb{I}(z_1 = z) \log \pi_z + \nonumber\\
&\sum\limits_{t = 2}^{T_{tr}} \sum\limits_{z \in \mathcal{Z}} \sum\limits_{\tilde{z} \in \mathcal{Z}} \mathbb{I}(z_t = \tilde{z}) \mathbb{I}(z_{t-1} = z) \log\xi_{\tilde{z}, z} + \nonumber\\
&\sum\limits_{t = 1}^{T_{tr}} \sum\limits_{i = 1}^{N_t} \sum\limits_{y \in \mathcal{Y}} \mathbb{I}\left(y_{i, t} = y\right) \log \phi_{x_{i, t}, y} + \nonumber\\
&\left. \sum\limits_{t = 1}^{T_{tr}} \sum\limits_{i = 1}^{N_t} \sum\limits_{z \in \mathcal{Z}} \sum\limits_{y \in \mathcal{Y}} \mathbb{I}(y_{i, t} = y) \mathbb{I}(z_t = z) \log \theta_{y, z} \right)
\end{align} 

One can notice that $\log q(\boldsymbol\Omega)$ is further factorised as in (\ref{eq:q_param_factorisation}). Now each factorisation term can be considered independently. Derivations of the equations (\ref{eq:VB:beta}) -- (\ref{eq:VB:gamma}) are very similar to each other. We provide the derivation only of the term $q(\boldsymbol{\Phi})$:
\begin{align}
&\log q(\boldsymbol\Phi) = Const + \nonumber\\
&\mathbb{E}_{q(\mathbf{y}_{1:T_{tr}}, \mathbf{z}_{1:T_{tr}})} \left(\vphantom{\sum\limits_{t = 1}^{T_{tr}} \sum\limits_{i = 1}^{N_t} \sum\limits_{y \in \mathcal{Y}}}\sum\limits_{y \in \mathcal{Y}} \sum\limits_{x \in \mathcal{X}} \left(\beta_x - 1 \right) \log \phi_{x, y} + \right. \nonumber\\
&\left.\sum\limits_{t = 1}^{T_{tr}} \sum\limits_{i = 1}^{N_t} \sum\limits_{y \in \mathcal{Y}} \mathbb{I}\left( y_{i, t} = y \right) \log \phi_{x_{i, t}, y} \right)  = \nonumber\\
&Const + \sum\limits_{y \in \mathcal{Y}} \sum\limits_{x \in \mathcal{X}} \left(\beta_x - 1 \right) \log \phi_{x, y} + \nonumber\\
&\sum\limits_{t = 1}^{T_{tr}} \sum\limits_{i = 1}^{N_t} \sum\limits_{y \in \mathcal{Y}} \log \phi_{x_{i, t}, y} \underbrace{\mathbb{E}_{q(\mathbf{y}_{1:T_{tr}}, \mathbf{z}_{1:T_{tr}})}\left( \mathbb{I}\left(y_{i, t} = y\right)\right)}_{q(y_{i, t} = y)} = \nonumber\\
&Const + \nonumber\\
\label{eq:log_q_phi}
&\sum\limits_{y \in \mathcal{Y}} \sum\limits_{x \in \mathcal{X}} \log \phi_{x, y} \left( \beta_x - 1 + \sum\limits_{t = 1}^{T_{tr}} \sum\limits_{i = 1}^{N_t} \mathbb{I}(x_{i, t} = x) q(y_{i,t} = y) \right)
\end{align}   
It can be noticed from (\ref{eq:log_q_phi}) that the distribution of $\boldsymbol\Phi$ is a product of the Dirichlet distributions (\ref{eq:VB:beta}).

The update formula in the variational Bayes inference scheme for the hidden variables is as follows:
\begin{align}
&\log q(\mathbf{y}_{1:T_{tr}}, \mathbf{z}_{1:T_{tr}}) =  Const + \nonumber\\
&\mathbb{E}_{q(\boldsymbol\pi)q(\boldsymbol\Xi)q(\boldsymbol\Theta)q(\boldsymbol\Phi)} \log p(\mathbf{x}_{1:T_{tr}}, \mathbf{y}_{1:T_{tr}}, \mathbf{z}_{1:T_{tr}}, \boldsymbol\Omega | \boldsymbol\eta, \boldsymbol\gamma, \boldsymbol\alpha, \boldsymbol\beta) = \nonumber\\
&Const + \sum\limits_{z \in \mathcal{Z}} \mathbb{I}\left(z_1 = z\right) \mathbb{E}_{q(\boldsymbol\pi)} \log \pi_z + \nonumber\\
&\sum\limits_{t = 2}^{T_{tr}} \sum\limits_{z \in \mathcal{Z}} \sum\limits_{\tilde{z} \in \mathcal{Z}} \mathbb{I}\left(z_t = \tilde{z}\right) \mathbb{I}\left(z_{t-1} = z\right) \mathbb{E}_{q(\boldsymbol\Xi)} \log \xi_{\tilde{z}, z} + \nonumber\\
&\sum\limits_{t = 1}^{T_{tr}} \sum\limits_{i = 1}^{N_t} \sum\limits_{y \in \mathcal{Y}} \mathbb{I}\left(y_{i, t} = y\right) \mathbb{E}_{q(\boldsymbol\Phi)} \log \phi_{x_{i, t}, y} + \nonumber\\
\label{eq:log_q_yz_beginning}
&\sum\limits_{t = 1}^{T_{tr}} \sum\limits_{i = 1}^{N_t} \sum\limits_{z \in \mathcal{Z}} \sum\limits_{y \in \mathcal{Y}} \mathbb{I}\left(y_{i, t} = y \right) \mathbb{I} \left(z_t = z\right) \mathbb{E}_{q(\boldsymbol\Theta)} \log \theta_{y, z}
\end{align}
We know from the parameters update~(\ref{eq:VB:beta}) -- (\ref{eq:VB:gamma}) that their distributions are Dirichlet. Therefore, $\mathbb{E}_{q(\boldsymbol\pi)}\log \pi_z = \psi\left(\tilde{\eta}_z\right) - \psi\left(\sum_{z' \in \mathcal{Z}} \tilde{\eta}_{z'}\right)$ and similarly for all the other expected value expressions.

Using the introduced notations (\ref{eq:VB:introduced_pi}) -- (\ref{eq:VB:introduced_theta}) the update formula~(\ref{eq:log_q_yz_beginning}) for the hidden variables can be then expressed as:   
\begin{align}
&\log q(\mathbf{y}_{1:{T_{tr}}}, \mathbf{z}_{1:T_{tr}}) = Const + 
\sum\limits_{z \in \mathcal{Z}} \mathbb{I}\left(z_1 = z\right) \log \tilde{\pi}_z + \nonumber\\
&\sum\limits_{t = 2}^{T_{tr}} \sum\limits_{z \in \mathcal{Z}} \sum\limits_{\tilde{z} \in \mathcal{Z}} \mathbb{I}\left(z_t = \tilde{z}\right) \mathbb{I}\left(z_{t-1} = z\right) \log \tilde{\xi}_{\tilde{z}, z} + \nonumber\\
&\sum\limits_{t = 1}^{T_{tr}} \sum\limits_{i = 1}^{N_t} \sum\limits_{y \in \mathcal{Y}} \mathbb{I}\left(y_{i, t} = y\right) \log\tilde{\phi}_{x_{i, t}, y} + \nonumber\\
&\sum\limits_{t = 1}^{T_{tr}} \sum\limits_{i = 1}^{N_t} \sum\limits_{z \in \mathcal{Z}} \sum\limits_{y \in \mathcal{Z}} \mathbb{I} \left(y_{i, t} = y\right) \mathbb{I}\left(z_t = z\right) \log\tilde{\theta}_{y, z}
\end{align}

The approximated distribution of the hidden variables is then:
\begin{multline}
\label{eq:q_yz}
q(\mathbf{y}_{1:T_{tr}}, \mathbf{z}_{1:T_{tr}}) = \\
\dfrac{1}{\tilde{K}} \tilde{\pi}_{z_1} \left[\prod\limits_{t = 2}^{T_{tr}} \tilde{\xi}_{z_t, z_{t-1}}\right] \prod\limits_{t = 1}^{T_{tr}} \prod\limits_{i = 1}^{N_t} \tilde{\phi}_{x_{i, t}, y_{i, t}} \tilde{\theta}_{y_{i, t}, z_t},
\end{multline}
where $\tilde{K}$ is a normalisation constant. Note that the expression of the true posterior distribution of the hidden variables is the same up to replacing the true parameters variables with the corresponding tilde variables:
\begin{multline}
p(\mathbf{y}_{1:T_{tr}}, \mathbf{z}_{1:T_{tr}} | \mathbf{x}_{1:T_{tr}}, \boldsymbol\Omega) = \\
\dfrac{1}{K} \pi_{z_1} \left[\prod\limits_{t = 2}^{T_{tr}} \xi_{z_t, z_{t-1}} \right] \prod\limits_{t = 1}^{T_{tr}} \prod\limits_{i = 1}^{N_t} \phi_{x_{i, t}, y_{i, t}} \theta_{y_{i, t}, z}
\end{multline}
Therefore, to compute the required expressions of the hidden variables $q(z_1 = z)$, $q(z_{t-1} = z, z_t = z')$, $q(y_{i, t} = y, z_t = z)$ and $q(y_{i, t} = y)$ one can use the same forward-backward procedure and update formula as in the E-step of the EM-algorithm replacing all the parameter variables with the corresponding introduced tilde variables. 

\section*{Acknowledgments}

Olga Isupova and Lyudmila Mihaylova would like to thank the support from the EC Seventh Framework Programme [FP7 2013-2017] TRAcking in compleX sensor systems (TRAX) Grant agreement no.: 607400. Lyudmila Mihaylova also acknowledges the UK Engineering and Physical Sciences Research Council (EPSRC) for the support via the Bayesian Tracking and Reasoning over Time (BTaRoT) grant EP/K021516/1.

\bibliographystyle{IEEEtran}
\bibliography{TNNLS-2016-P-6800-R1-Biblist}
\vspace{-2em}
\begin{IEEEbiography}[{\includegraphics[width=1in,height=1.25in,clip,keepaspectratio]{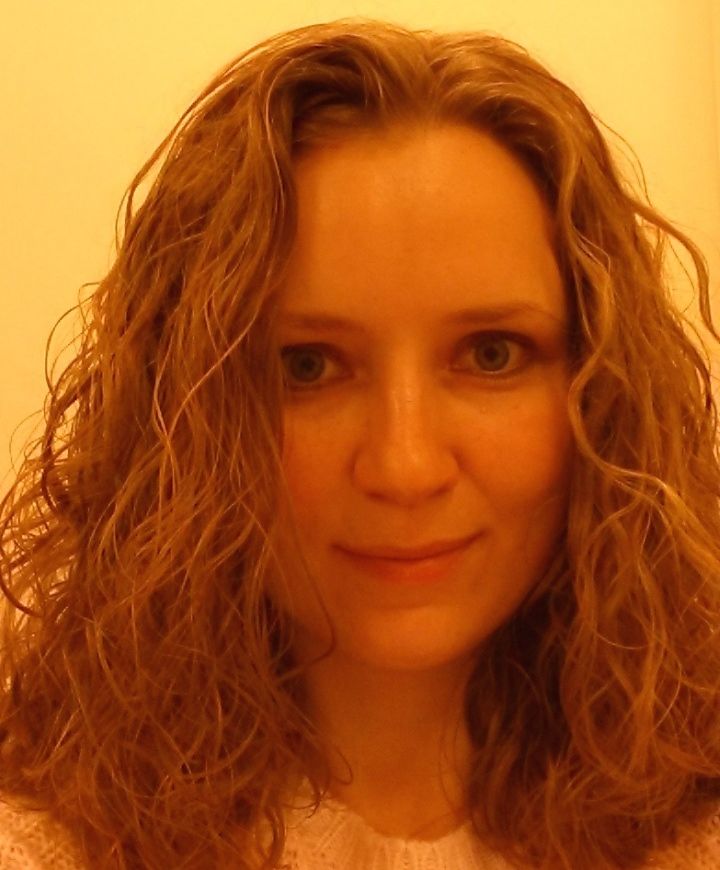}}]{Olga Isupova}
is a PhD student at the Department of Automatic Control and Systems Engineering at the University of Sheffield and an Early Stage Researcher in the FP7 Programme TRAX. She received the Specialist (eq. to M.Sc.) degree in Applied Mathematics and Computer Science, 2012, from Lomonosov Moscow State University, Moscow, Russia. Her research is on machine learning, Bayesian nonparametrics, anomaly detection.  
\end{IEEEbiography}
\vspace{-3.5em}
\begin{IEEEbiography}[{\includegraphics[width=1in,height=1.25in,clip,keepaspectratio]{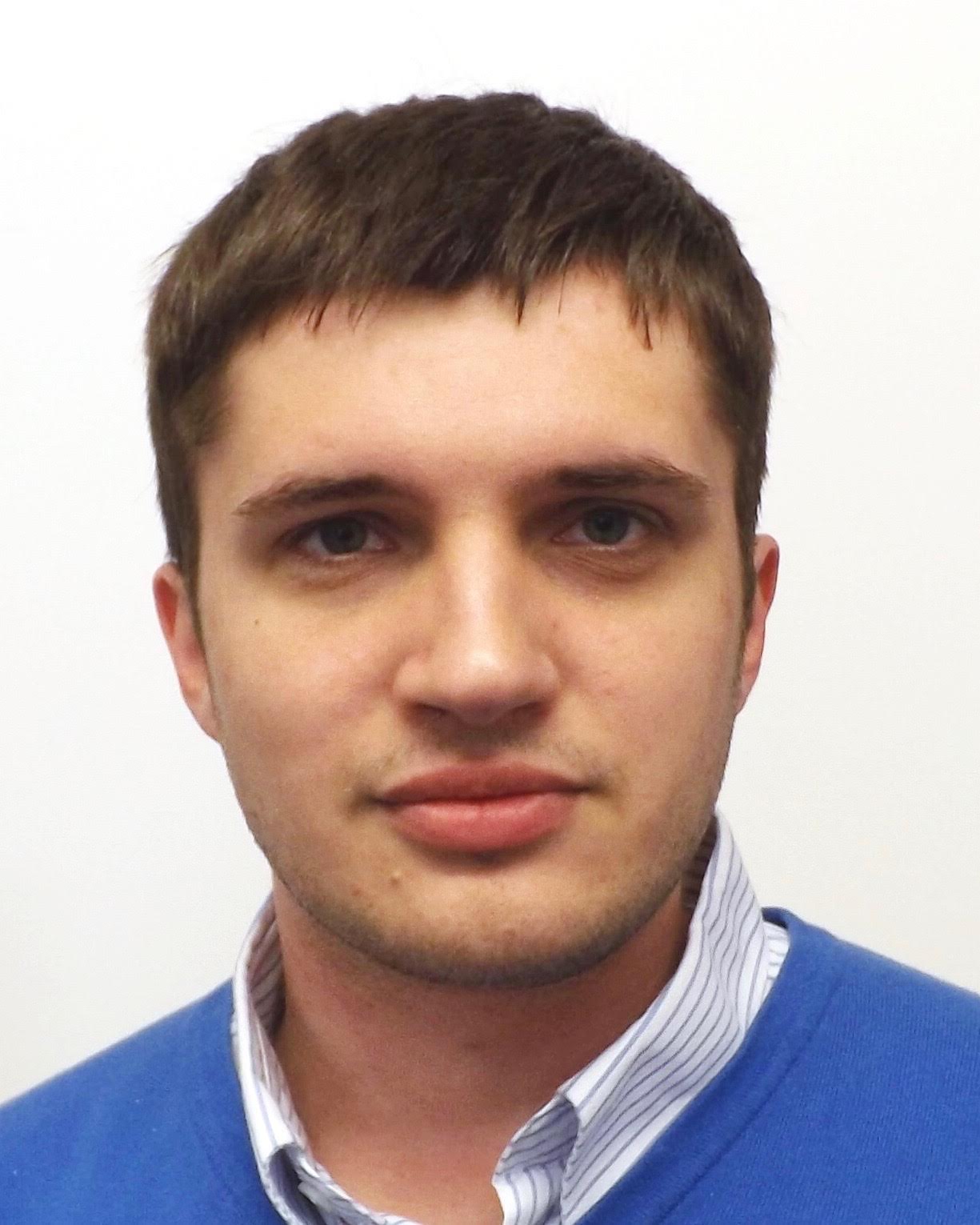}}]{Danil Kuzin}
is a PhD student at the Department of Automatic Control and Systems Engineering at the University of Sheffield and an engineer at the Rinicom, Ltd. He received the Specialist degree in Applied Mathematics and Computer Science, 2012, from Lomonosov Moscow State University, Moscow, Russia. His research is mainly in sparse modelling for video. His other research interests include nonparametric Bayes and deep reinforcement learning.
\end{IEEEbiography}
\vspace{-4em}
\begin{IEEEbiography}
[{\includegraphics[width=1in,height=1.25in,clip,keepaspectratio]{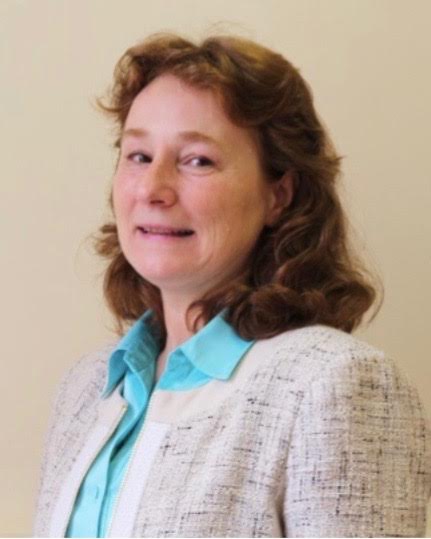}}] {Lyudmila Mihaylova} (M'98,
SM'2008) is Professor of Signal Processing and Control at the
Department of Automatic Control and Systems Engineering at the
University of Sheffield, United Kingdom. Her research is in the
areas of machine learning and autonomous systems with various
applications such as navigation, surveillance and sensor network
systems. She has given a number of talks and tutorials, including
the plenary talk for the IEEE Sensor Data Fusion 2015 (Germany),
invited talks University of California, Los Angeles, IPAMI Traffic
Workshop 2016 (USA), IET ICWMMN 2013 in Beijing, China. Prof.
Mihaylova is an Associate Editor of the IEEE Transactions on
Aerospace and Electronic Systems and of the Elsevier Signal
Processing Journal. She was elected in March 2016 as a president of
the International Society of Information Fusion (ISIF). She is on
the board of Directors of ISIF and a Senior IEEE member. She was the
general co-chair IET Data Fusion $\&$ Target Tracking 2014 and 2012
Conferences, Program co-chair for the 19th International Conference
on Information Fusion, 2016, academic chair of
Fusion 2010 conference.
\end{IEEEbiography}

\end{document}